\documentclass[11pt]{article}

\usepackage[preprint]{acl}

\usepackage{times}
\usepackage{latexsym}
\usepackage{amsmath}
\usepackage{algorithm}
\usepackage{algpseudocode}

\usepackage{fontawesome}

\usepackage[dvipsnames]{xcolor}
\definecolor{plot_red}{RGB}{234, 107, 102}
\definecolor{new_red}{RGB}{254, 109, 115}
\definecolor{new_yellow}{RGB}{255, 203, 1}

\usepackage{booktabs}
\usepackage{multirow} % for borders and merged ranges
\usepackage{soul}% for underlines
\usepackage{xcolor,colortbl} % for cell colors
\usepackage{changepage,threeparttable} % for wide tables
\usepackage{tcolorbox}
\tcbuselibrary{skins, breakable}

\usepackage[T1]{fontenc}
\usepackage[utf8]{inputenc}

\usepackage{microtype}

\usepackage{inconsolata}

\usepackage{graphicx}
\usepackage{subcaption}

\title{From Where Words Come: Efficient Regularization of Code Tokenizers Through Source Attribution}

\author{
\noindent
\begin{minipage}[t]{0.3\textwidth}
    \centering
    Pavel Chizhov$^{1,2}$\\
\end{minipage}% <--- Important: No newline here
\hfill
\begin{minipage}[t]{0.3\textwidth}
    \centering
    Egor Bogomolov$^{2,3}$\\
\end{minipage}% <--- Important: No newline here
\hfill
\begin{minipage}[t]{0.33\textwidth}
    \centering
    Ivan P. Yamshchikov$^{1}$\\
\end{minipage}
\vspace{3mm} \\
$^{1}$CAIRO, Technical University of Applied Sciences Würzburg-Schweinfurt \\
$^{2}$JetBrains Research ~~~$^{3}$TU Delft
\vspace{2mm}\\
Correspondence to: \texttt{pavel.chizhov@thws.de}
}

\begin{document}
\maketitle
\begin{abstract}
    Efficiency and safety of Large Language Models (LLMs), among other factors, rely on the quality of tokenization. A good tokenizer not only improves inference speed and language understanding but also provides extra defense against jailbreak attacks and lowers the risk of hallucinations. In this work, we investigate the efficiency of code tokenization, in particular from the perspective of data source diversity. We demonstrate that code tokenizers are prone to producing unused, and thus under-trained, tokens due to the imbalance in repository and language diversity in the training data, as well as the dominance of source-specific, repetitive tokens that are often unusable in future inference. By modifying the BPE objective and introducing merge skipping, we implement different techniques under the name Source-Attributed BPE (SA-BPE) to regularize BPE training and minimize overfitting, thereby substantially reducing the number of under-trained tokens while maintaining the same inference procedure as with regular BPE. This provides an effective tool suitable for production use.
\end{abstract}

\begin{center}
\faGithub~\footnotesize{\href{https://github.com/pchizhov/sa-bpe}{\texttt{pchizhov/sa-bpe}}}
\end{center}

\section{Introduction}
\label{sec:introduction}

Subword tokenization is the most common approach for vocabulary building in language models~\citep{Devlin2019BERTPO,NEURIPS2020_1457c0d6,grattafiori2024llama3herdmodels,gemmateam2025gemma3technicalreport}. This approach is normally free from linguistic rules and thus fits a variety of arbitrary mixtures of languages, which are commonly not fixed in large models. The main optimization criterion of subword tokenization is text compression. Currently, the most widely used subword tokenization approaches are variants of Byte-Pair Encoding (BPE; \citealp{gage1994new, sennrich-etal-2016-neural}). This algorithm builds a vocabulary by iteratively merging the most frequently co-occurring token pairs, starting from the initial byte sequence. On inference, the learned sequence of merges is applied to a text in the same order as during training.

However, the efficiency of tokenization algorithms comes at the cost of limited control over word segmentation and the resulting vocabulary. One common issue in BPE tokenizers is under-trained tokens~\citep{land-bartolo-2024-fishing}. Such tokens are present in tokenizers but are rarely or never seen during model training. Therefore, these tokens lack adequate embedding representations within the model and merely clutter the vocabulary. Furthermore, because the model is unaware of suitable contexts for these tokens, they can lead to hallucinations or even serve as a means of token-level jailbreak attacks.

Textual data in coding languages presents a unique challenge for tokenization algorithms. First of all, regardless of the variety of coding languages, English remains the main language for language syntax, while other languages might be present as natural language in comments and docstrings. Second, punctuation and formatting play a large role in code understanding. Finally, variable and function names are frequent and specifically formatted parts of the data: they often contain multiple words without whitespace separation, e.g., using naming conventions such as \texttt{CamelCase} and \texttt{snake\_case}.

\begin{figure*}[t!]
    \begin{subfigure}[b]{0.67\textwidth}
        \includegraphics[width=\textwidth]{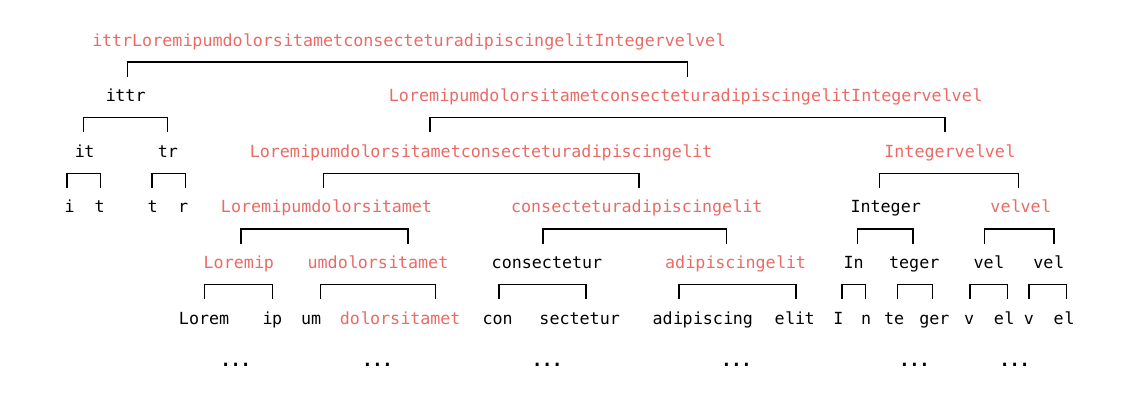}
        \subcaption{The last six levels of the merge tree for token number 48263 in StarCoder2. The token itself is shown at the top of the tree; merges are represented by square brackets.}
        \label{fig:starcoder-tree}
    \end{subfigure}
    \hfill
    \begin{subfigure}[b]{0.3\textwidth}
        \includegraphics[width=\textwidth]{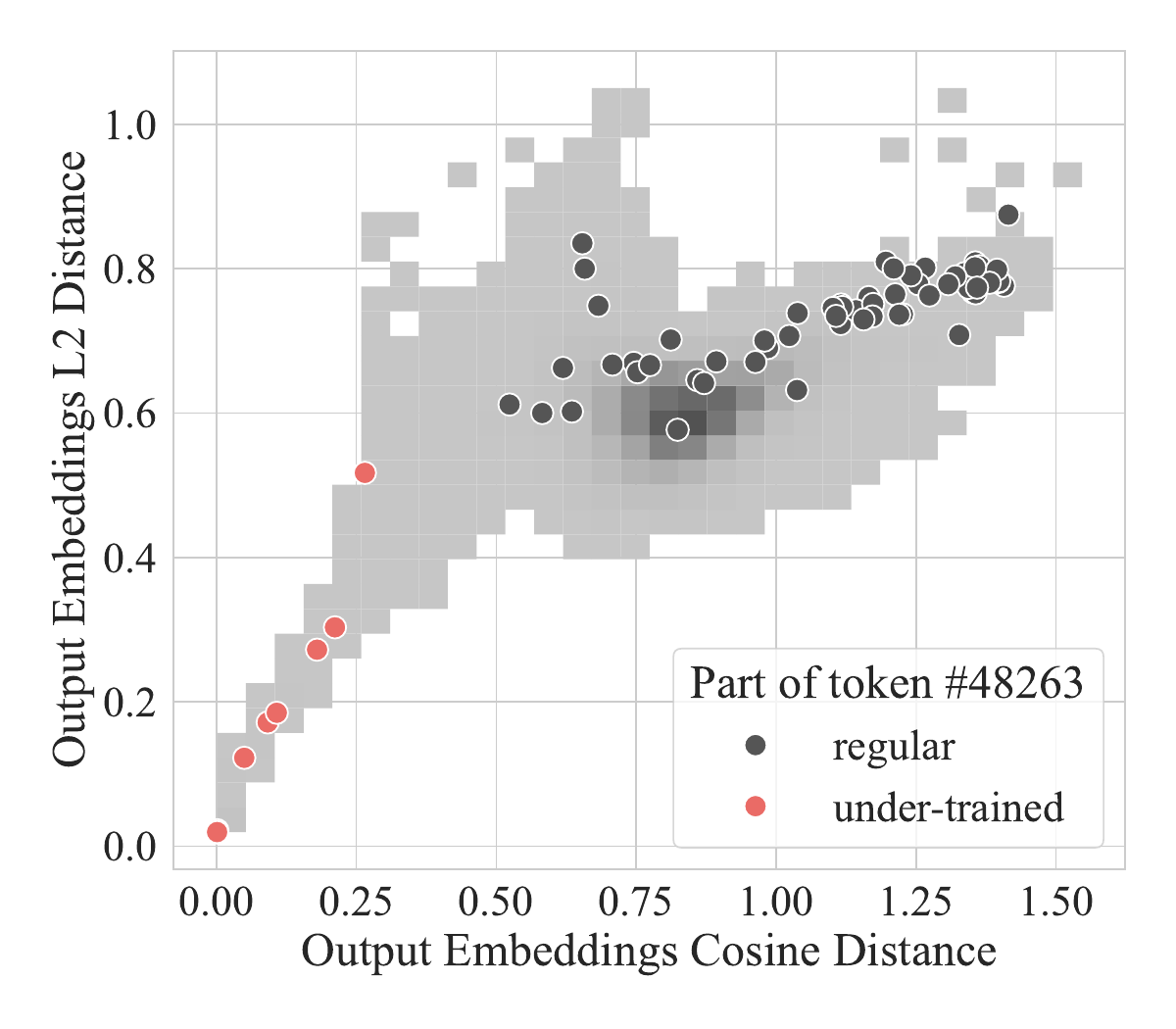}
        \subcaption{Under-trained token indicators for the StarCoder2 tokenizer.}
        \label{fig:starcoder-magikarp}
    \end{subfigure}
    \caption{Under-trained token example from StarCoder2, token number 48263. We show \textbf{(a)} the last steps of the merge tree and \textbf{(b)} the under-trained token indicators for StarCoder2 3B computed with the \texttt{magikarp} library. The tokens closer to $(0, 0)$ are similar to known under-trained tokens and thus are more likely to be under-trained. Such under-trained tokens that are used to build token number 48263 are highlighted in \textcolor{plot_red}{red} in both figures.}
    \label{fig:starcoder}
\end{figure*}

To illustrate the problem of under-trained tokens in coding models, we show an outstanding useless token in the StarCoder2 tokenizer (see Figure~\ref{fig:starcoder-tree}). The token is likely a name of a variable or a method written in the \texttt{CamelCase} style. The main content of the token is a placeholder string \textit{``Lorem ipsum dolor...''} written with a typo (``ipum'' instead of ``ipsum''). Thus, even if another project theoretically included a similar use of this common placeholder phrase, this token or its components would be useless due to the typo. As we highlight in Figure~\ref{fig:starcoder}, this token is included in the tokenizer along with all its constituent parts, many of which also become under-trained. Other examples include repositories where the names were compressed or randomized and are therefore not informative.

In this work, we investigate the issue of under-trained tokens in code language models and propose \textbf{Source-Attributed BPE (SA-BPE)}, a set of BPE modifications to address the problem of overfitting during tokenizer training. We approach this from the perspective of regularization and introduce it in a form opposite to TF-IDF~\cite{salton_introduction_1983}: we de-prioritize terms that appear in only a small subset of documents. In particular, our contributions include:

\begin{itemize}
    \item We analyze under-trained tokens in the BPE tokenizers across common code models and formulate the main sources of such tokens.
    \item We propose SA-BPE, several heuristic modifications that introduce regularization into the BPE algorithm and help mitigate overfitting in coding tokenizers.
    \item We train several small language models and show that our modifications substantially reduce the number of under-trained tokens.
\end{itemize}

\begin{table*}[htbp]
    \centering
    \small
    \begin{tabular}{lccc}
    \toprule
    & \textbf{Qwen2.5 Coder 3B} & \textbf{CodeGemma 7B} & \textbf{StarCoder2 3B}  \\
    \textbf{Category} & 2\% of 151k tokens & 2\% of 262k tokens & 2\% of 49k tokens  \\
    \midrule
    Special tokens & \hphantom{00}18 \hphantom{0}(0.6\%) & \hphantom{0}101 \hphantom{0}(2.0\%) & \hphantom{00}0 \hphantom{0}(0.0\%) \\
    Digits and numbers & \hphantom{000}2 \hphantom{0}(0.1\%) & \hphantom{00}77 \hphantom{0}(1.5\%) & \hphantom{00}0 \hphantom{0}(0.0\%) \\
    Punctuation & \hphantom{0}124 \hphantom{0}(4.1\%) & \hphantom{0}521 (10.2\%) & 125 (12.2\%) \\
    Full or partial variable or function names & \hphantom{000}4 \hphantom{0}(0.1\%) & \hphantom{00}64 \hphantom{0}(1.2\%) & 451 (44.0\%) \\
    ALL--CAPS Latin words & \hphantom{000}0 \hphantom{0}(0.0\%) & \hphantom{00}60 \hphantom{0}(1.2\%) & \hphantom{0}88 \hphantom{0}(8.6\%) \\
    Other Latin words & \hphantom{00}66 \hphantom{0}(2.2\%) & \hphantom{0}484 \hphantom{0}(9.5\%) & 253 (24.7\%) \\
    Non-Latin words or characters & 2819 (92.9\%) & 3120 (60.9\%) & \hphantom{0}27 \hphantom{0}(2.6\%) \\
    \midrule
    Other & \hphantom{000}0 \hphantom{0}(0.0\%) & \hphantom{0}693 (13.5\%) & \hphantom{0}80 \hphantom{0}(7.8\%) \\
    \bottomrule
    \end{tabular}
    \caption{Under-trained token classification for coding language models. For each under-trained token category, we present the number of tokens and their percentage out of the total studied token set (2\% of each vocabulary).}
    \label{tab:analysis}
\end{table*}

\section{Related Work}
\label{sec:related_work}

\paragraph{Tokenization and Regularization.} Typical alternatives to BPE include WordPiece and UnigramLM. WordPiece~\cite{schuster2012japanese,Wu2016GooglesNM} is an algorithm for vocabulary learning that iteratively chooses the merges that maximize the overall sequence likelihood the most, which can be reformulated as using the pointwise mutual information (PMI) metric~\citep{church-hanks-1990-word} as an objective. UnigramLM tokenizer~\cite{kudo-2018-subword}, employs a top-down approach, where the large heuristically constructed initial vocabulary is iteratively trimmed by removing tokens that can be split without compromising the sequence probability according to a unigram model. The term ``subword regularization''~\citep{kudo-2018-subword,provilkov-etal-2020-bpe} typically refers to utilizing different tokenization trajectories for the same sequence during language model training and does not imply changes in the BPE vocabulary.

\paragraph{Under-Trained Tokens.}

Under-trained tokens are primarily a result of overfitting to the tokenizer's training sample, but they can also originate from intermediate tokens that are a natural part of the BPE algorithm and are necessary for inference.
Multiple efforts were made to remove junk tokens from language models' vocabularies. Post-training trimming~\citep{yang-etal-2022-textpruner,cognetta-etal-2024-analysis} focuses on editing the trained BPE tokenizer through removing tokens that are chosen for pruning by their frequency. This method reduces the model size, but requires a task-specific choice of absolute thresholds.
\citet{chizhov-etal-2024-bpe} and \citet{10.1609/aaai.v39i23.34633} implemented vocabulary refinement during tokenizer training, mostly targeting intermediate tokens. Both approaches require an inference procedure different from the basic BPE.

\paragraph{BPE Modifications.}
S-BPE~\citep{vilar-federico-2021-statistical}, similarly to WordPiece, aims at maximizing the overall likelihood of the text. S-BPE includes the PMI metric in the objective, accounts for the low-frequency false positives, and adds a stopping criterion based on likelihood decay. Parity-aware BPE~\cite{foroutan2025parityawarebytepairencodingimproving}, mitigates the issue of language disparity and unfairness on LLM inference~\cite{ahia-etal-2023-languages}, by alternating the focus on different languages during tokenizer training, giving favor to less compressed languages. Other efforts focused on pre-tokenization, which has a large impact on the tokenization that follows. For example, numbers that are suggested to be split in groups of one, three left-to-right, or three right-to-left tokens, with the latter having superior performance in the model's numeracy~\cite{lee2024digitstodecisions}. Recent work has focused on superword merges, i.e., merges that occur across whitespaces, commonly used as pre-tokenization barriers. Such merges allow for better text compression and can be useful, especially in coding models, where many instructions are repeated, such as \texttt{``import numpy as np''}~\cite{fried2023incoder}. On the other hand, superword tokens might exacerbate overfitting; we investigate this separately in Section~\ref{sec:alternatives}.
\citet{schmidt2025boundlessbytepairencoding} and \citet{liu2025superbpespacetravellanguage} concurrently studied the impact of superword merges, introducing them either when pre-tokens are already merged or at a specific stage during BPE training. \citet{schmidt2025boundlessbytepairencoding} also highlighted that coding variable names are essentially superword tokens, since they consist of multiple words.

\section{Analysis}
\label{sec:analysis}

We study three common coding language models: CodeGemma 7B~\cite{codegemmateam2024codegemmaopencodemodels}, Qwen 2.5 Coder 3B~\cite{hui2024qwen25codertechnicalreport}, and StarCoder2 3B~\cite{lozhkov2024starcoder2stackv2} and analyze their under-trained tokens. We chose these because it is straightforward to distinguish their known under-trained tokens: CodeGemma has reserved \texttt{<unused>} tokens, Qwen 2.5 Coder has unused embedding vectors above known vocabulary size, and StarCoder2 has a documented set of unused tokens~\cite{land-bartolo-2024-fishing}. For each model, we perform an under-trained token analysis using the \texttt{magikarp} library. Using the undertrained token indicators as in Figure~\ref{fig:starcoder-magikarp}, we compute the distance from $(0, 0)$ for each token. Then, we take the lowest 2\% by this distance for the analysis. After a manual inspection, we highlight several under-trained token categories. We implement a rule-based classification procedure that we describe in Appendix~\ref{app:classification} and present the results in Table~\ref{tab:analysis}.

CodeGemma and Qwen have larger vocabularies, which allows for greater natural language diversity; thus, most of their under-trained tokens are non-Latin words or characters. StarCoder2, in its turn, represents a more condensed tokenizer, where most of the under-trained tokens come from variable names and their parts (44\%). Together with punctuation and uppercase Latin tokens, which are typically present in code, they constitute 64.8\% of the studied tokens. We argue that such tokens can be efficiently prevented by reducing overfitting to large repositories. Furthermore, since the BPE tokens are derived from frequency maximization, it is highly probable that other categories also originate from single repositories (e.g., project-specific words) or from single corrupted documents. Even the intermediate tokens, the removal of which normally requires changing the tokenization procedure~\cite{chizhov-etal-2024-bpe,10.1609/aaai.v39i23.34633}, can be reduced, since they are often building parts of largely overfitted tokens, as seen in Figure~\ref{fig:starcoder-tree}.

\section{Methodology}
\label{sec:methods}

We consider BPE as a greedy optimization algorithm for building a vocabulary $\mathcal{V}$ given a text corpus $\mathcal{C}$. It iteratively collects a list of merges $\mathcal{M}$, enlarging the vocabulary until it reaches the desired size $v$. For all our experiments, we use the byte-level version of BPE~\cite{Radford2019LanguageMA} that initializes the vocabulary with UTF-8 bytes and does not use special \texttt{<UNK>} tokens.

\subsection{BPE Modifications}

During BPE training, we track two properties of each pair: $\mathcal{R}$ --- the number of repositories where the pair is present, and $\mathcal{L}$ --- the number of languages where the pair is present. Using these numbers, we experiment with two BPE modifications that are highlighted in blue in Algorithm~\ref{alg:bpe}: 

\paragraph{Merge skip criterion.} We add an option to skip the merges that do not comply with a chosen criterion. For these criteria, we use thresholds $\mathcal{R}_t$ and $\mathcal{L}_t$, meaning that a merge should be present at least in $\mathcal{R}_t$ repositories and $\mathcal{L}_t$ languages to be executed. The intuition behind this approach relies on the fact that these counts do not increase over the merge tree: $\mathcal{R}\left(t\right) \le \mathcal{R}\left(t_l\right)$ and $\mathcal{R}\left(t\right) \le \mathcal{R}\left(t_r\right)$, where $t = t_l + t_r$; the same is true for the number of languages $\mathcal{L}$. Thus, if a merge is present in $k$ repositories, any merge that the resulting token can form in the future is present in at most $k$ repositories; if $k$ is low, this branch can be pruned. In a sense, increasing the thresholds for merge skipping is a form of adjusting the strength of regularization.

\paragraph{Priority criterion.} In the original BPE, pair frequency $\mathcal{F}$ is set as the criterion for choosing the next merge during training. We experiment with adding other components to this objective:

\begin{itemize}
    \item $\mathcal{F} \cdot \mathcal{L}$, prioritizing language-universal merges. 
    \item $\mathcal{F} \cdot \log\left(\mathcal{R} + 1\right)$, lowering the priority of pairs present in fewer repositories.
    \item $\mathcal{F} \cdot \log\mathcal{R}$, this variant also nullifies all pairs that are present only in one repository.
    \item $\mathcal{F} \cdot \log\left(\mathcal{R} + 1\right) \cdot \mathcal{L}$ and $\mathcal{F} \cdot \log\mathcal{R} \cdot \mathcal{L}$, combining the two properties.
\end{itemize}

\begin{algorithm}[t]
\caption{BPE Training}
\label{alg:bpe}
\begin{algorithmic}[1]
\Require Corpus $\mathcal{C}$; desired vocabulary size $v$
\Ensure Vocabulary $\mathcal{V}$, merge list $\mathcal{M}$

\State Initialize vocabulary $\mathcal{V}$ from $\mathcal{C}$
\State $\mathcal{M} \gets [\,]$

\While{$|\mathcal{V}| < v$}
    \State $(t_{\text{left}}, t_{\text{right}}) \gets \textcolor{blue}{\textsc{PriorityCriterion}}\left(\mathcal{C}\right)$ 
    % \Comment{\textcolor{blue}{Select the top-priority pair}}

    \If{$\textcolor{blue}{\textsc{SkipCriterion}}\left(t_{\text{left}}, t_{\text{right}}, \mathcal{C}\right)$}
        \State \textbf{continue} 
        % \Comment{\textcolor{blue}{Skip an undesirable pair}}
    \EndIf

    \State $t_{\text{new}} \gets t_{\text{left}} + t_{\text{right}}$ 
    \State $\mathcal{V} \gets \mathcal{V} \cup \{t_{\text{new}}\}$
    \State $\mathcal{M} \gets \mathcal{M} \mathbin{\|} \left[\left(t_{\text{left}}, t_{\text{right}}\right)\right]$

    \State Update corpus $\mathcal{C}$ to reflect the merge
\EndWhile

\State \Return $\mathcal{V}, \mathcal{M}$
\end{algorithmic}
\end{algorithm}

\begin{figure*}[t!]
    \begin{subfigure}[b]{\textwidth}
        \includegraphics[width=\linewidth]{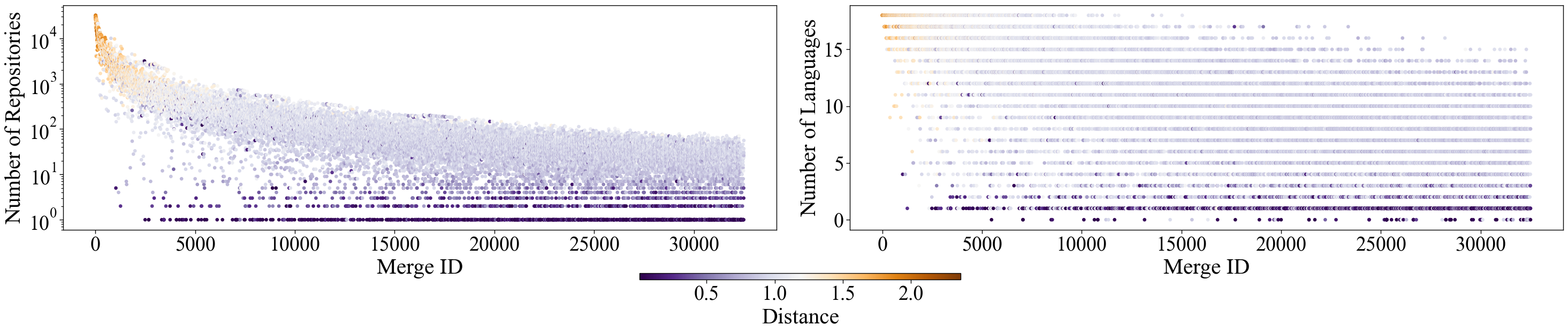}
        \subcaption{BPE merge history. For each merge, we calculate the number of repositories \textbf{(left)} and languages \textbf{(right)} that contain this merge. Color represents distance from $(0, 0)$ in the under-trained token identifiers space (lower = more similar to under-trained).}
        \label{fig:bpe_history}
    \end{subfigure}
    \begin{subfigure}[b]{0.31\textwidth}
        \includegraphics[width=\textwidth]{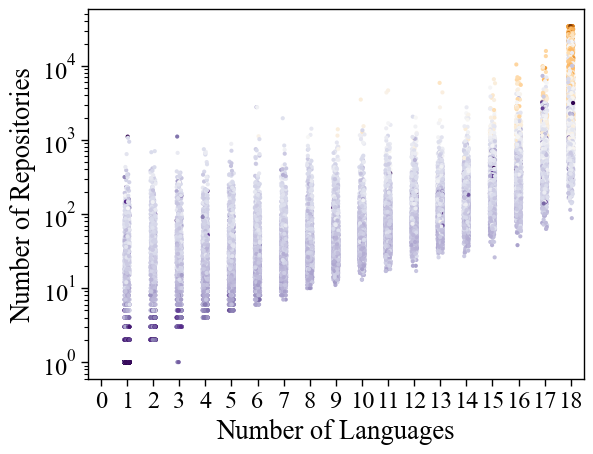}
        \subcaption{Number of languages vs number of repositories for each merge.}
        \label{fig:languages-vs-repos}
    \end{subfigure}
    \hfill
    \begin{subfigure}[b]{0.31\textwidth}
        \includegraphics[width=\textwidth]{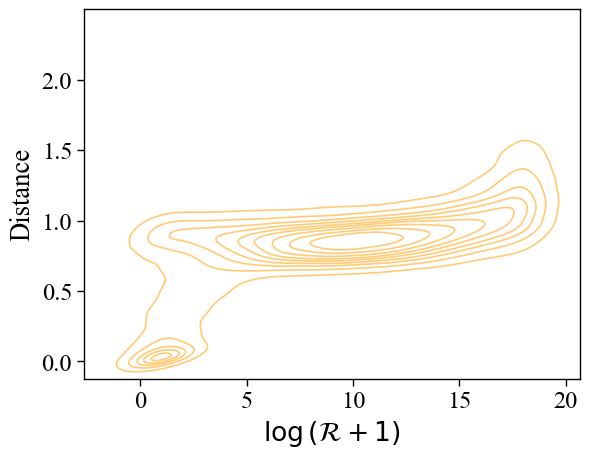}
        \subcaption{Under-trained token indicator distances vs the number of repositories.}
        \label{fig:criterion-log-rep}
    \end{subfigure}
    \hfill
    \begin{subfigure}[b]{0.31\textwidth}
        \includegraphics[width=\textwidth]{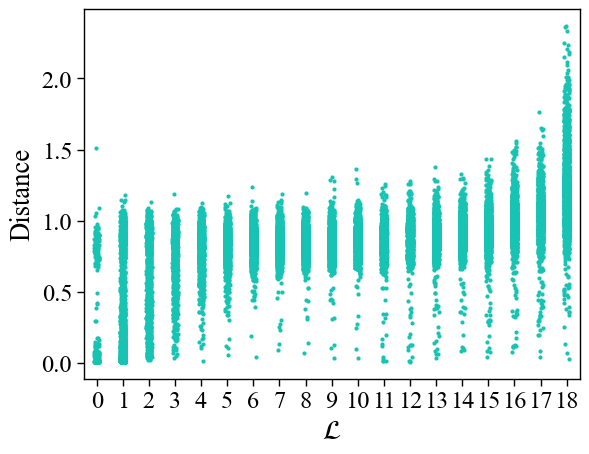}
        \subcaption{Under-trained token indicator distances vs the number of languages.}
        \label{fig:criterion-log-lang}
    \end{subfigure}
    \caption{BPE merge visualization. We show \textbf{(a)} BPE merge sequences with counts of repositories and languages, and \textbf{(b)} correspondence between the two counts, highlighting the distances from $(0, 0)$ in the under-trained token indicator space. We also show correlations between these distances and \textbf{(c)} repository and \textbf{(d)} language counts.}
    \label{fig:bpe_analysis}
\end{figure*}

We do not rule out frequencies completely from the criteria, since they are the main component for optimizing text compression, and there would be multiple candidate pairs with equal criterion values without the frequency component. We also experiment with combining merge skip criteria with priority criteria. For example, we add skipping merges where $\mathcal{L} = 1$ to the SA-BPE with the objective $\mathcal{F} \cdot \log\mathcal{R} \cdot \mathcal{L}$, so that all the single-language and single-repository merges are avoided.

\paragraph{Rationale.} We propose these modifications to introduce regularization directly into BPE training to prevent overfitting. The key idea behind SA-BPE originates from TF-IDF, which involves dividing by the logarithm of the number of documents to prioritize document-specific terms. Here, however, we want to \textbf{de-prioritize} them; therefore, $\log\mathcal{R}$ appears in the numerator. In the code domain, repositories and languages typically have similar structure and syntax, respectively, which makes it natural to use them as units of data.

\subsection{Data}

We assemble the training dataset using Common Corpus~\cite{langlais2025commoncorpuslargestcollection}. We select a random subset of 7\% of documents with a fixed set of 18 programming languages (see full list in Appendix~\ref{app:data}). The collected sample encompasses 14,428,162 documents and 56.22 GB of text; the per-language statistics are shown in Table~\ref{tab:language_stats}. For tokenizer training, we select a random subsample of 0.25\% of the collected data, comprising 36,052 files. We separately conduct single-language experiments using Java and Python subsets of the assembled data, as high-resource languages that differ in syntax and naming conventions.

For the evaluation set, we collected a sample from GitHub repositories that have at least 10 stars and 15 commits, and were created after March 1, 2025, which is later than the cutoff date for Common Corpus, last updated on February 11, 2025 (at the time of this work). For each language, we selected 200 repositories with the most stars; if there were fewer repositories, we included all of them. From each repository, we chose at most 20 files with fixed extensions (see Appendix~\ref{app:data} for details).

\subsection{Model Training}

We train a series of small Llama models~\citep{grattafiori2024llama3herdmodels} with 100 million parameters (see Appendix~\ref{app:architecture} for details). Each model is trained for 60,000 steps on two NVIDIA A100 GPUs with a context window size of 2,048, a batch size per GPU of 16, and four steps of gradient accumulation, resulting in an effective batch size of 262,144 tokens and a total training set size of 15.7 billion tokens. We reserve three unused tokens in each model to analyze the model embeddings with the \texttt{magikarp} package~\cite{land-bartolo-2024-fishing}: \texttt{<|unused\_token\_\{1,2,3\}|>}. These tokens never appear in the training set; hence, their embeddings do not change during training and can be used as seed vectors for under-trained token analysis.

\subsection{Evaluation}

We use intrinsic tokenizer quality measures, namely compression rate (the corpus length in bytes divided by its token count), coverage (the number of tokens used at least once in the evaluation sample), and mean token length. We consider tokenizers with better compression and higher coverage on the evaluation set to be more regularized, as they are better suited to unseen content. If a tokenizer with a better compression rate/higher coverage has a lower mean token length, this signals better generalization even further, since shorter types are more likely to be reused, and the long, unneeded tokens take up less space in the vocabulary.

\paragraph{Magikarp analysis.} We use the \texttt{magikarp} package~\citep{land-bartolo-2024-fishing} as the main extrinsic measure to study the impact of a tokenizer on a trained model. Namely, we compute the under-trained token indicators (Euclidean and cosine distances from the known under-trained tokens) and run the verification prompts. A verification is run on the bottom 2\% of the under-trained token indicator distribution, presenting the model with each token and asking it to repeat it. If the next-token probability for the token itself is below 1\%, the token is verified as under-trained.

\begin{table}[tbp]\centering
\small
\begin{tabular}{lrrrrr}\toprule
\textbf{Tokenizer}&\textbf{CR} &\textbf{Used} &\textbf{MTL} &\textbf{Steps} \\\midrule
BPE &3.90 &19585 &\cellcolor[HTML]{f8d8d5}6.75 &32508 \\
Skip $\mathcal{R} < 2$ &\cellcolor[HTML]{ebf7f2}3.90 &\cellcolor[HTML]{eef8f4}19862 &\cellcolor[HTML]{fdf4f3}6.57 &\cellcolor[HTML]{fef8f7}36263 \\
Skip $\mathcal{R} < 5$ &\cellcolor[HTML]{dcf1e7}3.91 &\cellcolor[HTML]{def2e8}20117 &\cellcolor[HTML]{fffdfd}6.51 &\cellcolor[HTML]{fdf0ee}40096 \\
Skip $\mathcal{R} < 10$ &\cellcolor[HTML]{ceebdd}3.91 &\cellcolor[HTML]{d0ecdf}20343 &6.50 &\cellcolor[HTML]{fbe8e6}43575 \\
Skip $\mathcal{R} < 20$ &\cellcolor[HTML]{b7e1cd}3.92 &\cellcolor[HTML]{b7e1cd}20741 &6.50 &\cellcolor[HTML]{f8d8d5}51093 \\
$\mathcal{F} \cdot \log\left(\mathcal{R} + 1\right)$ &\cellcolor[HTML]{c7e8d8}3.91 &\cellcolor[HTML]{c7e8d8}20491 &\cellcolor[HTML]{fef9f9}6.54 &32508 \\
$\mathcal{F} \cdot \log\mathcal{R}$ &\cellcolor[HTML]{c4e7d6}3.91 &\cellcolor[HTML]{c3e6d6}20552 &\cellcolor[HTML]{fffdfc}6.51 &32508 \\
\midrule
BPE &3.68 &17208 &\cellcolor[HTML]{f8d8d5}5.84 &32508 \\
Skip $\mathcal{R} < 2$ &\cellcolor[HTML]{eff9f4}3.68 &\cellcolor[HTML]{f0f9f5}17363 &5.78 &\cellcolor[HTML]{fffdfd}34186 \\
Skip $\mathcal{R} < 5$ &\cellcolor[HTML]{dcf1e7}3.69 &\cellcolor[HTML]{e3f4ec}17493 &5.78 &\cellcolor[HTML]{fefaf9}36201 \\
Skip $\mathcal{R} < 10$ &\cellcolor[HTML]{cceadc}3.69 &\cellcolor[HTML]{d3ede1}17651 &\cellcolor[HTML]{fef8f7}5.79 &\cellcolor[HTML]{fdf4f3}39745 \\
Skip $\mathcal{R} < 20$ &\cellcolor[HTML]{b7e1cd}3.70 &\cellcolor[HTML]{b7e1cd}17932 &\cellcolor[HTML]{f9d9d6}5.84 &\cellcolor[HTML]{f8d8d5}57570 \\
$\mathcal{F} \cdot \log\left(\mathcal{R} + 1\right)$ &\cellcolor[HTML]{d4ede1}3.69 &\cellcolor[HTML]{d2ece0}17670 &\cellcolor[HTML]{fdf0ee}5.80 &32508 \\
$\mathcal{F} \cdot \log\mathcal{R}$ &\cellcolor[HTML]{d2ece0}3.69 &\cellcolor[HTML]{ceebdd}17708 &\cellcolor[HTML]{fdf0ef}5.80 &32508 \\
\bottomrule
\end{tabular}
\caption{Evaluation of Java (top) and Python (bottom) SA-BPE tokenizers compared to BPE: compression rate (CR) and number of used tokens in the evaluation set, mean token length (MTL), and number of training steps.}\label{tab:java-python}
\end{table}

\begin{figure}[t]
    \begin{subfigure}[b]{\linewidth}
        \includegraphics[width=\textwidth]{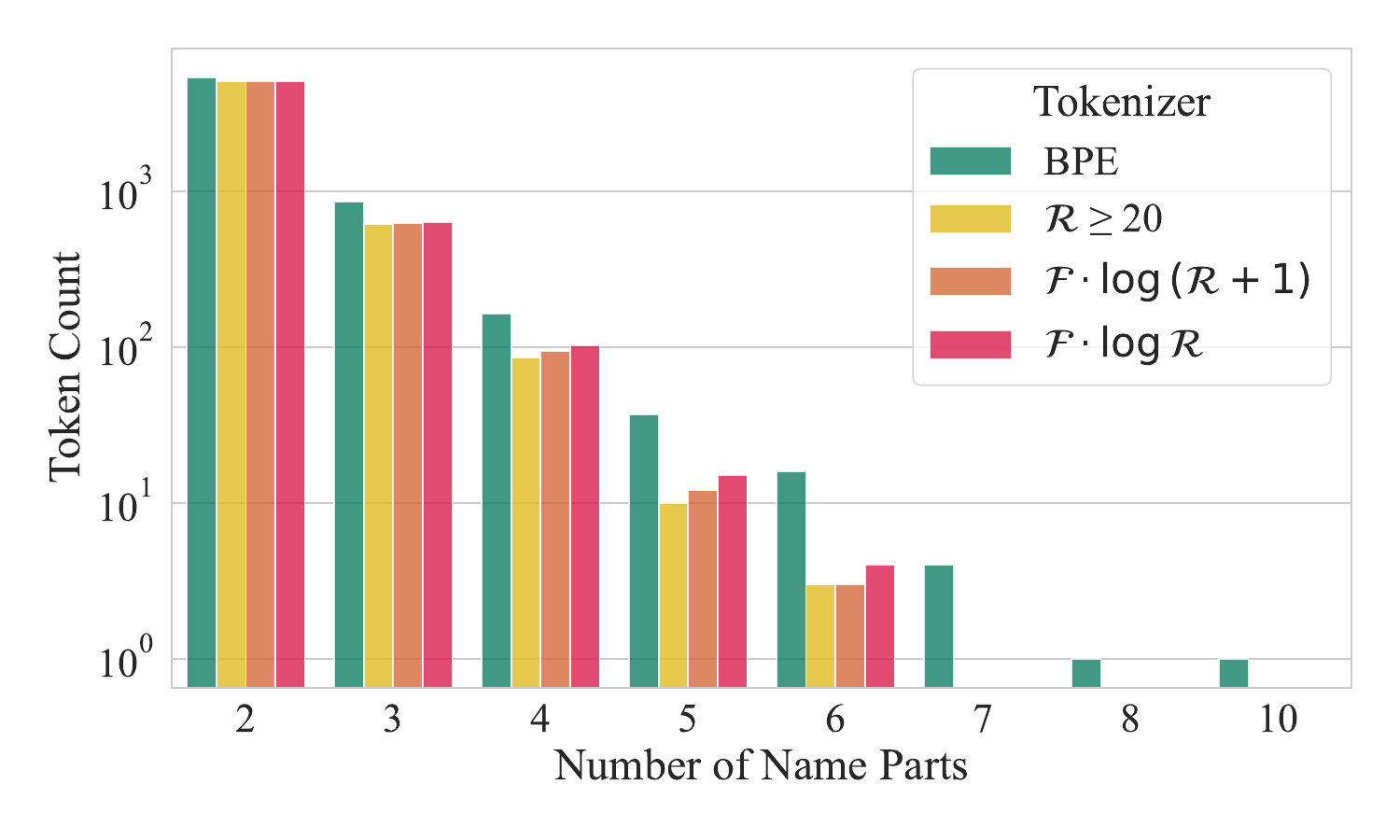}
        \subcaption{Variable name token counts in Java tokenizers.}
        \label{fig:java-cases}
    \end{subfigure}
    \hfill
    \begin{subfigure}[b]{\linewidth}
        \includegraphics[width=\textwidth]{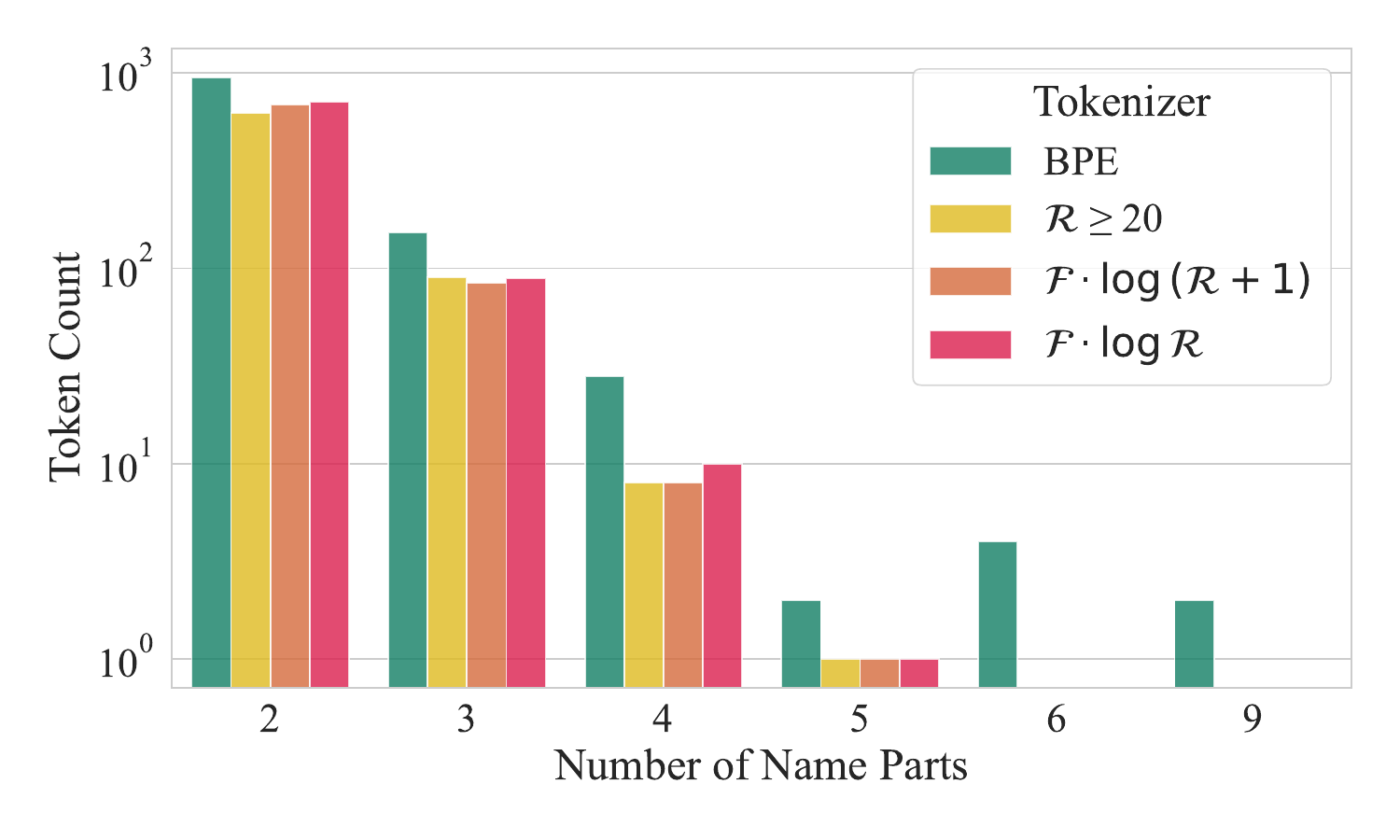}
        \subcaption{Variable name token counts in Python tokenizers.}
        \label{fig:python-cases}
    \end{subfigure}
    \hfill
    \begin{subfigure}[b]{\linewidth}
        \includegraphics[width=\textwidth]{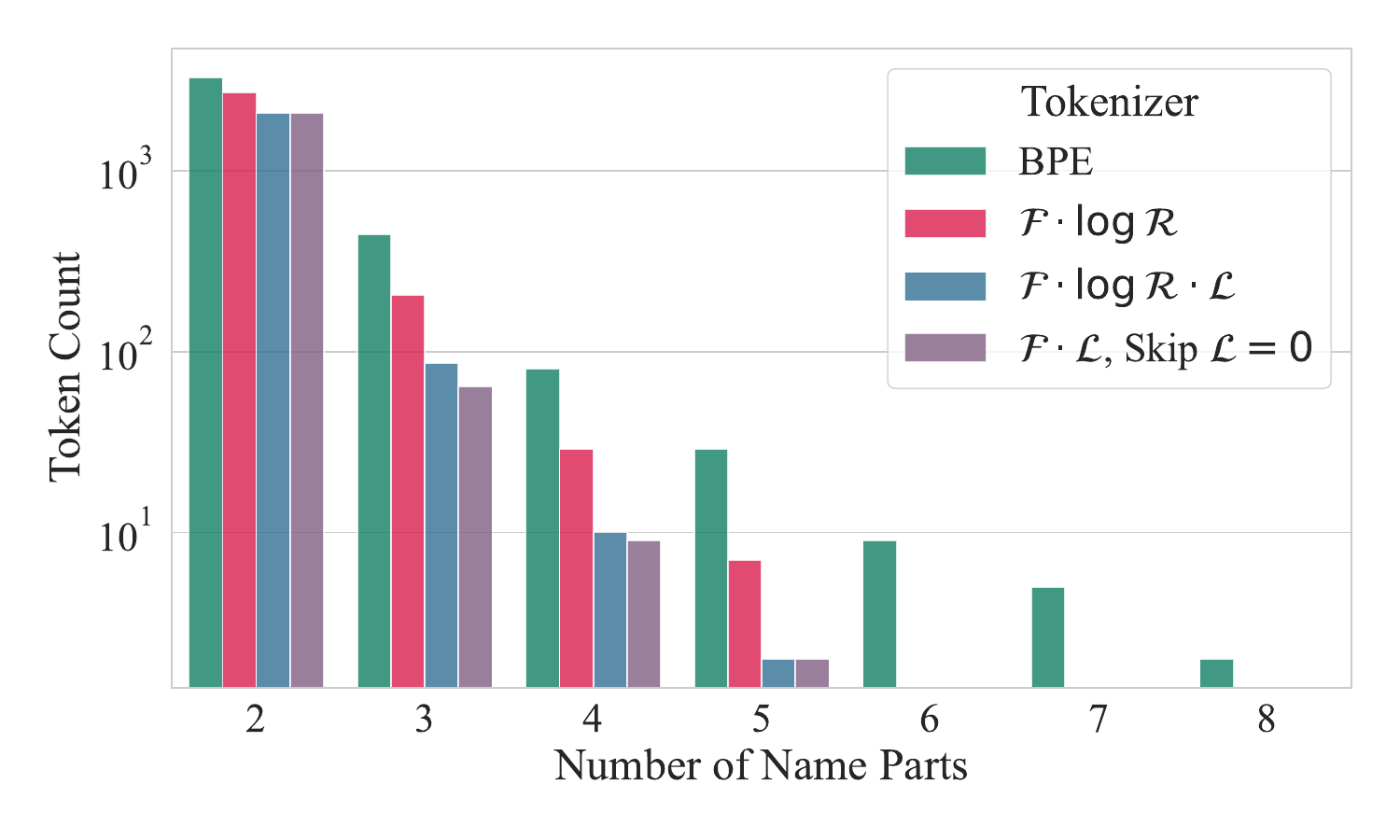}
        \subcaption{Variable name token counts in multilingual tokenizers.}
        \label{fig:all-names}
    \end{subfigure}
    \caption{\texttt{CamelCase} and \texttt{snake\_case} token lengths (in name parts) in basic BPE and our modifications.}
    \label{fig:cases}
\end{figure}

\section{Experimental Results}
\label{sec:results}

Before implementing our modifications, we train a basic BPE model on the selected tokenizer corpus and count the number of repositories and languages that each merge appears in. We also train a generative model using this tokenizer and compute under-trained token indicators. We show the BPE merge history in Figure~\ref{fig:bpe_history}. The number of repositories and languages the merge appears in decreases along the merge history. Furthermore, a certain number of merges occur only in a few repositories and languages, well below the general trend. These merges mostly comprise identifiable under-trained tokens. This justifies prioritizing merges that appear in a larger number of repositories and languages.

It is also notable that under-trained tokens, positioned low in the plots, appear as early as within the first 5\,000 merges. These tokens are, for example: \texttt{FHI}, \texttt{FHIR}, \texttt{PHPFHIR}, \texttt{HPFHIRConstants} --- tokens appearing in only one PHP project named FHIR, and \texttt{chord}, \texttt{chordie} --- tokens from a repository called \texttt{guitartools}. These are clear cases of overfitting to repository-specific naming and content. There are also some under-trained tokens positioned high in the plots. These are mostly intermediate word parts: \texttt{perty}, \texttt{verride}, \texttt{urrent}. These findings correspond well to our hypotheses formulated in Section~\ref{sec:analysis}. As we show in Figure~\ref{fig:languages-vs-repos}, there is a correlation between the number of repositories and the number of languages. Furthermore, Figures~\ref{fig:criterion-log-rep}~and~\ref{fig:criterion-log-lang} show that the number of repositories is superior to the number of languages in indicating tokens that are clearly under-trained.

\begin{table*}[!htp]\centering
\small
\begin{tabular}{lcccccc}\toprule
\textbf{Priority Criterion} &\textbf{Merge Skip} &\textbf{Compression} &\textbf{Coverage} &\textbf{MTL} &\textbf{\# 3-digit numbers} &\textbf{\# Under-trained} \\\midrule
$\mathcal{F}$ (BPE) &--- (BPE) &3.623 &29240 &\cellcolor[HTML]{f8d8d5}6.23 &517 &640 \\
\midrule
$\mathcal{F}$ &Skip $\mathcal{L} < 4$ &3.623 &\cellcolor[HTML]{bde4d1}31416 &\cellcolor[HTML]{fffafa}5.89 &\cellcolor[HTML]{ddf1e8}655 &\cellcolor[HTML]{b7e1cd}52 \\
$\mathcal{F} \cdot \mathcal{L}$ &--- &\cellcolor[HTML]{cdeadc}3.637 &\cellcolor[HTML]{c1e6d4}31267 &\cellcolor[HTML]{fffefe}5.85 &\cellcolor[HTML]{c0e5d4}772 &\cellcolor[HTML]{c3e6d5}151 \\
$\mathcal{F} \cdot \mathcal{L}$ &Skip $\mathcal{L} = 1$ &\cellcolor[HTML]{d4eee2}3.635 &\cellcolor[HTML]{bee4d2}31357 &5.84 &\cellcolor[HTML]{bfe4d2}779 &\cellcolor[HTML]{bae2cf}75 \\
$\mathcal{F} \cdot \log\left(\mathcal{R} + 1\right)$ &--- &\cellcolor[HTML]{bfe4d2}3.641 &\cellcolor[HTML]{cdebdd}30876 &\cellcolor[HTML]{fcedec}6.02 &\cellcolor[HTML]{ecf7f2}595 &\cellcolor[HTML]{cae8da}202 \\
$\mathcal{F} \cdot \log\mathcal{R}$ &--- &\cellcolor[HTML]{b7e1cd}3.643 &\cellcolor[HTML]{c7e8d8}31066 &\cellcolor[HTML]{fdf0ef}5.99 &\cellcolor[HTML]{e9f6f0}607 &\cellcolor[HTML]{bae2cf}72 \\
\midrule
$\mathcal{F} \cdot \log\left(\mathcal{R} + 1\right)$ &Skip $\mathcal{L} < 4$ &\cellcolor[HTML]{edf8f3}3.628 &\cellcolor[HTML]{c4e7d6}31187 &\cellcolor[HTML]{fefaf9}5.89 &\cellcolor[HTML]{e9f6f0}608 &\cellcolor[HTML]{b9e1ce}62 \\
$\mathcal{F} \cdot \log\left(\mathcal{R} + 1\right) \cdot \mathcal{L}$ &--- &\cellcolor[HTML]{bbe3d0}3.642 &\cellcolor[HTML]{b9e2cf}31533 &\cellcolor[HTML]{fffdfd}5.86 &\cellcolor[HTML]{b9e2cf}800 &\cellcolor[HTML]{b9e1ce}63 \\
$\mathcal{F} \cdot \log\mathcal{R} \cdot \mathcal{L}$ &--- &\cellcolor[HTML]{bbe3d0}3.642 &\cellcolor[HTML]{b8e2ce}31567 &\cellcolor[HTML]{fffdfd}5.86 &\cellcolor[HTML]{b8e2ce}804 &\cellcolor[HTML]{b7e1cd}44 \\
$\mathcal{F} \cdot \log\mathcal{R} \cdot \mathcal{L}$ &Skip $\mathcal{L} = 1$ &\cellcolor[HTML]{c6e8d8}3.639 &\cellcolor[HTML]{b7e1cd}31581 &\cellcolor[HTML]{fffefd}5.86 &\cellcolor[HTML]{b7e1cd}808 &\cellcolor[HTML]{b7e1cd}45 \\
\bottomrule
\end{tabular}
\caption{Evaluation of multilingual SA-BPE tokenizers compared to BPE: compression rate and number of total used tokens in the evaluation set, mean token length (MTL), three-digit numbers count, and number of verified under-trained tokens out of 646 tested. The first row represents the basic BPE algorithm.}\label{tab:multilingual-tokens}
\end{table*}

\begin{figure*}[t!]
    \begin{subfigure}[b]{0.49\textwidth}
        \includegraphics[width=\textwidth]{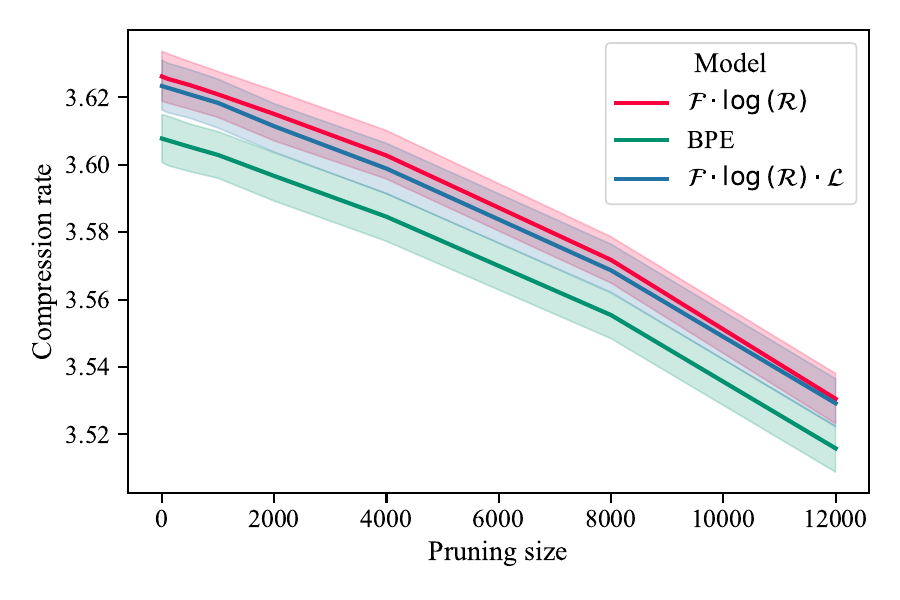}
        \subcaption{Pruning order: reverse merge order.}
        \label{fig:naive_all}
    \end{subfigure}
    \hfill
    \begin{subfigure}[b]{0.49\textwidth}
        \includegraphics[width=\textwidth]{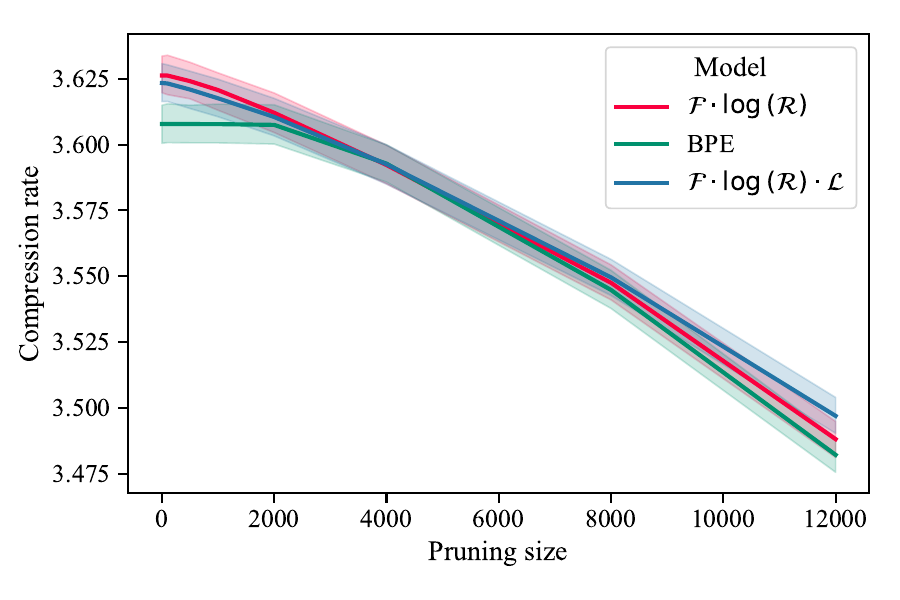}
        \subcaption{Pruning order: under-trained first.}
        \label{fig:scores_all}
    \end{subfigure}
    \caption{Compression rate for tokenizers with applied pruning \textbf{(a)} in the reverse order of token ids and \textbf{(b)} starting from the tokens with the lowest under-trained indicator values, here distance from $(0, 0)$ in indicator space.}
    \label{fig:pruning_all}
\end{figure*}

\subsection{Monolingual Experiments}

We train a series of tokenizers on Java and Python samples, using the minimum number of repositories as merge skip criteria and frequencies multiplied by the logarithm of the number of repositories (with and without nullifying single-repository merges) as priority criteria. As we show in Table~\ref{tab:java-python}, SA-BPE modifications slightly improve compression rate and token coverage in unseen repositories, both of which tend to increase along with the strength of regularization (increasing the threshold for $\mathcal{R}$). Compared to BPE, our tokenizers have a lower mean token length, which might serve as a proxy measure for overfitting to longer repository-specific names and, together with the non-compromised compression rate, suggests better generalization. While increasing the threshold for merge skips improves the tokenizer's generalizability, it comes with longer training, requiring more steps due to skips. Furthermore, increasing the threshold to 30 already resulted in running out of merges for the vocabulary of size 32\,768. 

In addition, we investigate the presence of \texttt{CamelCase} and \texttt{snake\_case} tokens in tokenizers. For each such token, we compute its length in name parts, splitting the \texttt{snake\_case} by underscores and \texttt{CamelCase} by capital letters (except for the abbreviations, e.g., \texttt{getHTTPStatus} has three parts) and show the distributions in Figures~\ref{fig:java-cases}~and~\ref{fig:python-cases}. Compared to BPE, our tokenizers have fewer long variable names. In all evaluations, priority criteria based on the logarithm of repository count showed comparable stable performance, without requiring a threshold value. Out of the two priority criterion versions, $\mathcal{F} \cdot \log\mathcal{R}$ shows slightly better scores as it avoids single-repository tokens.

\begin{figure*}[t!]
    \begin{subfigure}[b]{0.48\textwidth}
        \includegraphics[width=\textwidth]{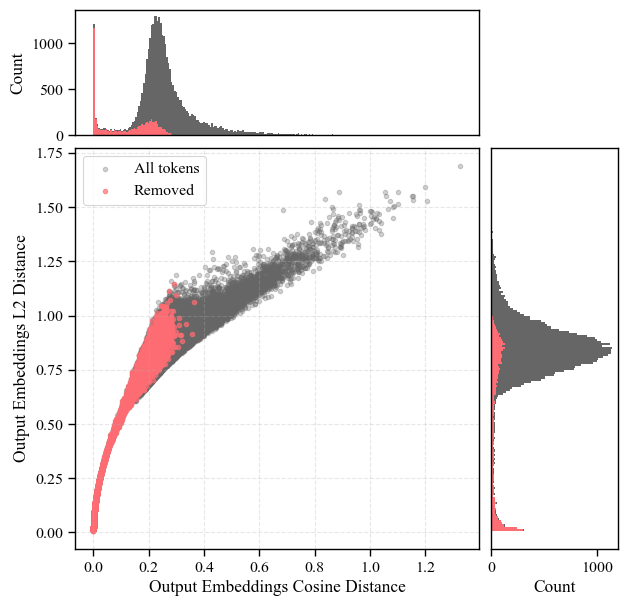}
        \subcaption{Under-trained token indicators for the model trained with basic BPE. The tokens present in this model but not in the $\mathcal{F} \cdot \log\mathcal{R} \cdot \mathcal{L}$ model are highlighted in color.}
        \label{fig:main-removed}
    \end{subfigure}
    \hfill
    \begin{subfigure}[b]{0.48\textwidth}
        \includegraphics[width=\textwidth]{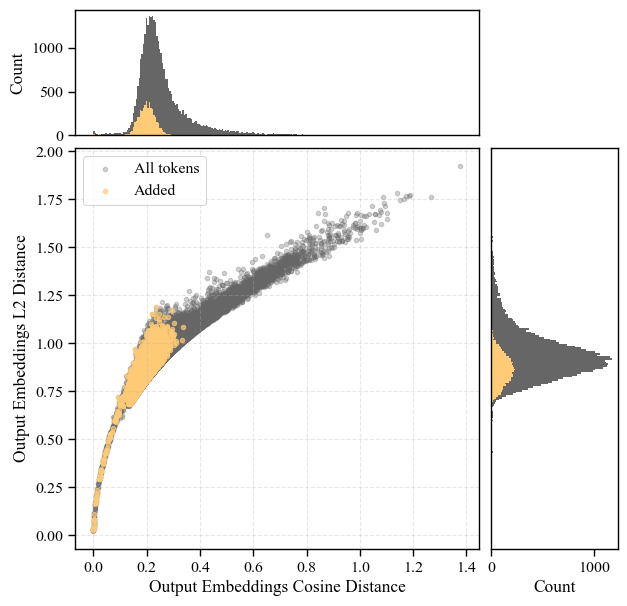}
        \subcaption{Under-trained token indicators for the $\mathcal{F} \cdot \log\mathcal{R} \cdot \mathcal{L}$ model. The tokens present in this model but not in the model trained with basic BPE are highlighted in color.}
        \label{fig:main-added}
    \end{subfigure}
    \caption{Under-trained token analysis in comparison for the model trained with basic BPE and the model trained with priority criterion $\mathcal{F} \cdot \log\mathcal{R} \cdot \mathcal{L}$. Each plot is accompanied by 1D histograms for better visual comparison.}
    \label{fig:main-magikarp}
\end{figure*}

\subsection{Multilingual Experiments}

\paragraph{Tokenizer metrics.}
In multilingual experiments, we incorporate the language component into the analysis as a multiplier or a threshold, choosing an intermediate value of 4 (we compare different threshold values in Appendix~\ref{sec:app-min_languages}). We present the main tokenizer metrics in Table~\ref{tab:multilingual-tokens}. Most of the tested SA-BPE modifications improve text compression, increase the coverage of the tokenized set, and yield generally shorter tokens than the basic BPE. Per-language scores indicate that modifications incorporating the language component provide stronger support for lower-resource languages (see Appendix~\ref{app:multilingual} for a comparison). Our regularized versions also produce more three-digit numbers as tokens, closer to 900, the total count in the best case. Comparing the lengths of variable names in Figure~\ref{fig:all-names}, we observe a situation similar to the monolingual scenarios: regularized tokenizers have fewer tokens that are long variable names; in particular, no longer than five name parts.

We separately evaluate compression during tokenizer pruning. We use two pruning strategies: starting from the end of the merge list (Figure~\ref{fig:naive_all}) and leaf-based pruning~\cite{purason2025teachingoldtokenizersnew} starting from the most under-trained tokens (Figure~\ref{fig:scores_all}). In the first case, we observe that our modifications result in superior compression across all vocabulary sizes. In the second case, the first 2,000 removed tokens have no influence on BPE's compression rate. This confirms that BPE produces useless tokens and that these tokens are well-correlated with the under-trained token indicators from \texttt{magikarp}. At the same time, our regularized tokenizers begin to lose compression even with as few as 100 removed tokens, highlighting the low redundancy in the vocabulary. We present and analyze the per-language plots in Appendix~\ref{app:pruning}.

In Figure~\ref{fig:freq_density}, we show token probability distributions in the training sample. BPE shows a heavy tail in the lowest-probability zone, which signals that more tokens will receive insufficient training.

\paragraph{Model evaluations.} The critical part of our analysis is the quantity of under-trained tokens. We used the verification prompting procedure with default parameters from \texttt{magikarp} and cosine distance as the indicator metric. For all tokenizers, the total number of tested tokens was 646. All SA-BPE variants reduce the number of verified under-trained tokens, in the best case, down to 44. Repository and language counts are efficient on their own, but combined modifications showed the best performance, especially as parts of the priority criterion.
The difference is also visible on the under-trained token indicator plots (see Figure~\ref{fig:main-magikarp}). The histograms of the basic BPE show large bins near $(0, 0)$, indicating under-trained tokens. In the modified algorithm, these regions are rather negligible. These tokens are unavoidable intermediate parts of frequent words, e.g., \texttt{OOLEAN}, \texttt{scriptors}, \texttt{trieve} --- parts of \texttt{BOOLEAN}, \texttt{descriptors}, \texttt{retrieve} (see examples from all models in Appendix~\ref{app:tokens}). To avoid such tokens, it is possible to use SA-BPE regularization in the algorithms aimed at intermediate tokens, e.g., PickyBPE. Examining the removed and added tokens in the same figure, we observe that SA-BPE excludes the majority of under-trained tokens from BPE and includes tokens closer to the general distribution. The difference in this case exceeds 10,000 tokens (see Appendix~\ref{app:multilingual} for all differences).

\begin{figure}[t]
    \centering
    \includegraphics[width=\linewidth]{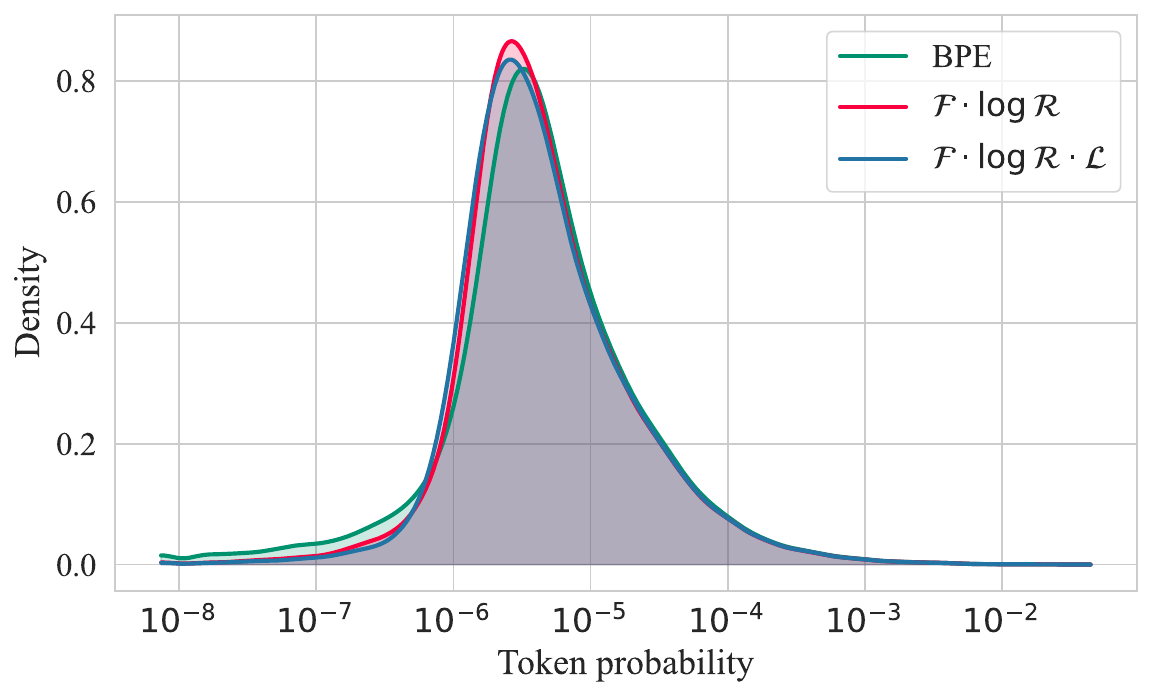}
    \caption{Kernel density estimation of the token probabilities calculated on the model training data.}
    \label{fig:freq_density}
\end{figure}

\subsection{Alternative Tokenization Methods}
\label{sec:alternatives}

We also test the performance of UnigramLM and Wordpiece\footnote{We use the \texttt{tokenizers} library from HuggingFace, where WordPiece is implemented as BPE with a greedy longest-prefix inference.} on code data in Appendix~\ref{app:unigram_wordpiece} and find those to be inferior even to basic BPE. We also compare our regularized BPE tokenizers to common BPE modifications: BoundlessBPE~\cite{schmidt2025boundlessbytepairencoding} and PickyBPE~\cite{chizhov-etal-2024-bpe}. Our best method, SA-BPE with a priority criterion $\mathcal{F} \cdot \log\mathcal{R} \cdot \mathcal{L}$, achieves the best compression rate and token coverage. Along with these scores, our method has a lower mean token length, which, combined with a better compression rate, signals less overfitting. Finally, because the inference procedure remains intact, our method demonstrates superior inference time\footnote{For both alternatives, we use byte-level implementations from \url{https://github.com/kensho-technologies/boundlessbpe}}.

\begin{table}[tbp]\centering
\small
%\resizebox{ extwidth}{!}{ % use this if the table is too large
\begin{tabular}{lcccc}\toprule
\textbf{Tokenizer}&\textbf{CR} &\textbf{Used} &\textbf{MTL} &\textbf{Time} \\\midrule
BPE & 3.62 & 29240 & 6.23 & \textbf{0:12} \\
PickyBPE & 3.61 & 28673 & 6.22 & 27:04$^*$ \\
BoundlessBPE & 3.56 & 29375 & 6.70 & 28:55$^*$ \\
SA-BPE (ours) & \textbf{3.64} & \textbf{31567} & 5.86 & \textbf{0:12} \\
\bottomrule
\end{tabular}
\caption{Evaluation of multilingual tokenizers: compression rate (CR) and number of used tokens in the evaluation set, mean token length (MTL), and total inference time. ``SA-BPE'' denotes our best model trained with priority criterion $\mathcal{F} \cdot \log\mathcal{R} \cdot \mathcal{L}$. $^*$Inference time is implementation-dependent.}\label{tab:alternatives}
\end{table}
\section{Discussion}
\label{sec:discussion}

The demonstrated SA-BPE variants introduce regularization into BPE training and demonstrate superiority in building a generalizable tokenizer for coding languages in both monolingual and multilingual scenarios, compared to both basic BPE and SOTA alternatives. Though results suggest that the repository count is a more informative metric and language count might be its proxy (Figure~\ref{fig:languages-vs-repos}), the models with the fewest under-trained tokens and the best performance include both components. This might be due to supporting the general language-agnostic syntax tokens in the multilingual environment. Also, language-related modifications show better parity and support for lower-resource languages (Appendix~\ref{app:multilingual}). Merge skip criteria are explicit, but require parameter tuning. Therefore, the more obvious cases for them are prohibiting single-language tokens in certain multilingual scenarios and setting explicit lower bounds when the project knowledge allows. Based on these findings, one might choose modifications depending on project needs, while the combined parameter-free priority criterion modifications ($\mathcal{F} \cdot \log\mathcal{R} \cdot \mathcal{L}$ in the multilingual case) are a good default choice.

One of the benefits of SA-BPE is that it brings changes only to the BPE training, while the tokenizer can be used for inference just as the regular BPE (see the discussion of time and space complexity in Appendix~\ref{app:complexity}). The demonstrated approach can also be applied to natural languages, using analogies of data sources such as documents, books, or document collections. However, the language aspect must be avoided in this case, since language diversity in natural domains is substantially larger than in programming languages.

\section{Conclusion}
\label{sec:conclusion}

In this paper, we propose SA-BPE, a set of modifications to the BPE training procedure that enable a substantial reduction in overfitting for code data, without altering the tokenization function and slowing down inference. Our modifications effectively target repository-specific names and variables, as well as the non-code-related content. This improves tokenizer efficiency and reduces the number of under-trained tokens, which are linked with various model issues in the literature. Algorithmically, SA-BPE is lightweight, which makes it possible to combine it with other BPE modifications and apply it to other domains and general natural language, using the analogy of data sources.

\section*{Limitations}
\label{sec:limitations}

In this work, we focus on the code data as the primary domain and utilize its specific characteristics to propose improvements. Even though we mention in Section~\ref{sec:discussion} that the proposed modifications are, in theory, applicable to other domains, we do not conduct such experiments, since they are out of scope of the current work. 

Due to computational resource limitations, the models trained in this work are small and not suitable for more sophisticated benchmarking with downstream code-related tasks. For the same reason, we also did not study how the proposed changes would affect the performance of larger language models with vocabulary transfer and continued pre-training on the code data.

\section*{Acknowledgements}
This work was funded by a joint project of the Center for Artificial Intelligence (CAIRO), THWS and JetBrains Research (project ERIC). The computational resources were provided by Erlangen National High-Performance Computing Center (NHR@FAU) of the Friedrich-Alexander-Universität Erlangen-Nürnberg (FAU) (joint project with CAIRO, THWS). The authors acknowledge the work of Taido Purason, specifically the repository \texttt{tokenizer-extension}, the BPE training implementation from which was adapted for the core implementation of this work.

\bibliography{custom}

@inproceedings{
langlais2025commoncorpuslargestcollection,
title={Common {C}orpus: {The Largest Collection of Ethical Data for {LLM} Pre-Training}},
author={Pierre-Carl Langlais and Pavel Chizhov and Catherine Arnett and Carlos Rosas Hinostroza and Mattia Nee and Eliot Krzysztof Jones and Ir{\`e}ne Girard and David Mach and Anastasia Stasenko and Ivan P. Yamshchikov},
booktitle={The Fourteenth International Conference on Learning Representations},
year={2026},
url={https://openreview.net/forum?id=0wSlFpMsGb}
}

@inproceedings{
schmidt2025boundlessbytepairencoding,
title={Boundless Byte Pair Encoding: Breaking the Pre-tokenization Barrier},
author={Craig W Schmidt and Varshini Reddy and Chris Tanner and Yuval Pinter},
booktitle={Second Conference on Language Modeling},
year={2025},
url={https://openreview.net/forum?id=oPAjXGV8qQ}
}

@misc{gemmateam2025gemma3technicalreport,
      title={{Gemma 3 Technical Report}}, 
      author={Aishwarya Kamath and Johan Ferret and Shreya Pathak and Nino Vieillard and Ramona Merhej and Sarah Perrin and Tatiana Matejovicova and Alexandre Ramé and Morgane Rivière and Louis Rouillard and Thomas Mesnard and Geoffrey Cideron and Jean-bastien Grill and Sabela Ramos and Edouard Yvinec and Michelle Casbon and Etienne Pot and Ivo Penchev and Gaël Liu and Francesco Visin and Kathleen Kenealy and Lucas Beyer and Xiaohai Zhai and Anton Tsitsulin and Robert Busa-Fekete and Alex Feng and Noveen Sachdeva and Benjamin Coleman and Yi Gao and Basil Mustafa and Iain Barr and Emilio Parisotto and David Tian and Matan Eyal and Colin Cherry and Jan-Thorsten Peter and Danila Sinopalnikov and Surya Bhupatiraju and Rishabh Agarwal and Mehran Kazemi and Dan Malkin and Ravin Kumar and David Vilar and Idan Brusilovsky and Jiaming Luo and Andreas Steiner and Abe Friesen and Abhanshu Sharma and Abheesht Sharma and Adi Mayrav Gilady and Adrian Goedeckemeyer and Alaa Saade and Alex Feng and Alexander Kolesnikov and Alexei Bendebury and Alvin Abdagic and Amit Vadi and András György and André Susano Pinto and Anil Das and Ankur Bapna and Antoine Miech and Antoine Yang and Antonia Paterson and Ashish Shenoy and Ayan Chakrabarti and Bilal Piot and Bo Wu and Bobak Shahriari and Bryce Petrini and Charlie Chen and Charline Le Lan and Christopher A. Choquette-Choo and CJ Carey and Cormac Brick and Daniel Deutsch and Danielle Eisenbud and Dee Cattle and Derek Cheng and Dimitris Paparas and Divyashree Shivakumar Sreepathihalli and Doug Reid and Dustin Tran and Dustin Zelle and Eric Noland and Erwin Huizenga and Eugene Kharitonov and Frederick Liu and Gagik Amirkhanyan and Glenn Cameron and Hadi Hashemi and Hanna Klimczak-Plucińska and Harman Singh and Harsh Mehta and Harshal Tushar Lehri and Hussein Hazimeh and Ian Ballantyne and Idan Szpektor and Ivan Nardini and Jean Pouget-Abadie and Jetha Chan and Joe Stanton and John Wieting and Jonathan Lai and Jordi Orbay and Joseph Fernandez and Josh Newlan and Ju-yeong Ji and Jyotinder Singh and Kat Black and Kathy Yu and Kevin Hui and Kiran Vodrahalli and Klaus Greff and Linhai Qiu and Marcella Valentine and Marina Coelho and Marvin Ritter and Matt Hoffman and Matthew Watson and Mayank Chaturvedi and Michael Moynihan and Min Ma and Nabila Babar and Natasha Noy and Nathan Byrd and Nick Roy and Nikola Momchev and Nilay Chauhan and Noveen Sachdeva and Oskar Bunyan and Pankil Botarda and Paul Caron and Paul Kishan Rubenstein and Phil Culliton and Philipp Schmid and Pier Giuseppe Sessa and Pingmei Xu and Piotr Stanczyk and Pouya Tafti and Rakesh Shivanna and Renjie Wu and Renke Pan and Reza Rokni and Rob Willoughby and Rohith Vallu and Ryan Mullins and Sammy Jerome and Sara Smoot and Sertan Girgin and Shariq Iqbal and Shashir Reddy and Shruti Sheth and Siim Põder and Sijal Bhatnagar and Sindhu Raghuram Panyam and Sivan Eiger and Susan Zhang and Tianqi Liu and Trevor Yacovone and Tyler Liechty and Uday Kalra and Utku Evci and Vedant Misra and Vincent Roseberry and Vlad Feinberg and Vlad Kolesnikov and Woohyun Han and Woosuk Kwon and Xi Chen and Yinlam Chow and Yuvein Zhu and Zichuan Wei and Zoltan Egyed and Victor Cotruta and Minh Giang and Phoebe Kirk and Anand Rao and Kat Black and Nabila Babar and Jessica Lo and Erica Moreira and Luiz Gustavo Martins and Omar Sanseviero and Lucas Gonzalez and Zach Gleicher and Tris Warkentin and Vahab Mirrokni and Evan Senter and Eli Collins and Joelle Barral and Zoubin Ghahramani and Raia Hadsell and Yossi Matias and D. Sculley and Slav Petrov and Noah Fiedel and Noam Shazeer and Oriol Vinyals and Jeff Dean and Demis Hassabis and Koray Kavukcuoglu and Clement Farabet and Elena Buchatskaya and Jean-Baptiste Alayrac and Rohan Anil and Dmitry and Lepikhin and Sebastian Borgeaud and Olivier Bachem and Armand Joulin and Alek Andreev and Cassidy Hardin and Robert Dadashi and Léonard Hussenot},
      year={2025},
      eprint={2503.19786},
      archivePrefix={arXiv},
      primaryClass={cs.CL},
      url={https://arxiv.org/abs/2503.19786}, 
}

@inproceedings{ahia-etal-2023-languages,
    title = "{Do All Languages Cost the Same? Tokenization in the Era of Commercial Language Models}",
    author = "Ahia, Orevaoghene  and
      Kumar, Sachin  and
      Gonen, Hila  and
      Kasai, Jungo  and
      Mortensen, David  and
      Smith, Noah  and
      Tsvetkov, Yulia",
    editor = "Bouamor, Houda  and
      Pino, Juan  and
      Bali, Kalika",
    booktitle = "Proceedings of the 2023 Conference on Empirical Methods in Natural Language Processing",
    month = dec,
    year = "2023",
    address = "Singapore",
    publisher = "Association for Computational Linguistics",
    url = "https://aclanthology.org/2023.emnlp-main.614/",
    doi = "10.18653/v1/2023.emnlp-main.614",
    pages = "9904--9923"
}

@misc{lozhkov2024starcoder2stackv2,
      title={{StarCoder 2 and The Stack v2: The Next Generation}}, 
      author={Anton Lozhkov and Raymond Li and Loubna Ben Allal and Federico Cassano and Joel Lamy-Poirier and Nouamane Tazi and Ao Tang and Dmytro Pykhtar and Jiawei Liu and Yuxiang Wei and Tianyang Liu and Max Tian and Denis Kocetkov and Arthur Zucker and Younes Belkada and Zijian Wang and Qian Liu and Dmitry Abulkhanov and Indraneil Paul and Zhuang Li and Wen-Ding Li and Megan Risdal and Jia Li and Jian Zhu and Terry Yue Zhuo and Evgenii Zheltonozhskii and Nii Osae Osae Dade and Wenhao Yu and Lucas Krauß and Naman Jain and Yixuan Su and Xuanli He and Manan Dey and Edoardo Abati and Yekun Chai and Niklas Muennighoff and Xiangru Tang and Muhtasham Oblokulov and Christopher Akiki and Marc Marone and Chenghao Mou and Mayank Mishra and Alex Gu and Binyuan Hui and Tri Dao and Armel Zebaze and Olivier Dehaene and Nicolas Patry and Canwen Xu and Julian McAuley and Han Hu and Torsten Scholak and Sebastien Paquet and Jennifer Robinson and Carolyn Jane Anderson and Nicolas Chapados and Mostofa Patwary and Nima Tajbakhsh and Yacine Jernite and Carlos Muñoz Ferrandis and Lingming Zhang and Sean Hughes and Thomas Wolf and Arjun Guha and Leandro von Werra and Harm de Vries},
      year={2024},
      eprint={2402.19173},
      archivePrefix={arXiv},
      primaryClass={cs.SE},
      url={https://arxiv.org/abs/2402.19173}, 
}

@misc{hui2024qwen25codertechnicalreport,
      title={{Qwen2.5-Coder Technical Report}}, 
      author={Binyuan Hui and Jian Yang and Zeyu Cui and Jiaxi Yang and Dayiheng Liu and Lei Zhang and Tianyu Liu and Jiajun Zhang and Bowen Yu and Keming Lu and Kai Dang and Yang Fan and Yichang Zhang and An Yang and Rui Men and Fei Huang and Bo Zheng and Yibo Miao and Shanghaoran Quan and Yunlong Feng and Xingzhang Ren and Xuancheng Ren and Jingren Zhou and Junyang Lin},
      year={2024},
      eprint={2409.12186},
      archivePrefix={arXiv},
      primaryClass={cs.CL},
      url={https://arxiv.org/abs/2409.12186}, 
}

@misc{codegemmateam2024codegemmaopencodemodels,
      title={{CodeGemma: Open Code Models Based on Gemma}}, 
      author={Heri Zhao and Jeffrey Hui and Joshua Howland and Nam Nguyen and Siqi Zuo and Andrea Hu and Christopher A. Choquette-Choo and Jingyue Shen and Joe Kelley and Kshitij Bansal and Luke Vilnis and Mateo Wirth and Paul Michel and Peter Choy and Pratik Joshi and Ravin Kumar and Sarmad Hashmi and Shubham Agrawal and Zhitao Gong and Jane Fine and Tris Warkentin and Ale Jakse Hartman and Bin Ni and Kathy Korevec and Kelly Schaefer and Scott Huffman},
      year={2024},
      eprint={2406.11409},
      archivePrefix={arXiv},
      primaryClass={cs.CL},
      url={https://arxiv.org/abs/2406.11409}, 
}

@inproceedings{
liu2025superbpespacetravellanguage,
title={Super{BPE}: {Space Travel for Language Models}},
author={Alisa Liu and Jonathan Hayase and Valentin Hofmann and Sewoong Oh and Noah A. Smith and Yejin Choi},
booktitle={Tokenization Workshop},
year={2026},
url={https://openreview.net/forum?id=LwTWkSXIpt}
}

@inproceedings{Devlin2019BERTPO,
  title={{BERT: Pre-training of Deep Bidirectional Transformers for Language Understanding}},
  author={Jacob Devlin and Ming-Wei Chang and Kenton Lee and Kristina Toutanova},
  booktitle={North American Chapter of the Association for Computational Linguistics},
  year={2019},
  url={https://api.semanticscholar.org/CorpusID:52967399}
}

@inproceedings{NEURIPS2020_1457c0d6,
 author = {Brown, Tom and Mann, Benjamin and Ryder, Nick and Subbiah, Melanie and Kaplan, Jared D and Dhariwal, Prafulla and Neelakantan, Arvind and Shyam, Pranav and Sastry, Girish and Askell, Amanda and Agarwal, Sandhini and Herbert-Voss, Ariel and Krueger, Gretchen and Henighan, Tom and Child, Rewon and Ramesh, Aditya and Ziegler, Daniel and Wu, Jeffrey and Winter, Clemens and Hesse, Chris and Chen, Mark and Sigler, Eric and Litwin, Mateusz and Gray, Scott and Chess, Benjamin and Clark, Jack and Berner, Christopher and McCandlish, Sam and Radford, Alec and Sutskever, Ilya and Amodei, Dario},
 booktitle = {Advances in Neural Information Processing Systems},
 editor = {H. Larochelle and M. Ranzato and R. Hadsell and M.F. Balcan and H. Lin},
 pages = {1877--1901},
 publisher = {Curran Associates, Inc.},
 title = {{Language Models are Few-Shot Learners}},
 url = {https://proceedings.neurips.cc/paper_files/paper/2020/file/1457c0d6bfcb4967418bfb8ac142f64a-Paper.pdf},
 volume = {33},
 year = {2020}
}

@inproceedings{yang-etal-2022-textpruner,
    title = "{T}ext{P}runer: {A Model Pruning Toolkit for Pre-Trained Language Models}",
    author = "Yang, Ziqing  and
      Cui, Yiming  and
      Chen, Zhigang",
    editor = "Basile, Valerio  and
      Kozareva, Zornitsa  and
      Stajner, Sanja",
    booktitle = "Proceedings of the 60th Annual Meeting of the Association for Computational Linguistics: System Demonstrations",
    month = may,
    year = "2022",
    address = "Dublin, Ireland",
    publisher = "Association for Computational Linguistics",
    url = "https://aclanthology.org/2022.acl-demo.4/",
    doi = "10.18653/v1/2022.acl-demo.4",
    pages = "35--43"
}

@misc{grattafiori2024llama3herdmodels,
      title={{The Llama 3 Herd of Models}}, 
      author={Aaron Grattafiori and Abhimanyu Dubey and Abhinav Jauhri and Abhinav Pandey and Abhishek Kadian and Ahmad Al-Dahle and Aiesha Letman and Akhil Mathur and Alan Schelten and Alex Vaughan and Amy Yang and Angela Fan and Anirudh Goyal and Anthony Hartshorn and Aobo Yang and Archi Mitra and Archie Sravankumar and Artem Korenev and Arthur Hinsvark and Arun Rao and Aston Zhang and Aurelien Rodriguez and Austen Gregerson and Ava Spataru and Baptiste Roziere and Bethany Biron and Binh Tang and Bobbie Chern and Charlotte Caucheteux and Chaya Nayak and Chloe Bi and Chris Marra and Chris McConnell and Christian Keller and Christophe Touret and Chunyang Wu and Corinne Wong and Cristian Canton Ferrer and Cyrus Nikolaidis and Damien Allonsius and Daniel Song and Danielle Pintz and Danny Livshits and Danny Wyatt and David Esiobu and Dhruv Choudhary and Dhruv Mahajan and Diego Garcia-Olano and Diego Perino and Dieuwke Hupkes and Egor Lakomkin and Ehab AlBadawy and Elina Lobanova and Emily Dinan and Eric Michael Smith and Filip Radenovic and Francisco Guzmán and Frank Zhang and Gabriel Synnaeve and Gabrielle Lee and Georgia Lewis Anderson and Govind Thattai and Graeme Nail and Gregoire Mialon and Guan Pang and Guillem Cucurell and Hailey Nguyen and Hannah Korevaar and Hu Xu and Hugo Touvron and Iliyan Zarov and Imanol Arrieta Ibarra and Isabel Kloumann and Ishan Misra and Ivan Evtimov and Jack Zhang and Jade Copet and Jaewon Lee and Jan Geffert and Jana Vranes and Jason Park and Jay Mahadeokar and Jeet Shah and Jelmer van der Linde and Jennifer Billock and Jenny Hong and Jenya Lee and Jeremy Fu and Jianfeng Chi and Jianyu Huang and Jiawen Liu and Jie Wang and Jiecao Yu and Joanna Bitton and Joe Spisak and Jongsoo Park and Joseph Rocca and Joshua Johnstun and Joshua Saxe and Junteng Jia and Kalyan Vasuden Alwala and Karthik Prasad and Kartikeya Upasani and Kate Plawiak and Ke Li and Kenneth Heafield and Kevin Stone and Khalid El-Arini and Krithika Iyer and Kshitiz Malik and Kuenley Chiu and Kunal Bhalla and Kushal Lakhotia and Lauren Rantala-Yeary and Laurens van der Maaten and Lawrence Chen and Liang Tan and Liz Jenkins and Louis Martin and Lovish Madaan and Lubo Malo and Lukas Blecher and Lukas Landzaat and Luke de Oliveira and Madeline Muzzi and Mahesh Pasupuleti and Mannat Singh and Manohar Paluri and Marcin Kardas and Maria Tsimpoukelli and Mathew Oldham and Mathieu Rita and Maya Pavlova and Melanie Kambadur and Mike Lewis and Min Si and Mitesh Kumar Singh and Mona Hassan and Naman Goyal and Narjes Torabi and Nikolay Bashlykov and Nikolay Bogoychev and Niladri Chatterji and Ning Zhang and Olivier Duchenne and Onur Çelebi and Patrick Alrassy and Pengchuan Zhang and Pengwei Li and Petar Vasic and Peter Weng and Prajjwal Bhargava and Pratik Dubal and Praveen Krishnan and Punit Singh Koura and Puxin Xu and Qing He and Qingxiao Dong and Ragavan Srinivasan and Raj Ganapathy and Ramon Calderer and Ricardo Silveira Cabral and Robert Stojnic and Roberta Raileanu and Rohan Maheswari and Rohit Girdhar and Rohit Patel and Romain Sauvestre and Ronnie Polidoro and Roshan Sumbaly and Ross Taylor and Ruan Silva and Rui Hou and Rui Wang and Saghar Hosseini and Sahana Chennabasappa and Sanjay Singh and Sean Bell and Seohyun Sonia Kim and Sergey Edunov and Shaoliang Nie and Sharan Narang and Sharath Raparthy and Sheng Shen and Shengye Wan and Shruti Bhosale and Shun Zhang and Simon Vandenhende and Soumya Batra and Spencer Whitman and Sten Sootla and Stephane Collot and Suchin Gururangan and Sydney Borodinsky and Tamar Herman and Tara Fowler and Tarek Sheasha and Thomas Georgiou and Thomas Scialom and Tobias Speckbacher and Todor Mihaylov and Tong Xiao and Ujjwal Karn and Vedanuj Goswami and Vibhor Gupta and Vignesh Ramanathan and Viktor Kerkez and Vincent Gonguet and Virginie Do and Vish Vogeti and Vítor Albiero and Vladan Petrovic and Weiwei Chu and Wenhan Xiong and Wenyin Fu and Whitney Meers and Xavier Martinet and Xiaodong Wang and Xiaofang Wang and Xiaoqing Ellen Tan and Xide Xia and Xinfeng Xie and Xuchao Jia and Xuewei Wang and Yaelle Goldschlag and Yashesh Gaur and Yasmine Babaei and Yi Wen and Yiwen Song and Yuchen Zhang and Yue Li and Yuning Mao and Zacharie Delpierre Coudert and Zheng Yan and Zhengxing Chen and Zoe Papakipos and Aaditya Singh and Aayushi Srivastava and Abha Jain and Adam Kelsey and Adam Shajnfeld and Adithya Gangidi and Adolfo Victoria and Ahuva Goldstand and Ajay Menon and Ajay Sharma and Alex Boesenberg and Alexei Baevski and Allie Feinstein and Amanda Kallet and Amit Sangani and Amos Teo and Anam Yunus and Andrei Lupu and Andres Alvarado and Andrew Caples and Andrew Gu and Andrew Ho and Andrew Poulton and Andrew Ryan and Ankit Ramchandani and Annie Dong and Annie Franco and Anuj Goyal and Aparajita Saraf and Arkabandhu Chowdhury and Ashley Gabriel and Ashwin Bharambe and Assaf Eisenman and Azadeh Yazdan and Beau James and Ben Maurer and Benjamin Leonhardi and Bernie Huang and Beth Loyd and Beto De Paola and Bhargavi Paranjape and Bing Liu and Bo Wu and Boyu Ni and Braden Hancock and Bram Wasti and Brandon Spence and Brani Stojkovic and Brian Gamido and Britt Montalvo and Carl Parker and Carly Burton and Catalina Mejia and Ce Liu and Changhan Wang and Changkyu Kim and Chao Zhou and Chester Hu and Ching-Hsiang Chu and Chris Cai and Chris Tindal and Christoph Feichtenhofer and Cynthia Gao and Damon Civin and Dana Beaty and Daniel Kreymer and Daniel Li and David Adkins and David Xu and Davide Testuggine and Delia David and Devi Parikh and Diana Liskovich and Didem Foss and Dingkang Wang and Duc Le and Dustin Holland and Edward Dowling and Eissa Jamil and Elaine Montgomery and Eleonora Presani and Emily Hahn and Emily Wood and Eric-Tuan Le and Erik Brinkman and Esteban Arcaute and Evan Dunbar and Evan Smothers and Fei Sun and Felix Kreuk and Feng Tian and Filippos Kokkinos and Firat Ozgenel and Francesco Caggioni and Frank Kanayet and Frank Seide and Gabriela Medina Florez and Gabriella Schwarz and Gada Badeer and Georgia Swee and Gil Halpern and Grant Herman and Grigory Sizov and Guangyi and Zhang and Guna Lakshminarayanan and Hakan Inan and Hamid Shojanazeri and Han Zou and Hannah Wang and Hanwen Zha and Haroun Habeeb and Harrison Rudolph and Helen Suk and Henry Aspegren and Hunter Goldman and Hongyuan Zhan and Ibrahim Damlaj and Igor Molybog and Igor Tufanov and Ilias Leontiadis and Irina-Elena Veliche and Itai Gat and Jake Weissman and James Geboski and James Kohli and Janice Lam and Japhet Asher and Jean-Baptiste Gaya and Jeff Marcus and Jeff Tang and Jennifer Chan and Jenny Zhen and Jeremy Reizenstein and Jeremy Teboul and Jessica Zhong and Jian Jin and Jingyi Yang and Joe Cummings and Jon Carvill and Jon Shepard and Jonathan McPhie and Jonathan Torres and Josh Ginsburg and Junjie Wang and Kai Wu and Kam Hou U and Karan Saxena and Kartikay Khandelwal and Katayoun Zand and Kathy Matosich and Kaushik Veeraraghavan and Kelly Michelena and Keqian Li and Kiran Jagadeesh and Kun Huang and Kunal Chawla and Kyle Huang and Lailin Chen and Lakshya Garg and Lavender A and Leandro Silva and Lee Bell and Lei Zhang and Liangpeng Guo and Licheng Yu and Liron Moshkovich and Luca Wehrstedt and Madian Khabsa and Manav Avalani and Manish Bhatt and Martynas Mankus and Matan Hasson and Matthew Lennie and Matthias Reso and Maxim Groshev and Maxim Naumov and Maya Lathi and Meghan Keneally and Miao Liu and Michael L. Seltzer and Michal Valko and Michelle Restrepo and Mihir Patel and Mik Vyatskov and Mikayel Samvelyan and Mike Clark and Mike Macey and Mike Wang and Miquel Jubert Hermoso and Mo Metanat and Mohammad Rastegari and Munish Bansal and Nandhini Santhanam and Natascha Parks and Natasha White and Navyata Bawa and Nayan Singhal and Nick Egebo and Nicolas Usunier and Nikhil Mehta and Nikolay Pavlovich Laptev and Ning Dong and Norman Cheng and Oleg Chernoguz and Olivia Hart and Omkar Salpekar and Ozlem Kalinli and Parkin Kent and Parth Parekh and Paul Saab and Pavan Balaji and Pedro Rittner and Philip Bontrager and Pierre Roux and Piotr Dollar and Polina Zvyagina and Prashant Ratanchandani and Pritish Yuvraj and Qian Liang and Rachad Alao and Rachel Rodriguez and Rafi Ayub and Raghotham Murthy and Raghu Nayani and Rahul Mitra and Rangaprabhu Parthasarathy and Raymond Li and Rebekkah Hogan and Robin Battey and Rocky Wang and Russ Howes and Ruty Rinott and Sachin Mehta and Sachin Siby and Sai Jayesh Bondu and Samyak Datta and Sara Chugh and Sara Hunt and Sargun Dhillon and Sasha Sidorov and Satadru Pan and Saurabh Mahajan and Saurabh Verma and Seiji Yamamoto and Sharadh Ramaswamy and Shaun Lindsay and Shaun Lindsay and Sheng Feng and Shenghao Lin and Shengxin Cindy Zha and Shishir Patil and Shiva Shankar and Shuqiang Zhang and Shuqiang Zhang and Sinong Wang and Sneha Agarwal and Soji Sajuyigbe and Soumith Chintala and Stephanie Max and Stephen Chen and Steve Kehoe and Steve Satterfield and Sudarshan Govindaprasad and Sumit Gupta and Summer Deng and Sungmin Cho and Sunny Virk and Suraj Subramanian and Sy Choudhury and Sydney Goldman and Tal Remez and Tamar Glaser and Tamara Best and Thilo Koehler and Thomas Robinson and Tianhe Li and Tianjun Zhang and Tim Matthews and Timothy Chou and Tzook Shaked and Varun Vontimitta and Victoria Ajayi and Victoria Montanez and Vijai Mohan and Vinay Satish Kumar and Vishal Mangla and Vlad Ionescu and Vlad Poenaru and Vlad Tiberiu Mihailescu and Vladimir Ivanov and Wei Li and Wenchen Wang and Wenwen Jiang and Wes Bouaziz and Will Constable and Xiaocheng Tang and Xiaojian Wu and Xiaolan Wang and Xilun Wu and Xinbo Gao and Yaniv Kleinman and Yanjun Chen and Ye Hu and Ye Jia and Ye Qi and Yenda Li and Yilin Zhang and Ying Zhang and Yossi Adi and Youngjin Nam and Yu and Wang and Yu Zhao and Yuchen Hao and Yundi Qian and Yunlu Li and Yuzi He and Zach Rait and Zachary DeVito and Zef Rosnbrick and Zhaoduo Wen and Zhenyu Yang and Zhiwei Zhao and Zhiyu Ma},
      year={2024},
      eprint={2407.21783},
      archivePrefix={arXiv},
      primaryClass={cs.AI},
      url={https://arxiv.org/abs/2407.21783}, 
}

@article{gage1994new,
  title={{A New Algorithm for Data Compression}},
  author={Gage, Philip},
  journal={The C Users Journal},
  volume={12},
  number={2},
  pages={23--38},
  year={1994},
  publisher={R \& D Publications, Inc. Lawrence, KS, USA}
}

@inproceedings{sennrich-etal-2016-neural,
    title = "Neural {M}achine {T}ranslation of {R}are {W}ords with {S}ubword {U}nits",
    author = "Sennrich, Rico  and
      Haddow, Barry  and
      Birch, Alexandra",
    editor = "Erk, Katrin  and
      Smith, Noah A.",
    booktitle = "Proceedings of the 54th Annual Meeting of the Association for Computational Linguistics (Volume 1: Long Papers)",
    month = aug,
    year = "2016",
    address = "Berlin, Germany",
    publisher = "Association for Computational Linguistics",
    url = "https://aclanthology.org/P16-1162",
    doi = "10.18653/v1/P16-1162",
    pages = "1715--1725",
}

@misc{foroutan2025parityawarebytepairencodingimproving,
      title={{Parity-Aware Byte-Pair Encoding: Improving Cross-lingual Fairness in Tokenization}}, 
      author={Negar Foroutan and Clara Meister and Debjit Paul and Joel Niklaus and Sina Ahmadi and Antoine Bosselut and Rico Sennrich},
      year={2025},
      eprint={2508.04796},
      archivePrefix={arXiv},
      primaryClass={cs.CL},
      url={https://arxiv.org/abs/2508.04796}, 
}

@article{Wu2016GooglesNM,
  title={{Google's Neural Machine Translation System: Bridging the Gap between Human and Machine Translation}},
  author={Yonghui Wu and Mike Schuster and Z. Chen and Quoc V. Le and Mohammad Norouzi and Wolfgang Macherey and Maxim Krikun and Yuan Cao and Qin Gao and Klaus Macherey and Jeff Klingner and Apurva Shah and Melvin Johnson and Xiaobing Liu and Lukasz Kaiser and Stephan Gouws and Yoshikiyo Kato and Taku Kudo and Hideto Kazawa and Keith Stevens and George Kurian and Nishant Patil and Wei Wang and Cliff Young and Jason R. Smith and Jason Riesa and Alex Rudnick and Oriol Vinyals and Gregory S. Corrado and Macduff Hughes and Jeffrey Dean},
  journal={ArXiv},
  year={2016},
  volume={abs/1609.08144},
  url={https://api.semanticscholar.org/CorpusID:3603249}
}

@inproceedings{kudo-2018-subword,
    title = "{Subword Regularization: Improving Neural Network Translation Models with Multiple Subword Candidates}",
    author = "Kudo, Taku",
    editor = "Gurevych, Iryna  and
      Miyao, Yusuke",
    booktitle = "Proceedings of the 56th Annual Meeting of the Association for Computational Linguistics (Volume 1: Long Papers)",
    month = jul,
    year = "2018",
    address = "Melbourne, Australia",
    publisher = "Association for Computational Linguistics",
    url = "https://aclanthology.org/P18-1007/",
    doi = "10.18653/v1/P18-1007",
    pages = "66--75"
}

@inproceedings{provilkov-etal-2020-bpe,
    title = "{BPE}-{Dropout: Simple and Effective Subword Regularization}",
    author = "Provilkov, Ivan  and
      Emelianenko, Dmitrii  and
      Voita, Elena",
    editor = "Jurafsky, Dan  and
      Chai, Joyce  and
      Schluter, Natalie  and
      Tetreault, Joel",
    booktitle = "Proceedings of the 58th Annual Meeting of the Association for Computational Linguistics",
    month = jul,
    year = "2020",
    address = "Online",
    publisher = "Association for Computational Linguistics",
    url = "https://aclanthology.org/2020.acl-main.170/",
    doi = "10.18653/v1/2020.acl-main.170",
    pages = "1882--1892"
}

@inproceedings{Radford2019LanguageMA,
  title={{Language Models are Unsupervised Multitask Learners}},
  author={Alec Radford and Jeff Wu and Rewon Child and David Luan and Dario Amodei and Ilya Sutskever},
  year={2019},
  url={https://api.semanticscholar.org/CorpusID:160025533}
}

@misc{lee2024digitstodecisions,
      title={{From Digits to Decisions: How Tokenization Impacts Arithmetic in LLMs}},
      author={Garreth Lee and Guilherme Penedo and Leandro von Werra and Thomas Wolf},
      url={https://huggingface.co/spaces/huggingface/number-tokenization-blog},
      year={2024}
}

@inproceedings{
fried2023incoder,
title={{InCoder: A Generative Model for Code Infilling and Synthesis}},
author={Daniel Fried and Armen Aghajanyan and Jessy Lin and Sida Wang and Eric Wallace and Freda Shi and Ruiqi Zhong and Scott Yih and Luke Zettlemoyer and Mike Lewis},
booktitle={The Eleventh International Conference on Learning Representations },
year={2023},
url={https://openreview.net/forum?id=hQwb-lbM6EL}
}

@inproceedings{cognetta-etal-2024-analysis,
    title = "{An Analysis of {BPE} Vocabulary Trimming in Neural Machine Translation}",
    author = "Cognetta, Marco  and
      Hiraoka, Tatsuya  and
      Sennrich, Rico  and
      Pinter, Yuval  and
      Okazaki, Naoaki",
    editor = "Tafreshi, Shabnam  and
      Akula, Arjun  and
      Sedoc, Jo{\~a}o  and
      Drozd, Aleksandr  and
      Rogers, Anna  and
      Rumshisky, Anna",
    booktitle = "Proceedings of the Fifth Workshop on Insights from Negative Results in NLP",
    month = jun,
    year = "2024",
    address = "Mexico City, Mexico",
    publisher = "Association for Computational Linguistics",
    url = "https://aclanthology.org/2024.insights-1.7/",
    doi = "10.18653/v1/2024.insights-1.7",
    pages = "48--50"
}

@inproceedings{vilar-federico-2021-statistical,
    title = "{A Statistical Extension of Byte-Pair Encoding}",
    author = "Vilar, David  and
      Federico, Marcello",
    editor = "Federico, Marcello  and
      Waibel, Alex  and
      Costa-juss{\`a}, Marta R.  and
      Niehues, Jan  and
      Stuker, Sebastian  and
      Salesky, Elizabeth",
    booktitle = "Proceedings of the 18th International Conference on Spoken Language Translation (IWSLT 2021)",
    month = aug,
    year = "2021",
    address = "Bangkok, Thailand (online)",
    publisher = "Association for Computational Linguistics",
    url = "https://aclanthology.org/2021.iwslt-1.31/",
    doi = "10.18653/v1/2021.iwslt-1.31",
    pages = "263--275"
}

@article{church-hanks-1990-word,
    title = "{Word Association Norms, Mutual Information, and Lexicography}",
    author = "Church, Kenneth Ward  and
      Hanks, Patrick",
    journal = "Computational Linguistics",
    volume = "16",
    number = "1",
    year = "1990",
    url = "https://aclanthology.org/J90-1003/",
    pages = "22--29"
}

@INPROCEEDINGS{schuster2012japanese,
  author={Schuster, Mike and Nakajima, Kaisuke},
  booktitle={2012 IEEE International Conference on Acoustics, Speech and Signal Processing (ICASSP)}, 
  title={{Japanese and Korean voice search}}, 
  year={2012},
  volume={},
  number={},
  pages={5149-5152},
  doi={10.1109/ICASSP.2012.6289079}}

@inproceedings{purason2025teachingoldtokenizersnew,
    title = "Teaching Old Tokenizers New Words: Efficient Tokenizer Adaptation for Pretrained Models",
    author = "Purason, Taido  and
      Chizhov, Pavel  and
      Yamshchikov, Ivan P.  and
      Fishel, Mark",
    editor = "Demberg, Vera  and
      Inui, Kentaro  and
      Marquez, Llu{\'i}s",
    booktitle = "Findings of the {A}ssociation for {C}omputational {L}inguistics: {EACL} 2026",
    month = mar,
    year = "2026",
    address = "Rabat, Morocco",
    publisher = "Association for Computational Linguistics",
    url = "https://aclanthology.org/2026.findings-eacl.341/",
    doi = "10.18653/v1/2026.findings-eacl.341",
    pages = "6492--6516",
    ISBN = "979-8-89176-386-9"
}

@book{salton_introduction_1983,
  address = {New York, NY},
  author = {Salton, Gerard and McGill, Michael},
  publisher = {McGraw-Hill},
  title = {{Introduction to Modern Information Retrieval}},
  year = 1983
}

@inproceedings{10.1609/aaai.v39i23.34633,
author = {Lian, Haoran and Xiong, Yizhe and Niu, Jianwei and Mo, Shasha and Su, Zhenpeng and Lin, Zijia and Chen, Hui and Han, Jungong and Ding, Guiguang},
title = {{Scaffold-BPE: Enhancing Byte Pair Encoding for Large Language Models with Simple and Effective Scaffold Token Removal}},
year = {2025},
isbn = {978-1-57735-897-8},
publisher = {AAAI Press},
url = {https://doi.org/10.1609/aaai.v39i23.34633},
doi = {10.1609/aaai.v39i23.34633},
booktitle = {Proceedings of the Thirty-Ninth AAAI Conference on Artificial Intelligence and Thirty-Seventh Conference on Innovative Applications of Artificial Intelligence and Fifteenth Symposium on Educational Advances in Artificial Intelligence},
articleno = {2735},
numpages = {10},
series = {AAAI'25/IAAI'25/EAAI'25}
}

@inproceedings{chizhov-etal-2024-bpe,
    title = "{BPE} {Gets Picky: Efficient Vocabulary Refinement During Tokenizer Training}",
    author = "Chizhov, Pavel  and
      Arnett, Catherine  and
      Korotkova, Elizaveta  and
      Yamshchikov, Ivan P.",
    editor = "Al-Onaizan, Yaser  and
      Bansal, Mohit  and
      Chen, Yun-Nung",
    booktitle = "Proceedings of the 2024 Conference on Empirical Methods in Natural Language Processing",
    month = nov,
    year = "2024",
    address = "Miami, Florida, USA",
    publisher = "Association for Computational Linguistics",
    url = "https://aclanthology.org/2024.emnlp-main.925/",
    doi = "10.18653/v1/2024.emnlp-main.925",
    pages = "16587--16604"
}

@inproceedings{land-bartolo-2024-fishing,
    title = "{Fishing for Magikarp: Automatically Detecting Under-trained Tokens in Large Language Models}",
    author = "Land, Sander  and
      Bartolo, Max",
    editor = "Al-Onaizan, Yaser  and
      Bansal, Mohit  and
      Chen, Yun-Nung",
    booktitle = "Proceedings of the 2024 Conference on Empirical Methods in Natural Language Processing",
    month = nov,
    year = "2024",
    address = "Miami, Florida, USA",
    publisher = "Association for Computational Linguistics",
    url = "https://aclanthology.org/2024.emnlp-main.649/",
    doi = "10.18653/v1/2024.emnlp-main.649",
    pages = "11631--11646"
}

\appendix

\section{Rule-Based Classification}
\label{app:classification}

To classify the tokens, we use the following rules and regular expressions:

\begin{itemize}
    \item \textbf{Special tokens:} tokens fitting the regular expression \texttt{r``<[\^{}>]+>''}.
    \item \textbf{Digits and numbers:} digit sequences, optionally preceeded by a \texttt{``+''} or a \texttt{``-''} sign.
    \item \textbf{Punctuation:} tokens where symbols that are present in \texttt{string.punctuation} or that have a Unicode category that assumes punctuation constitute more than 80\%. 
    \item \textbf{Variable or function names:} tokens that consist of words merged via underscores (\texttt{``\_''}) or matching the regular expression \texttt{r``\^{}[a-zA-Z]*[A-Z][a-zA-Z]*\$''}.
    \item \textbf{ALL–CAPS Latin words:} tokens consisting of Latin letters, all in upper case.
    \item \textbf{Other Latin words:} tokens consisting of Latin letters that did not classify into previous categories.
    \item \textbf{Non-Latin words or characters:} tokens that contain non-Latin characters.
    \item \textbf{Other:} tokens not falling into any category.
\end{itemize}

\section{Model and Training}
\label{app:architecture}

In Tables~\ref{tab:model_params} and \ref{tab:training_params}, we show the main model and training parameters, respectively.

\begin{table}[!h]
\small
\centering
\begin{tabular}{lc}
\toprule
\textbf{Model Parameter} & \textbf{Value} \\
\midrule
Hidden size & 768 \\
Intermediate size & 2048 \\
Number of hidden layers & 12 \\
Number of attention heads & 12 \\
Number of key-value heads & 4 \\
Max position embeddings & 2048 \\
Vocabulary size & 32768 \\
Hidden activation & \texttt{silu} \\
Initializer range & 0.02 \\
Tie word embeddings & \texttt{true} \\
RMS norm epsilon & $1.0 \times 10^{-5}$ \\
ROPE theta & 10000 \\
\bottomrule
\end{tabular}
\caption{Main model configuration parameters}
\label{tab:model_params}
\end{table}

\begin{table}[!t]
\small
\centering
\begin{tabular}{ll}
\toprule
\textbf{Training Parameter} & \textbf{Value} \\
\midrule
Optimizer & \texttt{adamW} \\
Weight decay & 0.01 \\
Gradient clipping & 1.0 \\
Adam beta1 & 0.9 \\
Adam beta2 & 0.95 \\
Adam epsilon & $1.0 \times 10^{-8}$ \\
Warmup steps & 5000 \\
Total steps & 60000 \\
Learning rate decay style & \texttt{linear} \\
Learning rate warmup style & \texttt{linear} \\
Max learning rate & 0.0003 \\
End learning rate & 0.000003 \\
\bottomrule
\end{tabular}
\caption{Main training configuration parameters}
\label{tab:training_params}
\end{table}

\section{Data}
\label{app:data}

In Tables~\ref{tab:language_stats}~and~\ref{tab:language_extensions}, we show language statistics for the training corpus and the extensions chosen from the repositories for each language in the process of assembling the evaluation dataset, respectively.

\begin{table}[!tbp]
\small
\centering
\begin{tabular}{lrc}
\toprule
\textbf{Language} & \textbf{\# Documents} & \textbf{Size (in GB)} \\
\midrule
Java & 2,741,833 & 11.15 \\
JavaScript & 2,757,358 & 8.40 \\
Python & 1,770,216 & 7.90 \\
PHP & 2,006,486 & 7.02 \\
C++ & 990,971 & 6.67 \\
C\# & 1,468,542 & 5.70 \\
Go & 615,934 & 2.81 \\
Rust & 185,878 & 1.13 \\
Ruby & 589,494 & 1.07 \\
Kotlin & 320,726 & 0.78 \\
Scala & 242,087 & 0.76 \\
Swift & 234,583 & 0.76 \\
Vue & 182,438 & 0.75 \\
Dart & 116,085 & 0.42 \\
Lua & 71,166 & 0.33 \\
Haskell & 73,609 & 0.28 \\
Julia & 37,525 & 0.16 \\
OCaml & 23,231 & 0.14 \\
\bottomrule
\end{tabular}
\caption{Per-language dataset statistics}
\label{tab:language_stats}
\end{table}

\begin{table}[!tbp]
\small
\centering
\begin{tabular}{ll}
\toprule
\textbf{Language} & \textbf{Extensions} \\ \midrule
Java & \texttt{java} \\
C\# & \texttt{cs} \\
C++ & \texttt{cpp}, \texttt{hpp} \\
Python & \texttt{py} \\
Haskell & \texttt{hs} \\
Dart & \texttt{dart} \\
Go & \texttt{go} \\
JavaScript & \texttt{js} \\
Julia & \texttt{jl} \\
Kotlin & \texttt{kt} \\
Ruby & \texttt{rb} \\
Rust & \texttt{rs} \\
Scala & \texttt{scala}, \texttt{sc} \\
Swift & \texttt{swift} \\
Vue & \texttt{vue} \\
PHP & \texttt{php} \\
Lua & \texttt{lua} \\
OCaml & \texttt{ml}, \texttt{mli} \\ \bottomrule
\end{tabular}
\caption{File extensions chosen for each language for the evaluation set.}
\label{tab:language_extensions}
\end{table}

\section{Language Skip Criterion}
\label{sec:app-min_languages}

\begin{figure}[htbp]
    \centering
    \includegraphics[width=\linewidth]{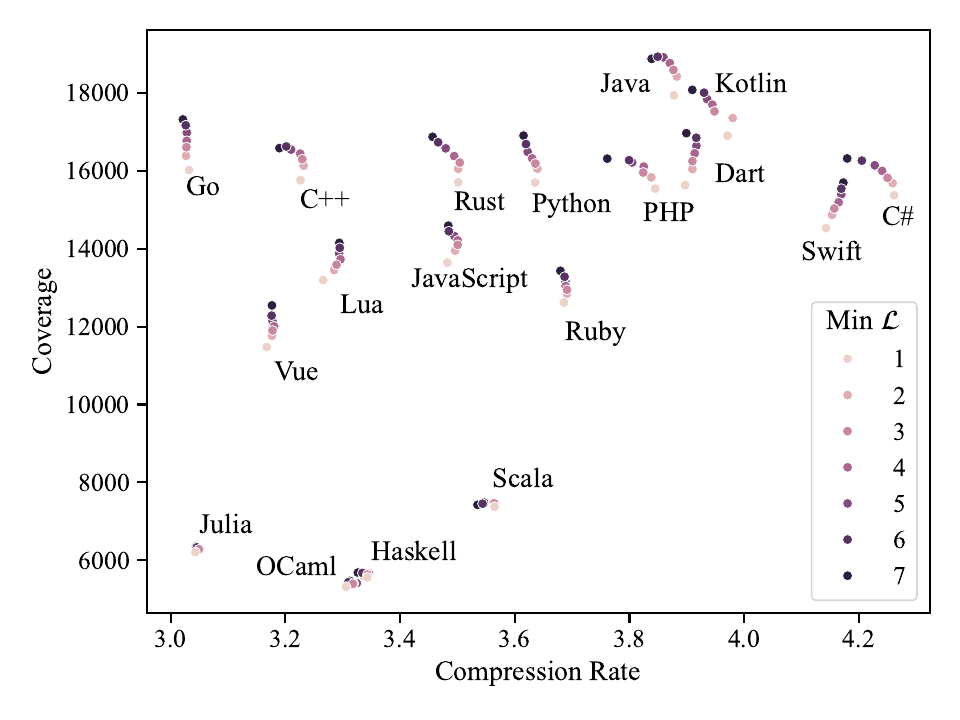}
    \caption{Models with thresholds for the minimum number of languages evaluated for compression rate and coverage. Minimum $\mathcal{L} = 1$ is regular BPE. Tokenizers with $\mathcal{L} >7$ resulted in exhausting the merge queue.}
    \label{fig:min_languages}
\end{figure}

We separately investigate different numbers of languages $\mathcal{L}$ chosen as skip criteria. Our results in Figure~\ref{fig:min_languages} show that with the increase of the threshold, for some languages (mainly higher-resource ones: C\#, Java, C++), the compression tends to decrease, while slightly increasing for others (mainly low-resource ones: Lua, Dart, Vue). This corresponds well to findings of~\citet{foroutan2025parityawarebytepairencodingimproving}, where the authors also observe a slight decrease in overall compression rate while improving it for lower-resource languages. For most of the languages, the increase in threshold is accompanied by improved coverage. 

In addition, in Table~\ref{tab:gini-compression}, we show that there is a tradeoff between compression rate and gini index for the compression rates per language.

\section{Time and Space Complexity}
\label{app:complexity}

\paragraph{Training.} Out of the two modifications, only the merge skip criterion increases the number of main BPE loop steps. However, the most computation-intensive algorithm parts are related to post-merge frequency and corpus updates, which do not happen if a merge is skipped. Therefore, this part potentially adds a new component with an upper bound of $\mathcal{O}\left(\left|Q\right|\right)$ to the time complexity, where $\left|Q\right|$ is the size of the priority queue. This does not change the overall time complexity as $\mathcal{O}\left(\left|Q\right|\right)$ is already the complexity of the queue initialization.

The main difference from the original BPE, however, lies in memory consumption, as our modifications require separate structures to maintain the mappings of each pair to its repositories and languages, along with the separate frequency counters. Therefore, the memory consumption increases by $\mathcal{O}\left(P\cdot\left( \mathcal{R}_{max} + \mathcal{L}_{max}\right)\right)$, where $P$ is the number of pairs in the queue. The updates in these structures also scale the merge time complexity during training comparably to the frequency-based updates, based on the number of repositories and/or languages in which the pair is present.

\paragraph{Inference.}The inference stage of all our modifications is identical to that of the basic BPE, as every modification only affects the order of merges and the resulting vocabulary. Thus, we do not bring any overhead to the inference procedure, in terms of both memory and time. Instead, having a more optimal vocabulary leads to a slight increase in inference speed due to improved vocabulary utilization and compression.

\begin{table}[!tbp]
\small
\centering
\begin{tabular}{ccc}
\toprule
\textbf{Min $\mathcal{L}$} &\textbf{Gini} &\textbf{Compression} \\\midrule
1 (BPE) &0.0576 &\cellcolor[HTML]{7ac9a2}3.5693 \\
2 &0.0576 &\cellcolor[HTML]{57bb8a}3.5743 \\
3 &\cellcolor[HTML]{c2e6d4}0.0570 &\cellcolor[HTML]{68c296}3.5719 \\
4 &\cellcolor[HTML]{b2dfc9}0.0568 &\cellcolor[HTML]{6ec59a}3.5710 \\
5 &\cellcolor[HTML]{a5dac0}0.0567 &\cellcolor[HTML]{96d5b6}3.5651 \\
6 &\cellcolor[HTML]{9dd7ba}0.0566 &\cellcolor[HTML]{c1e6d4}3.5588 \\
7 &\cellcolor[HTML]{57bb8a}0.0559 &3.5498 \\
\bottomrule
\end{tabular}
\caption{Gini index and compression rate for tokenizers regularized with language-based merge skip criteria.}
\label{tab:gini-compression}
\end{table}

\begin{figure}[tbp]
    \centering
    \includegraphics[width=\linewidth]{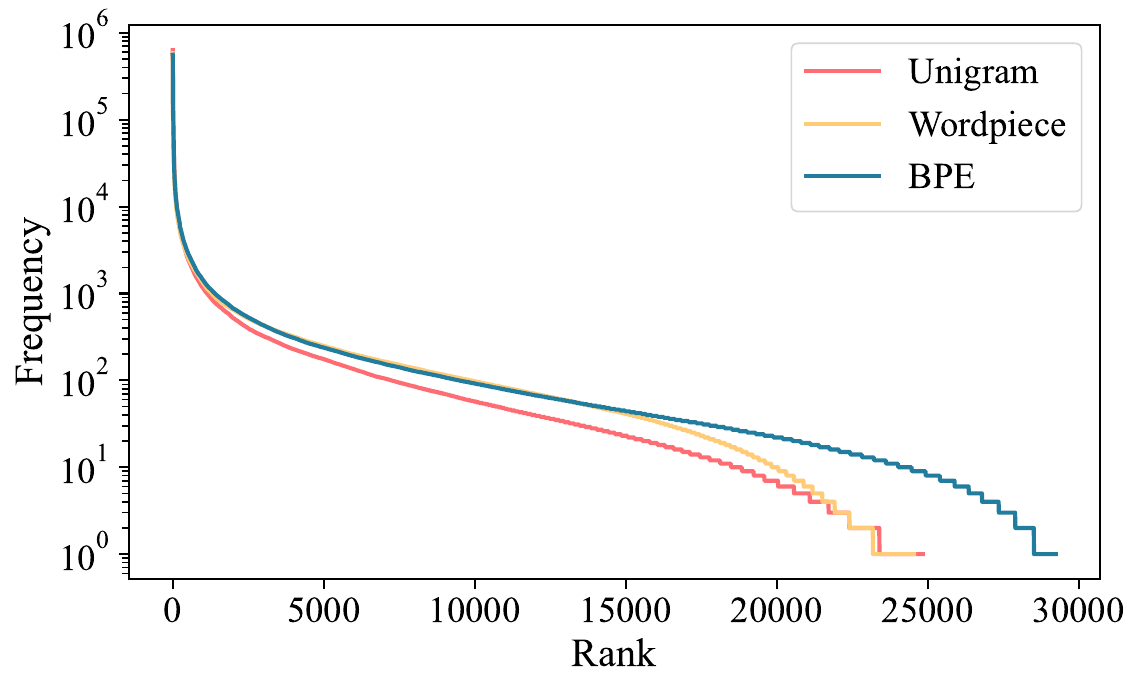}
    \caption{Token frequencies distribution in the evaluation corpus for basic BPE, Unigram, and Wordpiece tokenizers. Total coverage for BPE is 89.23\%, for Unigram 75.79\%, and for Wordpiece 74.92\%.}
    \label{fig:unigram-wordpiece}
\end{figure}

\begin{table*}[t]
\centering
\scriptsize
%\resizebox{ extwidth}{!}{ % use this if the table is too large
\begin{tabular}{lcrrrrrrrrrr}\toprule
\textbf{Priority Criterion} &\textbf{Merge Skip} &\textbf{Java} &\textbf{JavaScript} &\textbf{Python} &\textbf{PHP} &\textbf{C++} &\textbf{C\#} &\textbf{Go} &\textbf{Rust} &\textbf{Ruby} \\\midrule
$\mathcal{F}$ (BPE) &--- (BPE) &\cellcolor[HTML]{f1f9f5}3.875 &3.481 &\cellcolor[HTML]{f0f9f4}3.634 &\cellcolor[HTML]{daf0e5}3.842 &\cellcolor[HTML]{fdfefe}3.225 &\cellcolor[HTML]{dbf0e6}4.265 &\cellcolor[HTML]{ecf7f2}3.030 &\cellcolor[HTML]{e8f6ef}3.499 &3.683 \\
\midrule
$\mathcal{F}$ &Skip $\mathcal{L} < 4$ &3.867 &\cellcolor[HTML]{daf0e5}3.499 &3.629 &3.822 &3.224 &4.244 &3.026 &3.492 &\cellcolor[HTML]{f5fbf8}3.686 \\
$\mathcal{F} \cdot \mathcal{L}$ &--- &\cellcolor[HTML]{d8efe4}3.887 &\cellcolor[HTML]{c9e9da}3.507 &\cellcolor[HTML]{cfebde}3.644 &\cellcolor[HTML]{ceebdd}3.848 &\cellcolor[HTML]{c3e6d5}3.238 &\cellcolor[HTML]{d9efe5}4.267 &\cellcolor[HTML]{cae9da}3.037 &\cellcolor[HTML]{cfebde}3.507 &\cellcolor[HTML]{def2e9}3.692 \\
$\mathcal{F} \cdot \mathcal{L}$ &Skip $\mathcal{L} = 1$ &\cellcolor[HTML]{e5f4ed}3.880 &\cellcolor[HTML]{c6e8d8}3.508 &\cellcolor[HTML]{d2ede0}3.643 &\cellcolor[HTML]{e3f3eb}3.837 &\cellcolor[HTML]{d0ecdf}3.235 &\cellcolor[HTML]{def2e9}4.263 &\cellcolor[HTML]{ebf7f1}3.030 &\cellcolor[HTML]{d2ece0}3.506 &\cellcolor[HTML]{ddf1e7}3.693 \\
$\mathcal{F} \cdot \log\left(\mathcal{R} + 1\right)$ &--- &\cellcolor[HTML]{bce3d0}3.901 &\cellcolor[HTML]{ceebdd}3.505 &\cellcolor[HTML]{bde4d1}3.650 &\cellcolor[HTML]{bae2cf}3.859 &\cellcolor[HTML]{bce3d0}3.239 &\cellcolor[HTML]{bae2cf}4.284 &\cellcolor[HTML]{bde4d1}3.040 &\cellcolor[HTML]{bbe3d0}3.513 &\cellcolor[HTML]{bee4d2}3.702 \\
$\mathcal{F} \cdot \log\mathcal{R}$ &--- &\cellcolor[HTML]{b7e1cd}3.903 &\cellcolor[HTML]{c9e9da}3.507 &\cellcolor[HTML]{b7e1cd}3.652 &\cellcolor[HTML]{b7e1cd}3.860 &\cellcolor[HTML]{b7e1cd}3.240 &\cellcolor[HTML]{b7e1cd}4.286 &\cellcolor[HTML]{b7e1cd}3.041 &\cellcolor[HTML]{b7e1cd}3.514 &\cellcolor[HTML]{b7e1cd}3.703 \\
\midrule
$\mathcal{F} \cdot \log\left(\mathcal{R} + 1\right)$ &Skip $\mathcal{L} < 4$ &\cellcolor[HTML]{f0f9f5}3.875 &\cellcolor[HTML]{c9e9da}3.507 &\cellcolor[HTML]{f1f9f5}3.633 &\cellcolor[HTML]{f8fcfa}3.826 &\cellcolor[HTML]{f7fcf9}3.226 &\cellcolor[HTML]{f8fcfa}4.249 &\cellcolor[HTML]{f2faf6}3.029 &\cellcolor[HTML]{f7fcfa}3.494 &\cellcolor[HTML]{e3f3eb}3.691 \\
$\mathcal{F} \cdot \log\left(\mathcal{R} + 1\right) \cdot \mathcal{L}$ &--- &\cellcolor[HTML]{c6e7d7}3.896 &\cellcolor[HTML]{b9e2cf}3.514 &\cellcolor[HTML]{b9e2cf}3.651 &\cellcolor[HTML]{c4e7d6}3.853 &\cellcolor[HTML]{bce3d1}3.239 &\cellcolor[HTML]{cae9da}4.275 &\cellcolor[HTML]{bde4d1}3.040 &\cellcolor[HTML]{c1e5d4}3.511 &\cellcolor[HTML]{d5eee2}3.695 \\
$\mathcal{F} \cdot \log\mathcal{R} \cdot \mathcal{L}$ &--- &\cellcolor[HTML]{c5e7d7}3.896 &\cellcolor[HTML]{b9e2ce}3.515 &\cellcolor[HTML]{b9e2ce}3.651 &\cellcolor[HTML]{c4e7d6}3.854 &\cellcolor[HTML]{bee4d2}3.239 &\cellcolor[HTML]{c9e9d9}4.276 &\cellcolor[HTML]{bce3d1}3.040 &\cellcolor[HTML]{c1e5d4}3.511 &\cellcolor[HTML]{d3ede0}3.696 \\
$\mathcal{F} \cdot \log\mathcal{R} \cdot \mathcal{L}$ &Skip $\mathcal{L} = 1$ &\cellcolor[HTML]{d2ede0}3.890 &\cellcolor[HTML]{b7e1cd}3.515 &\cellcolor[HTML]{bee4d2}3.650 &\cellcolor[HTML]{daf0e6}3.842 &\cellcolor[HTML]{cae9da}3.236 &\cellcolor[HTML]{d3ede0}4.270 &\cellcolor[HTML]{def1e8}3.033 &\cellcolor[HTML]{cceadc}3.508 &\cellcolor[HTML]{d1ecdf}3.696 \\
\bottomrule
\end{tabular}
\caption{Text compression by language for the nine higher-resource languages.}\label{tab:lang-compression-1}
\end{table*}

\begin{table*}[!t]
\centering
\scriptsize
%\resizebox{ extwidth}{!}{ % use this if the table is too large
\begin{tabular}{lcrrrrrrrrrr}\toprule
\textbf{Priority Criterion} &\textbf{Merge Skip} &\textbf{Kotlin} &\textbf{Scala} &\textbf{Swift} &\textbf{Vue} &\textbf{Dart} &\textbf{Lua} &\textbf{Haskell} &\textbf{Julia} &\textbf{OCaml} \\\midrule
$\mathcal{F}$ (BPE) &--- (BPE) &\cellcolor[HTML]{d8efe4}3.968 &\cellcolor[HTML]{f4fbf7}3.562 &4.140 &3.165 &3.894 &3.264 &3.340 &3.041 &3.305 \\
\midrule
$\mathcal{F}$ &Skip $\mathcal{L} < 4$ &3.941 &3.559 &\cellcolor[HTML]{ceebdd}4.162 &\cellcolor[HTML]{daf0e6}3.178 &\cellcolor[HTML]{d3ede1}3.911 &\cellcolor[HTML]{bee4d2}3.294 &3.340 &\cellcolor[HTML]{dbf0e6}3.049 &\cellcolor[HTML]{def2e8}3.319 \\
$\mathcal{F} \cdot \mathcal{L}$ &--- &\cellcolor[HTML]{c6e7d7}3.980 &\cellcolor[HTML]{dff2e9}3.567 &\cellcolor[HTML]{c2e6d5}4.167 &\cellcolor[HTML]{c3e6d6}3.186 &\cellcolor[HTML]{d1ecdf}3.912 &\cellcolor[HTML]{bbe3d0}3.296 &\cellcolor[HTML]{e3f3eb}3.350 &\cellcolor[HTML]{bde4d1}3.055 &\cellcolor[HTML]{b9e2ce}3.334 \\
$\mathcal{F} \cdot \mathcal{L}$ &Skip $\mathcal{L} = 1$ &\cellcolor[HTML]{c4e7d6}3.980 &\cellcolor[HTML]{ebf7f1}3.564 &\cellcolor[HTML]{c4e7d6}4.166 &\cellcolor[HTML]{bfe5d3}3.187 &\cellcolor[HTML]{d0ecde}3.912 &\cellcolor[HTML]{b9e2cf}3.296 &\cellcolor[HTML]{f9fdfb}3.342 &\cellcolor[HTML]{b7e1cd}3.056 &\cellcolor[HTML]{b7e1cd}3.335 \\
$\mathcal{F} \cdot \log\left(\mathcal{R} + 1\right)$ &--- &\cellcolor[HTML]{bde4d2}3.985 &\cellcolor[HTML]{c0e5d4}3.574 &\cellcolor[HTML]{cceadc}4.162 &\cellcolor[HTML]{d0ecdf}3.181 &\cellcolor[HTML]{bce3d1}3.920 &\cellcolor[HTML]{d4ede1}3.284 &\cellcolor[HTML]{d0ecde}3.356 &\cellcolor[HTML]{cbe9db}3.052 &\cellcolor[HTML]{dff2e9}3.318 \\
$\mathcal{F} \cdot \log\mathcal{R}$ &--- &\cellcolor[HTML]{bae3cf}3.987 &\cellcolor[HTML]{b7e1cd}3.576 &\cellcolor[HTML]{c3e6d6}4.166 &\cellcolor[HTML]{cbeadb}3.183 &\cellcolor[HTML]{b7e1cd}3.921 &\cellcolor[HTML]{d2ece0}3.285 &\cellcolor[HTML]{cbeadb}3.357 &\cellcolor[HTML]{bfe5d3}3.055 &\cellcolor[HTML]{e4f4ec}3.316 \\
\midrule
$\mathcal{F} \cdot \log\left(\mathcal{R} + 1\right)$ &Skip $\mathcal{L} < 4$ &\cellcolor[HTML]{f5fbf8}3.948 &\cellcolor[HTML]{eff9f4}3.563 &\cellcolor[HTML]{c0e5d4}4.167 &\cellcolor[HTML]{cfebde}3.182 &\cellcolor[HTML]{c0e5d3}3.918 &\cellcolor[HTML]{bfe4d2}3.294 &\cellcolor[HTML]{f2faf6}3.344 &\cellcolor[HTML]{bfe5d3}3.055 &\cellcolor[HTML]{d5eee2}3.322 \\
$\mathcal{F} \cdot \log\left(\mathcal{R} + 1\right) \cdot \mathcal{L}$ &--- &\cellcolor[HTML]{b9e2ce}3.988 &\cellcolor[HTML]{dff2e9}3.567 &\cellcolor[HTML]{b8e2ce}4.171 &\cellcolor[HTML]{bae2cf}3.189 &\cellcolor[HTML]{bee4d2}3.919 &\cellcolor[HTML]{b9e2ce}3.296 &\cellcolor[HTML]{b7e1cd}3.364 &\cellcolor[HTML]{c0e5d4}3.054 &\cellcolor[HTML]{bfe5d3}3.331 \\
$\mathcal{F} \cdot \log\mathcal{R} \cdot \mathcal{L}$ &--- &\cellcolor[HTML]{b8e2ce}3.989 &\cellcolor[HTML]{def1e8}3.567 &\cellcolor[HTML]{b7e1cd}4.171 &\cellcolor[HTML]{b9e2ce}3.190 &\cellcolor[HTML]{bee4d2}3.919 &\cellcolor[HTML]{b8e2ce}3.296 &\cellcolor[HTML]{b9e2cf}3.363 &\cellcolor[HTML]{c0e5d3}3.055 &\cellcolor[HTML]{bee4d2}3.332 \\
$\mathcal{F} \cdot \log\mathcal{R} \cdot \mathcal{L}$ &Skip $\mathcal{L} = 1$ &\cellcolor[HTML]{b7e1cd}3.989 &\cellcolor[HTML]{edf8f3}3.564 &\cellcolor[HTML]{bee4d2}4.169 &\cellcolor[HTML]{b7e1cd}3.190 &\cellcolor[HTML]{bce3d1}3.920 &\cellcolor[HTML]{b7e1cd}3.297 &\cellcolor[HTML]{f0f9f5}3.345 &\cellcolor[HTML]{c1e5d4}3.054 &\cellcolor[HTML]{bee4d2}3.332 \\
\bottomrule
\end{tabular}
\caption{Text compression by language for the nine lower-resource languages.}\label{tab:lang-compression-2}
\end{table*}

\begin{table*}[!t]
\centering
\scriptsize
%\resizebox{ extwidth}{!}{ % use this if the table is too large
\begin{tabular}{lcrrrrrrrrrr}\toprule
\textbf{Priority Criterion} &\textbf{Merge Skip} &\textbf{Java} &\textbf{JavaScript} &\textbf{Python} &\textbf{PHP} &\textbf{C++} &\textbf{C\#} &\textbf{Go} &\textbf{Rust} &\textbf{Ruby} \\\midrule
$\mathcal{F}$ (BPE) &--- (BPE) &\cellcolor[HTML]{fff2cc}17935 &\cellcolor[HTML]{fff2cc}13643 &\cellcolor[HTML]{fff2cc}15693 &\cellcolor[HTML]{fff2cc}15541 &\cellcolor[HTML]{fff2cc}15755 &\cellcolor[HTML]{fff2cc}15370 &\cellcolor[HTML]{fff2cc}16019 &\cellcolor[HTML]{fff2cc}15701 &\cellcolor[HTML]{fff2cc}12618 \\
\midrule
$\mathcal{F}$ &Skip $\mathcal{L} < 4$ &\cellcolor[HTML]{fbd5a8}18763 &\cellcolor[HTML]{fcd9ad}14211 &\cellcolor[HTML]{fcdaae}16315 &\cellcolor[HTML]{fcd9ad}16106 &\cellcolor[HTML]{fbd4a7}16437 &\cellcolor[HTML]{fbd7ab}15993 &\cellcolor[HTML]{fbd8ab}16767 &\cellcolor[HTML]{fbd7aa}16376 &\cellcolor[HTML]{fcdbb0}13041 \\
$\mathcal{F} \cdot \mathcal{L}$ &--- &\cellcolor[HTML]{fbd3a6}18816 &\cellcolor[HTML]{fad2a4}14359 &\cellcolor[HTML]{fad1a3}16541 &\cellcolor[HTML]{fbd2a5}16266 &\cellcolor[HTML]{facfa1}16549 &\cellcolor[HTML]{fad0a2}16163 &\cellcolor[HTML]{facfa1}16993 &\cellcolor[HTML]{face9f}16594 &\cellcolor[HTML]{fad0a2}13251 \\
$\mathcal{F} \cdot \mathcal{L}$ &Skip $\mathcal{L} = 1$ &\cellcolor[HTML]{fad2a4}18850 &\cellcolor[HTML]{fad1a3}14377 &\cellcolor[HTML]{fad0a2}16561 &\cellcolor[HTML]{fad2a4}16278 &\cellcolor[HTML]{facfa0}16566 &\cellcolor[HTML]{facfa1}16188 &\cellcolor[HTML]{facfa1}17007 &\cellcolor[HTML]{facc9d}16625 &\cellcolor[HTML]{facfa1}13267 \\
$\mathcal{F} \cdot \log\left(\mathcal{R} + 1\right)$ &--- &\cellcolor[HTML]{fbd8ab}18688 &\cellcolor[HTML]{fcdcb1}14130 &\cellcolor[HTML]{fcdcb1}16247 &\cellcolor[HTML]{fcd9ad}16113 &\cellcolor[HTML]{fcdbb0}16279 &\cellcolor[HTML]{fcdcb1}15890 &\cellcolor[HTML]{fcdfb4}16574 &\cellcolor[HTML]{fcdfb4}16180 &\cellcolor[HTML]{fde0b5}12965 \\
$\mathcal{F} \cdot \log\mathcal{R}$ &--- &\cellcolor[HTML]{fbd5a9}18756 &\cellcolor[HTML]{fcdaae}14184 &\cellcolor[HTML]{fcdbaf}16289 &\cellcolor[HTML]{fbd7aa}16160 &\cellcolor[HTML]{fcdaae}16314 &\cellcolor[HTML]{fcdaaf}15926 &\cellcolor[HTML]{fcddb2}16627 &\cellcolor[HTML]{fcddb2}16231 &\cellcolor[HTML]{fcdeb3}13000 \\
\midrule
$\mathcal{F} \cdot \log\left(\mathcal{R} + 1\right)$ &Skip $\mathcal{L} < 4$ &\cellcolor[HTML]{facd9e}18990 &\cellcolor[HTML]{fad1a3}14386 &\cellcolor[HTML]{fbd2a5}16497 &\cellcolor[HTML]{fad0a2}16314 &\cellcolor[HTML]{facfa1}16548 &\cellcolor[HTML]{fad1a3}16140 &\cellcolor[HTML]{fad1a3}16962 &\cellcolor[HTML]{fad0a2}16530 &\cellcolor[HTML]{fbd4a7}13181 \\
$\mathcal{F} \cdot \log\left(\mathcal{R} + 1\right) \cdot \mathcal{L}$ &--- &\cellcolor[HTML]{facd9e}18991 &\cellcolor[HTML]{facd9e}14471 &\cellcolor[HTML]{facc9d}16655 &\cellcolor[HTML]{facc9d}16397 &\cellcolor[HTML]{facc9e}16614 &\cellcolor[HTML]{facd9e}16238 &\cellcolor[HTML]{facc9e}17079 &\cellcolor[HTML]{facc9d}16636 &\cellcolor[HTML]{facc9d}13317 \\
$\mathcal{F} \cdot \log\mathcal{R} \cdot \mathcal{L}$ &--- &\cellcolor[HTML]{facc9e}19004 &\cellcolor[HTML]{facc9e}14481 &\cellcolor[HTML]{facc9d}16663 &\cellcolor[HTML]{f9cb9c}16413 &\cellcolor[HTML]{facc9d}16625 &\cellcolor[HTML]{facc9d}16250 &\cellcolor[HTML]{facc9d}17087 &\cellcolor[HTML]{facc9d}16635 &\cellcolor[HTML]{facc9d}13321 \\
$\mathcal{F} \cdot \log\mathcal{R} \cdot \mathcal{L}$ &Skip $\mathcal{L} = 1$ &\cellcolor[HTML]{f9cb9c}19030 &\cellcolor[HTML]{f9cb9c}14499 &\cellcolor[HTML]{f9cb9c}16669 &\cellcolor[HTML]{facc9d}16404 &\cellcolor[HTML]{f9cb9c}16635 &\cellcolor[HTML]{f9cb9c}16265 &\cellcolor[HTML]{f9cb9c}17103 &\cellcolor[HTML]{f9cb9c}16644 &\cellcolor[HTML]{f9cb9c}13331 \\
\bottomrule
\end{tabular}
\caption{Language coverage for the nine higher-resource languages.}\label{tab:lang-coverage-1}
\end{table*}

\begin{table*}[!t]
\centering
\scriptsize
%\resizebox{ extwidth}{!}{ % use this if the table is too large
\begin{tabular}{lcrrrrrrrrrr}\toprule
\textbf{Priority Criterion} &\textbf{Merge Skip} &\textbf{Kotlin} &\textbf{Scala} &\textbf{Swift} &\textbf{Vue} &\textbf{Dart} &\textbf{Lua} &\textbf{Haskell} &\textbf{Julia} &\textbf{OCaml} \\\midrule
$\mathcal{F}$ (BPE) &--- (BPE) &\cellcolor[HTML]{fff2cc}16898 &\cellcolor[HTML]{fff2cc}7375 &\cellcolor[HTML]{fff2cc}14528 &\cellcolor[HTML]{fff2cc}11475 &\cellcolor[HTML]{fff2cc}15626 &\cellcolor[HTML]{fff2cc}13189 &\cellcolor[HTML]{fff2cc}5565 &\cellcolor[HTML]{fff2cc}6214 &\cellcolor[HTML]{fff2cc}5325 \\
\midrule
$\mathcal{F}$ &Skip $\mathcal{L} < 4$ &\cellcolor[HTML]{fbd7aa}17693 &\cellcolor[HTML]{facfa0}7475 &\cellcolor[HTML]{fbd6a9}15191 &\cellcolor[HTML]{fcdcb1}12006 &\cellcolor[HTML]{fbd7ab}16443 &\cellcolor[HTML]{fbd7ab}13726 &\cellcolor[HTML]{facc9d}5653 &\cellcolor[HTML]{fcddb2}6296 &\cellcolor[HTML]{fbd6a9}5405 \\
$\mathcal{F} \cdot \mathcal{L}$ &--- &\cellcolor[HTML]{fad2a4}17834 &\cellcolor[HTML]{f9cb9c}7484 &\cellcolor[HTML]{facfa1}15353 &\cellcolor[HTML]{fad2a4}12262 &\cellcolor[HTML]{fad0a2}16649 &\cellcolor[HTML]{fad0a2}13875 &\cellcolor[HTML]{facd9f}5650 &\cellcolor[HTML]{face9f}6354 &\cellcolor[HTML]{facd9e}5431 \\
$\mathcal{F} \cdot \mathcal{L}$ &Skip $\mathcal{L} = 1$ &\cellcolor[HTML]{fad0a3}17871 &\cellcolor[HTML]{facc9d}7483 &\cellcolor[HTML]{face9f}15380 &\cellcolor[HTML]{fad0a2}12295 &\cellcolor[HTML]{facfa1}16681 &\cellcolor[HTML]{facea0}13908 &\cellcolor[HTML]{f9cb9c}5654 &\cellcolor[HTML]{facc9d}6360 &\cellcolor[HTML]{f9cb9c}5434 \\
$\mathcal{F} \cdot \log\left(\mathcal{R} + 1\right)$ &--- &\cellcolor[HTML]{fcdbb0}17561 &\cellcolor[HTML]{fbd4a7}7460 &\cellcolor[HTML]{fcdeb3}15005 &\cellcolor[HTML]{fde0b5}11924 &\cellcolor[HTML]{fcdeb3}16229 &\cellcolor[HTML]{fddfb5}13568 &\cellcolor[HTML]{fcdfb4}5610 &\cellcolor[HTML]{fcdeb3}6293 &\cellcolor[HTML]{fcdeb4}5381 \\
$\mathcal{F} \cdot \log\mathcal{R}$ &--- &\cellcolor[HTML]{fcd9ad}17627 &\cellcolor[HTML]{facfa1}7473 &\cellcolor[HTML]{fcdcb1}15053 &\cellcolor[HTML]{fcdeb3}11975 &\cellcolor[HTML]{fcdcb1}16291 &\cellcolor[HTML]{fcddb3}13608 &\cellcolor[HTML]{fcdcb0}5617 &\cellcolor[HTML]{fcdcb1}6298 &\cellcolor[HTML]{fcdcb1}5387 \\
\midrule
$\mathcal{F} \cdot \log\left(\mathcal{R} + 1\right)$ &Skip $\mathcal{L} < 4$ &\cellcolor[HTML]{facea0}17940 &\cellcolor[HTML]{facc9d}7482 &\cellcolor[HTML]{facfa1}15359 &\cellcolor[HTML]{fbd4a7}12203 &\cellcolor[HTML]{facfa0}16685 &\cellcolor[HTML]{fad0a3}13866 &\cellcolor[HTML]{fbd4a7}5634 &\cellcolor[HTML]{fbd5a8}6326 &\cellcolor[HTML]{fad2a4}5417 \\
$\mathcal{F} \cdot \log\left(\mathcal{R} + 1\right) \cdot \mathcal{L}$ &--- &\cellcolor[HTML]{facd9e}17979 &\cellcolor[HTML]{facea0}7476 &\cellcolor[HTML]{facc9e}15413 &\cellcolor[HTML]{facd9e}12374 &\cellcolor[HTML]{facd9e}16742 &\cellcolor[HTML]{facd9e}13940 &\cellcolor[HTML]{facd9f}5650 &\cellcolor[HTML]{facd9e}6357 &\cellcolor[HTML]{facc9d}5432 \\
$\mathcal{F} \cdot \log\mathcal{R} \cdot \mathcal{L}$ &--- &\cellcolor[HTML]{facc9e}17987 &\cellcolor[HTML]{facd9f}7479 &\cellcolor[HTML]{facc9d}15422 &\cellcolor[HTML]{facc9e}12385 &\cellcolor[HTML]{facc9d}16760 &\cellcolor[HTML]{facc9d}13948 &\cellcolor[HTML]{facd9f}5650 &\cellcolor[HTML]{facc9d}6359 &\cellcolor[HTML]{facd9e}5431 \\
$\mathcal{F} \cdot \log\mathcal{R} \cdot \mathcal{L}$ &Skip $\mathcal{L} = 1$ &\cellcolor[HTML]{f9cb9c}18011 &\cellcolor[HTML]{facd9e}7481 &\cellcolor[HTML]{f9cb9c}15435 &\cellcolor[HTML]{f9cb9c}12407 &\cellcolor[HTML]{f9cb9c}16781 &\cellcolor[HTML]{f9cb9c}13963 &\cellcolor[HTML]{face9f}5649 &\cellcolor[HTML]{f9cb9c}6362 &\cellcolor[HTML]{facc9d}5432 \\
\bottomrule
\end{tabular}
\caption{Language coverage for the nine lower-resource languages.}\label{tab:lang-coverage-2}
\end{table*}

\begin{table*}[!t]
\centering
\scriptsize
%\resizebox{ extwidth}{!}{ % use this if the table is too large
\begin{tabular}{rlcrrrrrrrrrr}\toprule
&\textbf{Priority Criterion} &\textbf{Merge Skip} &\textbf{1} &\textbf{2} &\textbf{3} &\textbf{4} &\textbf{5} &\textbf{6} &\textbf{7} &\textbf{8} &\textbf{9} &\textbf{10} \\\midrule
\textbf{1} &$\mathcal{F}$ (BPE) &--- (BPE) &0 &\cellcolor[HTML]{ffde83}8304 &\cellcolor[HTML]{ffdb76}9168 &\cellcolor[HTML]{ffda73}9386 &\cellcolor[HTML]{ffe8a8}5848 &\cellcolor[HTML]{ffe6a0}6362 &\cellcolor[HTML]{ffd970}9580 &\cellcolor[HTML]{ffd76a}10004 &\cellcolor[HTML]{ffd769}10088 &\cellcolor[HTML]{ffd666}10228 \\
\textbf{2} &$\mathcal{F}$ &$\mathcal{L} < 4$ &\cellcolor[HTML]{ffde83}8304 &0 &\cellcolor[HTML]{ffe9ac}5572 &\cellcolor[HTML]{ffeaaf}5392 &\cellcolor[HTML]{ffe7a4}6132 &\cellcolor[HTML]{ffe8a8}5878 &\cellcolor[HTML]{fff2ce}3292 &\cellcolor[HTML]{ffe69f}6434 &\cellcolor[HTML]{ffe69f}6472 &\cellcolor[HTML]{ffe6a0}6392 \\
\textbf{3} &$\mathcal{F} \cdot \mathcal{L}$ &--- &\cellcolor[HTML]{ffdb76}9168 &\cellcolor[HTML]{ffe9ac}5572 &0 &\cellcolor[HTML]{fffefa}372 &\cellcolor[HTML]{ffe6a0}6366 &\cellcolor[HTML]{ffe6a1}6316 &\cellcolor[HTML]{ffebb5}5006 &\cellcolor[HTML]{fff5d8}2666 &\cellcolor[HTML]{fff4d5}2826 &\cellcolor[HTML]{fff4d4}2894 \\
\textbf{4} &$\mathcal{F} \cdot \mathcal{L}$ &$\mathcal{L} = 1$ &\cellcolor[HTML]{ffda73}9386 &\cellcolor[HTML]{ffeaaf}5392 &\cellcolor[HTML]{fffefa}372 &0 &\cellcolor[HTML]{ffe59d}6562 &\cellcolor[HTML]{ffe6a1}6346 &\cellcolor[HTML]{ffecb7}4842 &\cellcolor[HTML]{fff5d8}2614 &\cellcolor[HTML]{fff4d6}2764 &\cellcolor[HTML]{fff5d7}2678 \\
\textbf{5} &$\mathcal{F} \cdot \log(\mathcal{R} + 1)$ &--- &\cellcolor[HTML]{ffe8a8}5848 &\cellcolor[HTML]{ffe7a4}6132 &\cellcolor[HTML]{ffe6a0}6366 &\cellcolor[HTML]{ffe59d}6562 &0 &\cellcolor[HTML]{fffdf7}576 &\cellcolor[HTML]{ffecb7}4834 &\cellcolor[HTML]{ffe8a7}5892 &\cellcolor[HTML]{ffe8a6}5956 &\cellcolor[HTML]{ffe7a4}6086 \\
\textbf{6} &$\mathcal{F} \cdot \log\mathcal{R}$ &--- &\cellcolor[HTML]{ffe6a0}6362 &\cellcolor[HTML]{ffe8a8}5878 &\cellcolor[HTML]{ffe6a1}6316 &\cellcolor[HTML]{ffe6a1}6346 &\cellcolor[HTML]{fffdf7}576 &0 &\cellcolor[HTML]{ffeebd}4454 &\cellcolor[HTML]{ffe9ac}5590 &\cellcolor[HTML]{ffe9ab}5618 &\cellcolor[HTML]{ffe9aa}5728 \\
\textbf{7} &$\mathcal{F} \cdot \log(\mathcal{R} + 1)$ &$\mathcal{L} < 4$ &\cellcolor[HTML]{ffd970}9580 &\cellcolor[HTML]{fff2ce}3292 &\cellcolor[HTML]{ffebb5}5006 &\cellcolor[HTML]{ffecb7}4842 &\cellcolor[HTML]{ffecb7}4834 &\cellcolor[HTML]{ffeebd}4454 &0 &\cellcolor[HTML]{ffeebe}4368 &\cellcolor[HTML]{ffeebe}4352 &\cellcolor[HTML]{ffeebf}4282 \\
\textbf{8} &$\mathcal{F} \cdot \log(\mathcal{R} + 1) \cdot \mathcal{L}$ &--- &\cellcolor[HTML]{ffd76a}10004 &\cellcolor[HTML]{ffe69f}6434 &\cellcolor[HTML]{fff5d8}2666 &\cellcolor[HTML]{fff5d8}2614 &\cellcolor[HTML]{ffe8a7}5892 &\cellcolor[HTML]{ffe9ac}5590 &\cellcolor[HTML]{ffeebe}4368 &0 &\cellcolor[HTML]{fffffc}216 &\cellcolor[HTML]{fffefb}332 \\
\textbf{9} &$\mathcal{F} \cdot \log\mathcal{R} \cdot \mathcal{L}$ &--- &\cellcolor[HTML]{ffd769}10088 &\cellcolor[HTML]{ffe69f}6472 &\cellcolor[HTML]{fff4d5}2826 &\cellcolor[HTML]{fff4d6}2764 &\cellcolor[HTML]{ffe8a6}5956 &\cellcolor[HTML]{ffe9ab}5618 &\cellcolor[HTML]{ffeebe}4352 &\cellcolor[HTML]{fffffc}216 &0 &\cellcolor[HTML]{fffffd}200 \\
\textbf{10} &$\mathcal{F} \cdot \log\mathcal{R} \cdot \mathcal{L}$ &$\mathcal{L} = 1$ &\cellcolor[HTML]{ffd666}10228 &\cellcolor[HTML]{ffe6a0}6392 &\cellcolor[HTML]{fff4d4}2894 &\cellcolor[HTML]{fff5d7}2678 &\cellcolor[HTML]{ffe7a4}6086 &\cellcolor[HTML]{ffe9aa}5728 &\cellcolor[HTML]{ffeebf}4282 &\cellcolor[HTML]{fffefb}332 &\cellcolor[HTML]{fffffd}200 &0 \\
\bottomrule
\end{tabular}
\caption{Counts of different tokens in the trained tokenizers. The columns with numbers represent the models in the same order, enumerated by identifiers in the first column.}\label{tab:token-diff}
\end{table*}

\begin{table*}[!t]\centering
\small
%\resizebox{ extwidth}{!}{ % use this if the table is too large
\begin{tabular}{lcc}\toprule
\textbf{Priority Criterion} &\textbf{Merge Skip} &\textbf{Under-trained token examples} \\\midrule
$\mathcal{F}$ (BPE) &--- (BPE) & \texttt{0123351271903}, \texttt{\_AtaSmartAttributeDisplayType}, \texttt{minCuSize}, \texttt{DISKinematics} \\
\midrule
$\mathcal{F}$ &Skip $\mathcal{L} < 4$ & \texttt{IMARY}, \texttt{ientation}, \texttt{vanced}, \texttt{pendencies}, \texttt{ventory} \\
$\mathcal{F} \cdot \mathcal{L}$ &--- & \texttt{OfSurgeServices}, \texttt{AceHighFlush}, \texttt{PeriodoDeclara}, \texttt{OXWSMTGS}  \\
$\mathcal{F} \cdot \mathcal{L}$ &Skip $\mathcal{L} = 1$ & \texttt{arency}, \texttt{ICENSE}, \texttt{ployment}, \texttt{ereum}, \texttt{avascript} \\
$\mathcal{F} \cdot \log\left(\mathcal{R} + 1\right)$ &--- & \texttt{gridBagConstraintsPeriodoDeclara}, \texttt{HighFlush}, \texttt{ablytyped} \\
$\mathcal{F} \cdot \log\mathcal{R}$ &--- & \texttt{IntoConstraints}, \texttt{scalably}, \texttt{ilibj}, \texttt{VisualStyleBackColor}, \texttt{arency} \\
\midrule
$\mathcal{F} \cdot \log\left(\mathcal{R} + 1\right)$ &Skip $\mathcal{L} < 4$ & \texttt{AWSCloud}, \texttt{crets}, \texttt{ooser}, \texttt{NECTION}, \texttt{ereum} \\
$\mathcal{F} \cdot \log\left(\mathcal{R} + 1\right) \cdot \mathcal{L}$ &--- & \texttt{\_HandRankName}, \texttt{chordie}, \texttt{TINGS}, \texttt{ramimages}, \texttt{ekUSDI} \\
$\mathcal{F} \cdot \log\mathcal{R} \cdot \mathcal{L}$ &--- & \texttt{scriptors}, \texttt{TEGER}, \texttt{versation}, \texttt{FTWARE}, \texttt{clare} \\
$\mathcal{F} \cdot \log\mathcal{R} \cdot \mathcal{L}$ &Skip $\mathcal{L} = 1$ & \texttt{FTWARE}, \texttt{NECTION}, \texttt{IMARY}, \texttt{trieve}, \texttt{elcome} \\
\bottomrule
\end{tabular}
\caption{Examples of verified under-trained tokens in multilingual models.}\label{tab:under-trained-examples}
\end{table*}

\section{Unigram and Wordpiece}
\label{app:unigram_wordpiece}

We separately train Wordpiece and Unigram tokenizers on the same multilingual data sample and compare coverage and frequency distributions in Figure~\ref{fig:unigram-wordpiece}. BPE shows superior coverage, and its frequency distribution is substantially higher than that of the other tokenizers. By manual inspection, the Unigram tokenizer contains more than 5000 tokens as single Chinese characters or emojis, while the Wordpiece tokenizer contains many overly long variable names, e.g., ``\texttt{testTypeConstantsDefinedRistekUSDI\_FHIR \_R4B\_FHIRResource\_FHIR}'' as a single token. Based on these results, we excluded these methods from the analysis and compared only the BPE-based algorithms.

\section{Multilingual Models}
\label{app:multilingual}

In Tables~\ref{tab:lang-compression-1} and \ref{tab:lang-compression-2}, we present the compression rates by language for BPE and our modified models. In Tables~\ref{tab:lang-coverage-1} and \ref{tab:lang-coverage-2}, we present language coverage in the same manner. Modifications that include a language component lead to better per-language scores. This effect becomes more pronounced in lower-resource languages, suggesting that these modifications provide better support for them. This is most evident in compression rates for modifications that include $\mathcal{L} < 4$: for higher-resource languages, these models exhibit inferior compression, whereas they achieve better scores for most lower-resource languages. We hypothesize that this modification leads to stronger regularization of the language component.

In Table~\ref{tab:token-diff}, we present a pairwise vocabulary comparison of tokenizers. The tokenizers that combine both language and repository components show the largest difference from BPE, with more than 10,000 tokens. Similar tokenizer modifications naturally lead to similar vocabularies and have smaller differences. Thus, there is no clearly best modification, as we discuss in Section~\ref{sec:discussion}.

\section{Under-Trained Tokens}
\label{app:tokens}

In Table~\ref{tab:under-trained-examples}, we show representative examples of under-trained tokens in the multilingual models. The criteria that allow single-language or single-repository still contain variable names among the under-trained tokens. The other criteria are better regularized and contain predominantly intermediate tokens among the under-trained tokens. We show in Section~\ref{sec:results} that the number of such tokens is low when the general overfitting is reduced.

\section{Pruning and Compression}
\label{app:pruning}

\begin{figure*}[t!]
    \begin{subfigure}[b]{0.48\textwidth}
        \includegraphics[width=\textwidth]{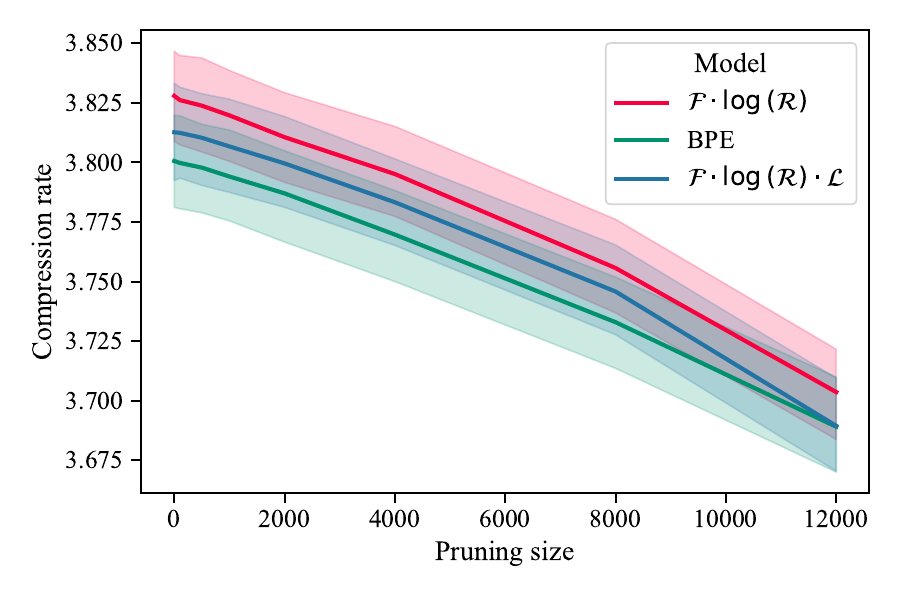}
        \subcaption{Pruning order: reverse merge order.}
        \label{fig:naive_Java}
    \end{subfigure}
    \hfill
    \begin{subfigure}[b]{0.48\textwidth}
        \includegraphics[width=\textwidth]{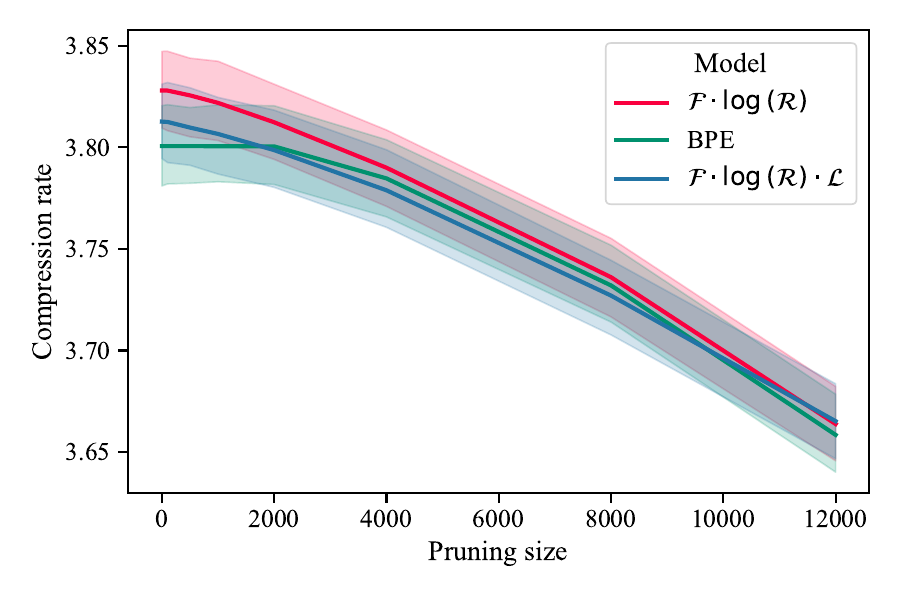}
        \subcaption{Pruning order: under-trained first.}
        \label{fig:scores_Java}
    \end{subfigure}
    \caption{Compression rate for Java for tokenizers with applied pruning \textbf{(a)} in the reverse order of token ids and \textbf{(b)} starting from the tokens with the lowest under-trained indicator values (distance from (0, 0) in indicator space).}
    \label{fig:pruning_Java}
\end{figure*}

\begin{figure*}[t!]
    \begin{subfigure}[b]{0.48\textwidth}
        \includegraphics[width=\textwidth]{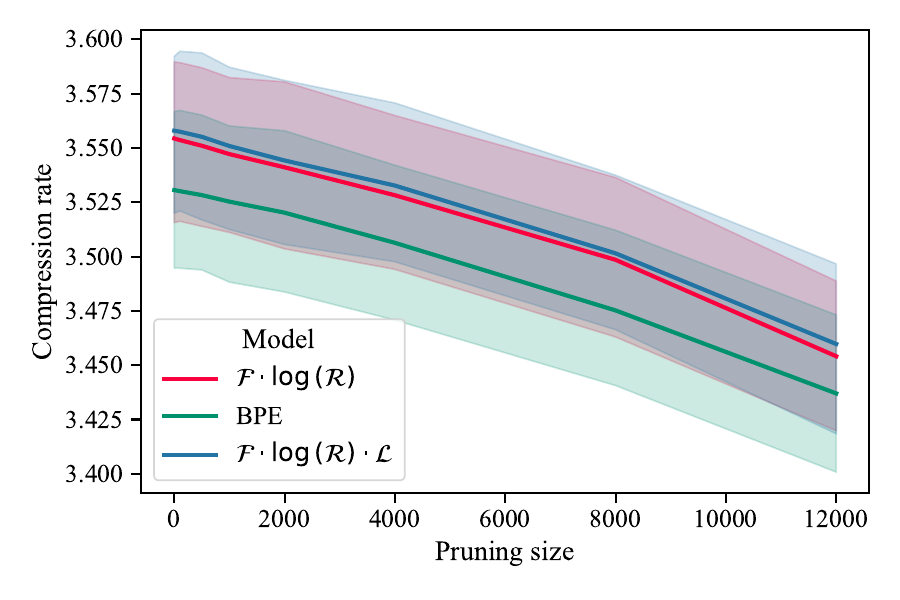}
        \subcaption{Pruning order: reverse merge order.}
        \label{fig:naive_JavaScript}
    \end{subfigure}
    \hfill
    \begin{subfigure}[b]{0.48\textwidth}
        \includegraphics[width=\textwidth]{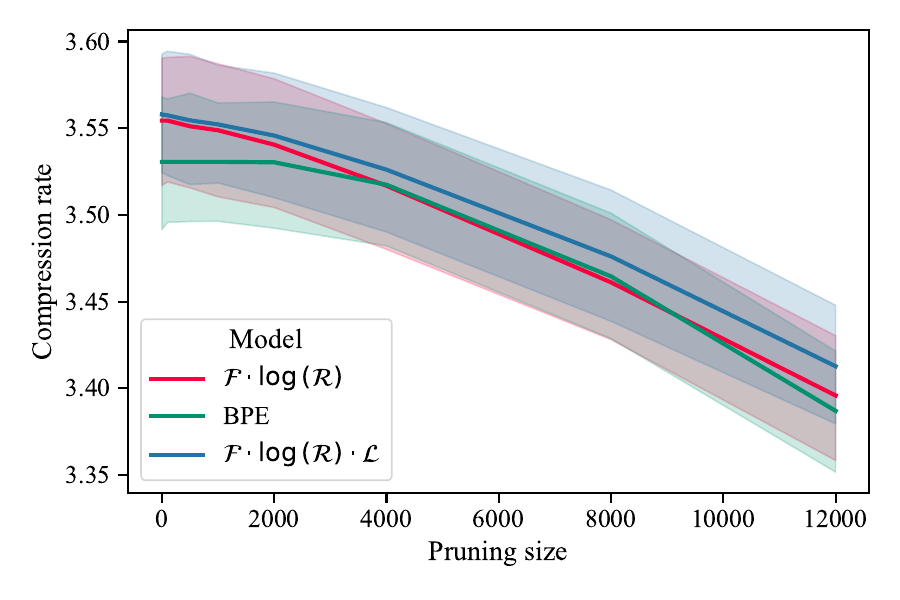}
        \subcaption{Pruning order: under-trained first.}
        \label{fig:scores_JavaScript}
    \end{subfigure}
    \caption{Compression rate for JavaScript for tokenizers with applied pruning \textbf{(a)} in the reverse order of token ids and \textbf{(b)} starting from the tokens with the lowest under-trained indicator values (distance from (0, 0) in indicator space).}
    \label{fig:pruning_JavaScript}
\end{figure*}

\begin{figure*}[t!]
    \begin{subfigure}[b]{0.48\textwidth}
        \includegraphics[width=\textwidth]{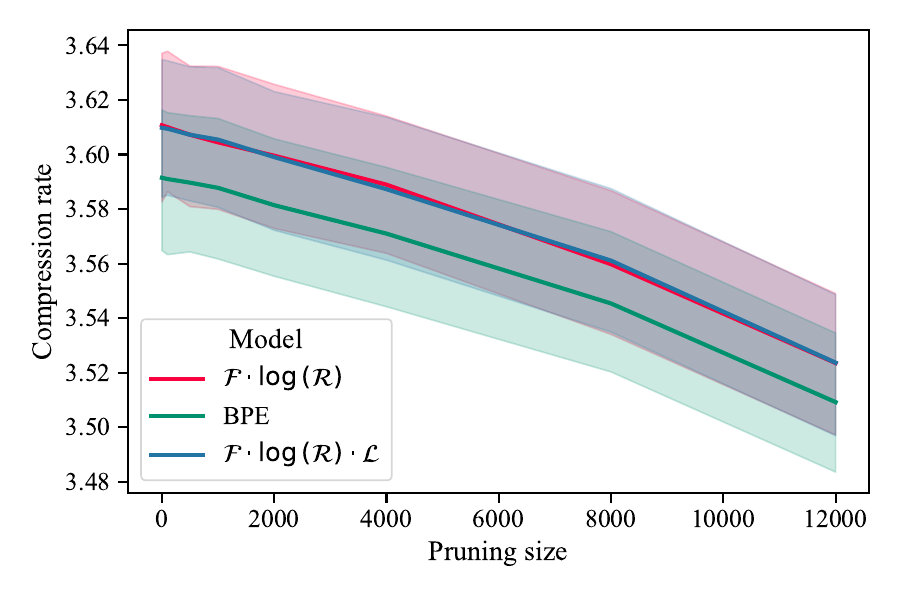}
        \subcaption{Pruning order: reverse merge order.}
        \label{fig:naive_Python}
    \end{subfigure}
    \hfill
    \begin{subfigure}[b]{0.48\textwidth}
        \includegraphics[width=\textwidth]{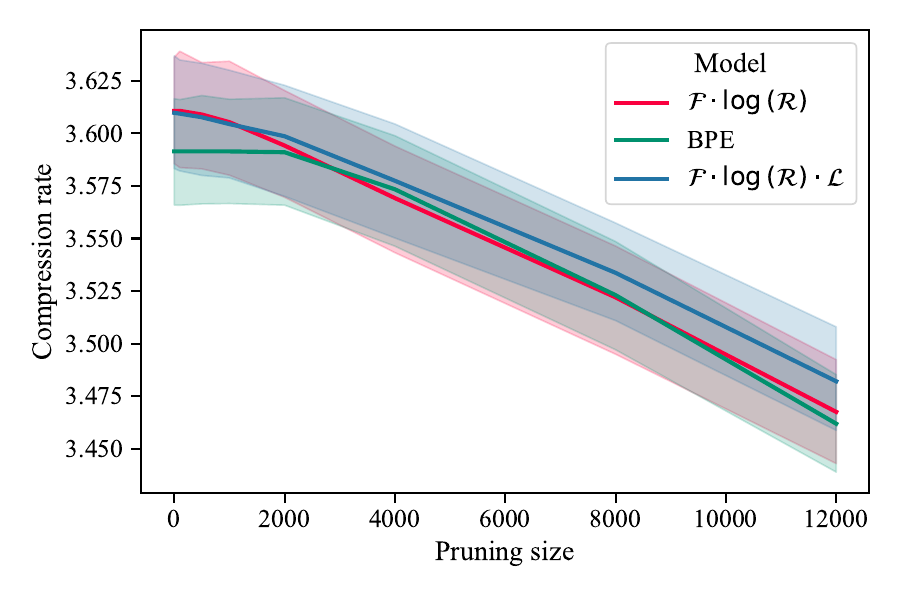}
        \subcaption{Pruning order: under-trained first.}
        \label{fig:scores_Python}
    \end{subfigure}
    \caption{Compression rate for Python for tokenizers with applied pruning \textbf{(a)} in the reverse order of token ids and \textbf{(b)} starting from the tokens with the lowest under-trained indicator values (distance from (0, 0) in indicator space).}
    \label{fig:pruning_Python}
\end{figure*}

\begin{figure*}[t!]
    \begin{subfigure}[b]{0.48\textwidth}
        \includegraphics[width=\textwidth]{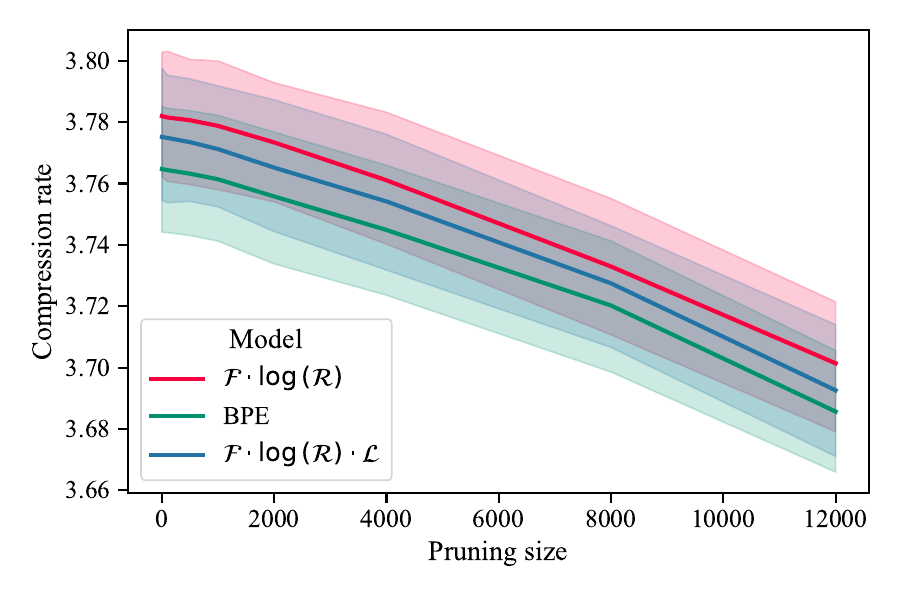}
        \subcaption{Pruning order: reverse merge order.}
        \label{fig:naive_PHP}
    \end{subfigure}
    \hfill
    \begin{subfigure}[b]{0.48\textwidth}
        \includegraphics[width=\textwidth]{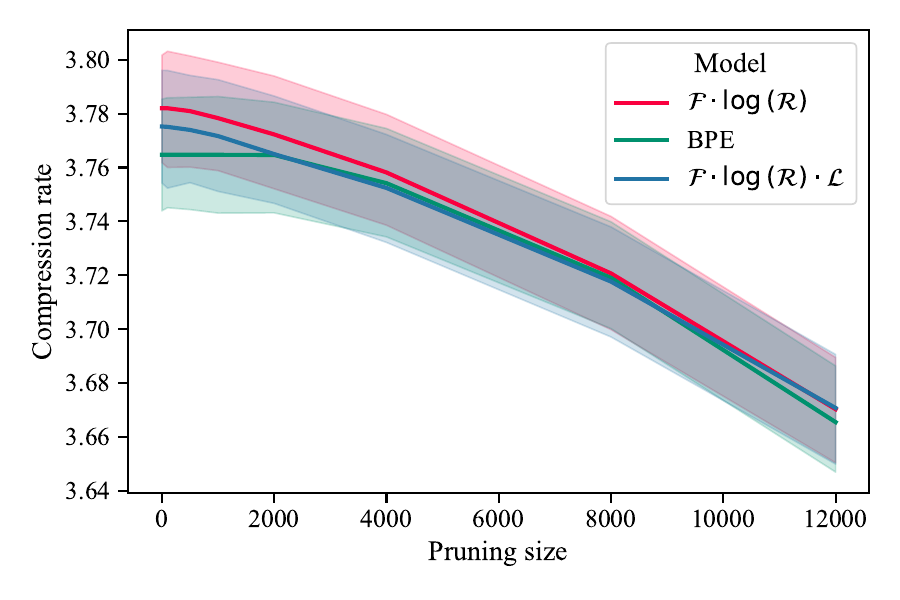}
        \subcaption{Pruning order: under-trained first.}
        \label{fig:scores_PHP}
    \end{subfigure}
    \caption{Compression rate for PHP for tokenizers with applied pruning \textbf{(a)} in the reverse order of token ids and \textbf{(b)} starting from the tokens with the lowest under-trained indicator values (distance from (0, 0) in indicator space).}
    \label{fig:pruning_PHP}
\end{figure*}

\begin{figure*}[t!]
    \begin{subfigure}[b]{0.48\textwidth}
        \includegraphics[width=\textwidth]{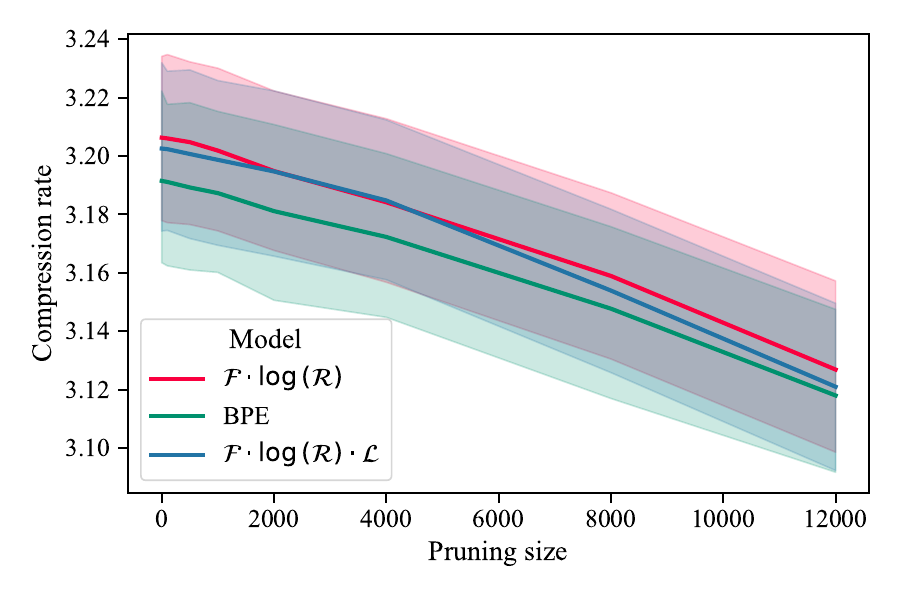}
        \subcaption{Pruning order: reverse merge order.}
        \label{fig:naive_C++}
    \end{subfigure}
    \hfill
    \begin{subfigure}[b]{0.48\textwidth}
        \includegraphics[width=\textwidth]{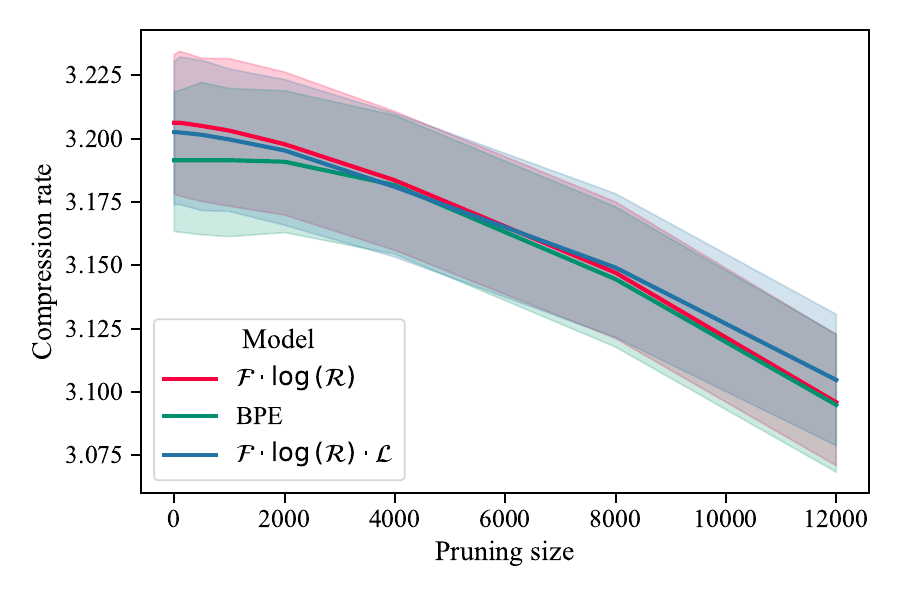}
        \subcaption{Pruning order: under-trained first.}
        \label{fig:scores_C++}
    \end{subfigure}
    \caption{Compression rate for C++ for tokenizers with applied pruning \textbf{(a)} in the reverse order of token ids and \textbf{(b)} starting from the tokens with the lowest under-trained indicator values (distance from (0, 0) in indicator space).}
    \label{fig:pruning_C++}
\end{figure*}

\begin{figure*}[t!]
    \begin{subfigure}[b]{0.48\textwidth}
        \includegraphics[width=\textwidth]{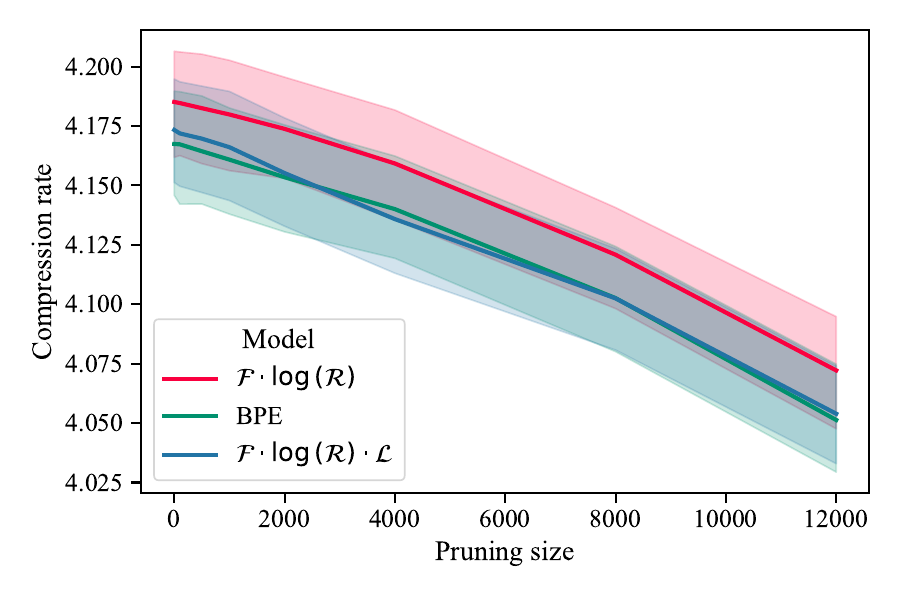}
        \subcaption{Pruning order: reverse merge order.}
        \label{fig:naive_Csharp}
    \end{subfigure}
    \hfill
    \begin{subfigure}[b]{0.48\textwidth}
        \includegraphics[width=\textwidth]{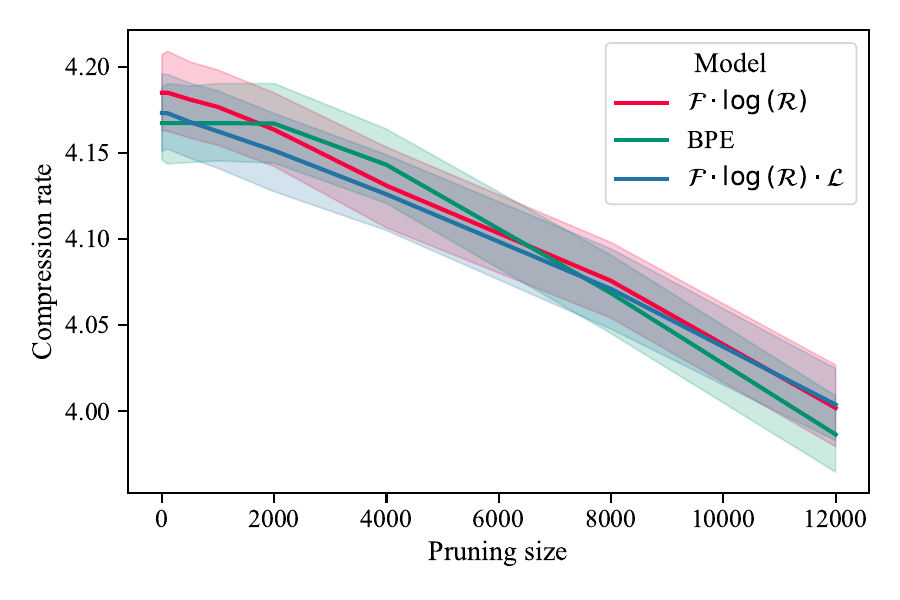}
        \subcaption{Pruning order: under-trained first.}
        \label{fig:scores_Csharp}
    \end{subfigure}
    \caption{Compression rate for C\# for tokenizers with applied pruning \textbf{(a)} in the reverse order of token ids and \textbf{(b)} starting from the tokens with the lowest under-trained indicator values (distance from (0, 0) in indicator space).}
    \label{fig:pruning_Csharp}
\end{figure*}

\begin{figure*}[t!]
    \begin{subfigure}[b]{0.48\textwidth}
        \includegraphics[width=\textwidth]{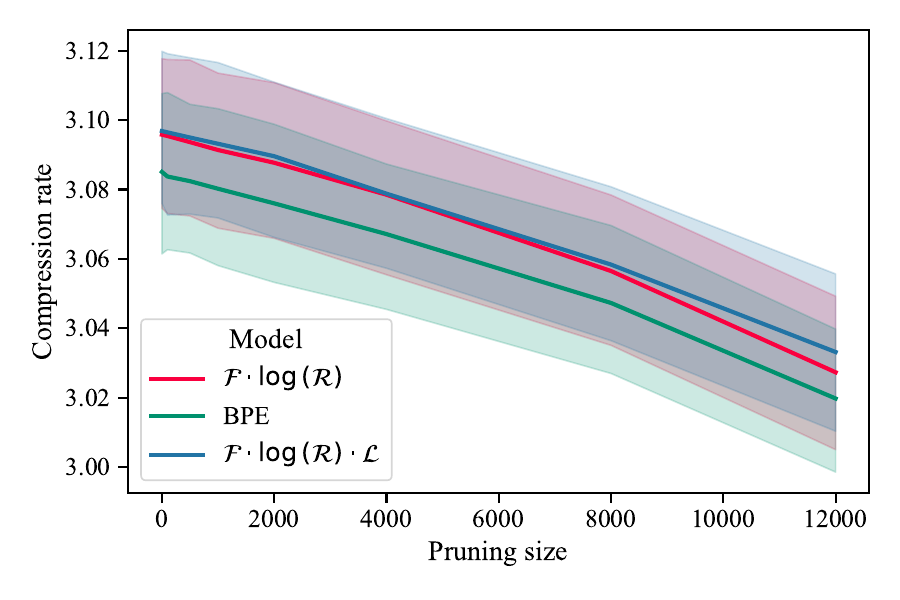}
        \subcaption{Pruning order: reverse merge order.}
        \label{fig:naive_Go}
    \end{subfigure}
    \hfill
    \begin{subfigure}[b]{0.48\textwidth}
        \includegraphics[width=\textwidth]{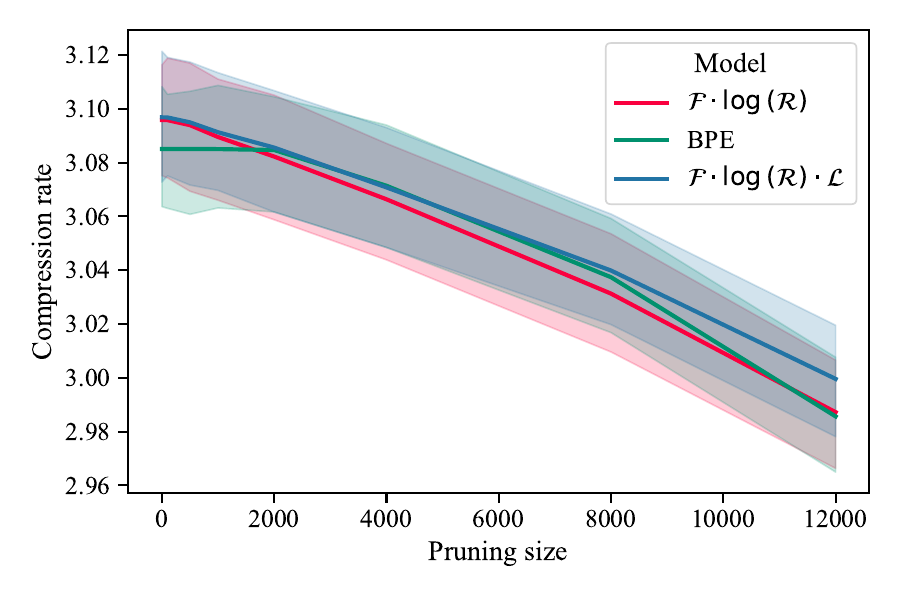}
        \subcaption{Pruning order: under-trained first.}
        \label{fig:scores_Go}
    \end{subfigure}
    \caption{Compression rate for Go for tokenizers with applied pruning \textbf{(a)} in the reverse order of token ids and \textbf{(b)} starting from the tokens with the lowest under-trained indicator values (distance from (0, 0) in indicator space).}
    \label{fig:pruning_Go}
\end{figure*}

\begin{figure*}[t!]
    \begin{subfigure}[b]{0.48\textwidth}
        \includegraphics[width=\textwidth]{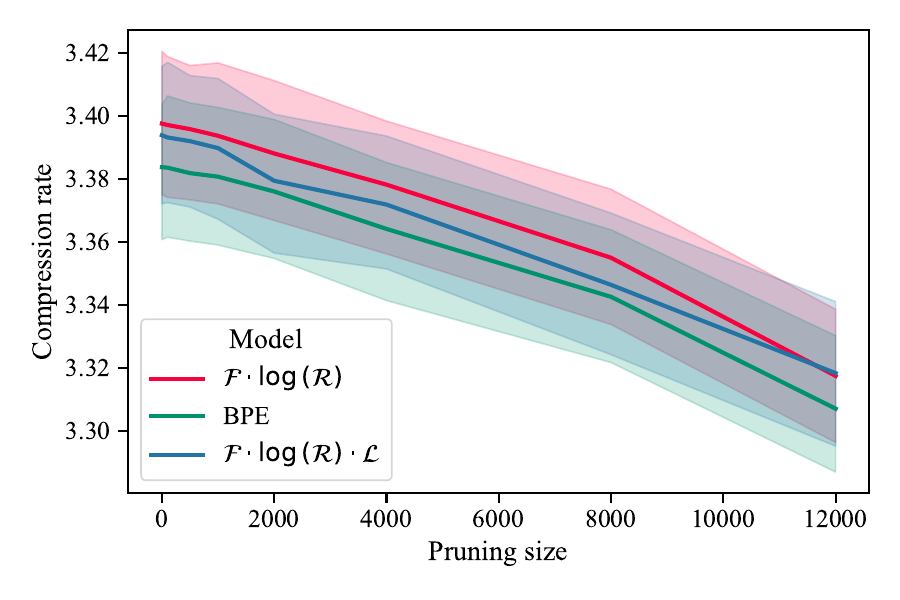}
        \subcaption{Pruning order: reverse merge order.}
        \label{fig:naive_Rust}
    \end{subfigure}
    \hfill
    \begin{subfigure}[b]{0.48\textwidth}
        \includegraphics[width=\textwidth]{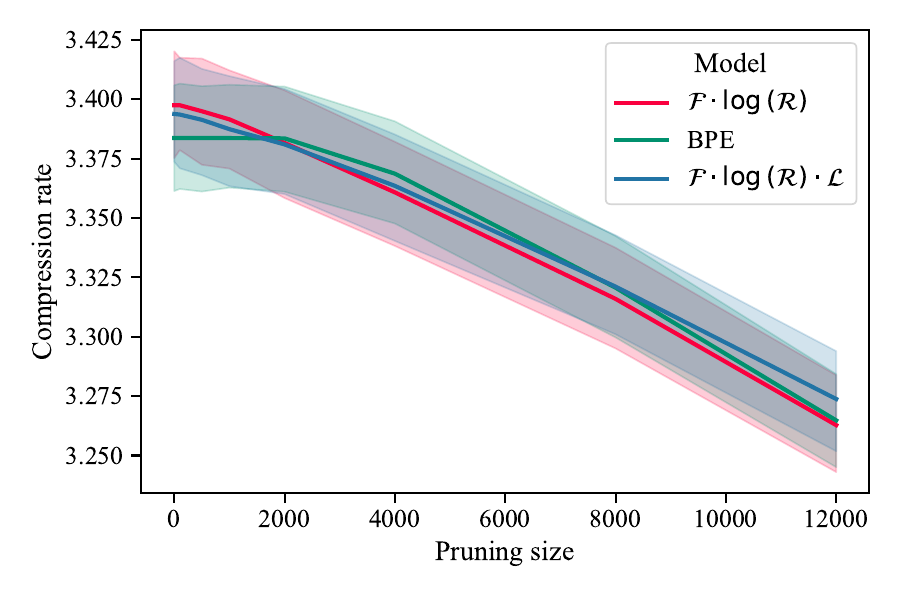}
        \subcaption{Pruning order: under-trained first.}
        \label{fig:scores_Rust}
    \end{subfigure}
    \caption{Compression rate for Rust for tokenizers with applied pruning \textbf{(a)} in the reverse order of token ids and \textbf{(b)} starting from the tokens with the lowest under-trained indicator values (distance from (0, 0) in indicator space).}
    \label{fig:pruning_Rust}
\end{figure*}

\begin{figure*}[t!]
    \begin{subfigure}[b]{0.48\textwidth}
        \includegraphics[width=\textwidth]{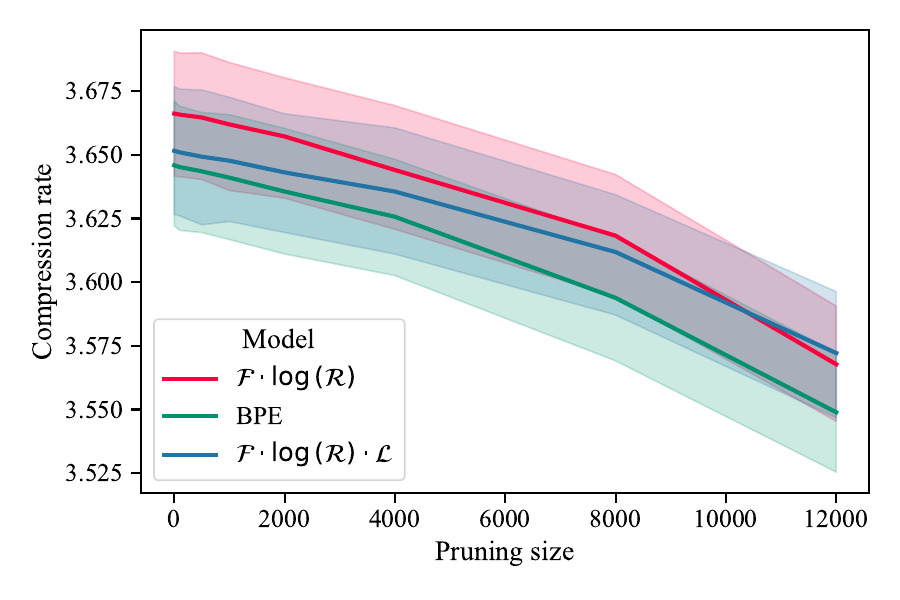}
        \subcaption{Pruning order: reverse merge order.}
        \label{fig:naive_Ruby}
    \end{subfigure}
    \hfill
    \begin{subfigure}[b]{0.48\textwidth}
        \includegraphics[width=\textwidth]{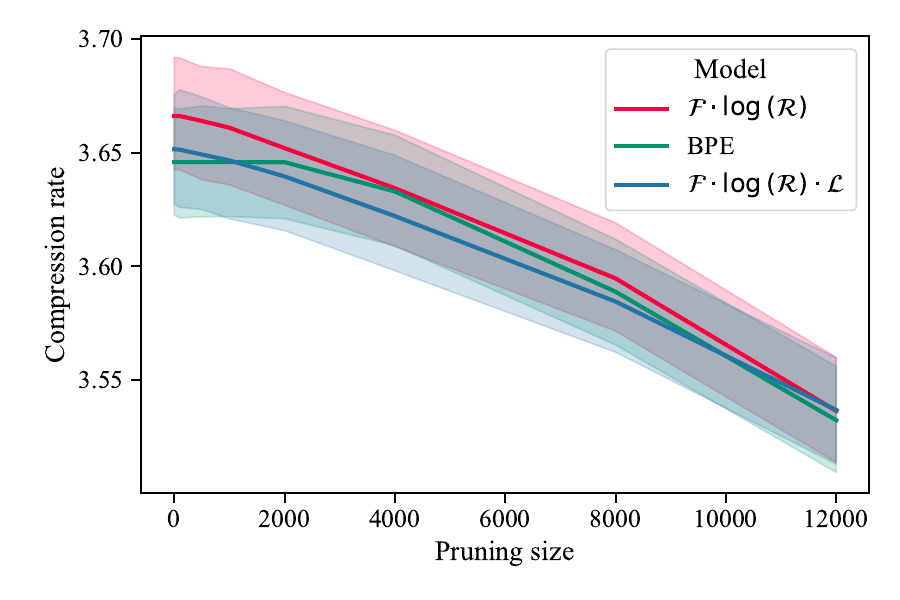}
        \subcaption{Pruning order: under-trained first.}
        \label{fig:scores_Ruby}
    \end{subfigure}
    \caption{Compression rate for Ruby for tokenizers with applied pruning \textbf{(a)} in the reverse order of token ids and \textbf{(b)} starting from the tokens with the lowest under-trained indicator values (distance from (0, 0) in indicator space).}
    \label{fig:pruning_Ruby}
\end{figure*}

\begin{figure*}[t!]
    \begin{subfigure}[b]{0.48\textwidth}
        \includegraphics[width=\textwidth]{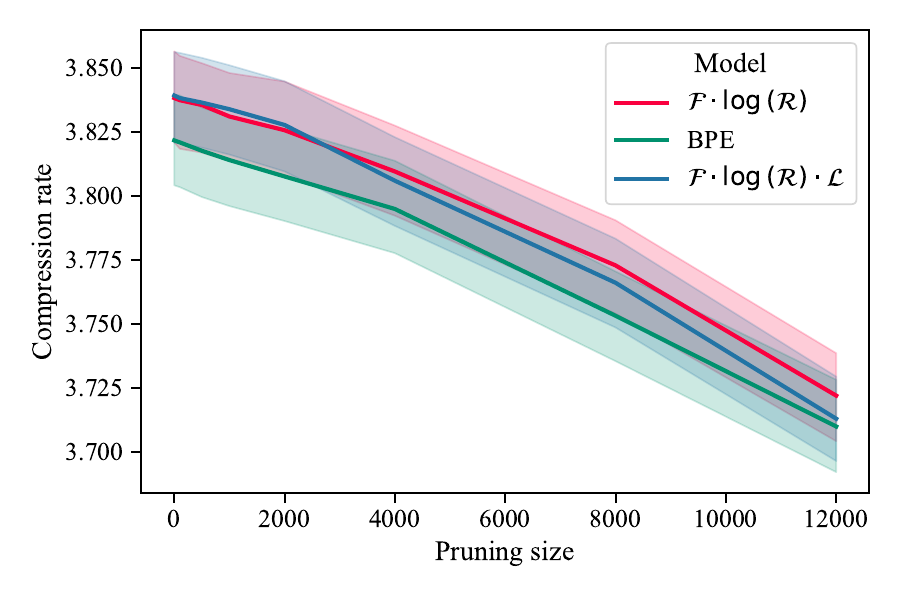}
        \subcaption{Pruning order: reverse merge order.}
        \label{fig:naive_Kotlin}
    \end{subfigure}
    \hfill
    \begin{subfigure}[b]{0.48\textwidth}
        \includegraphics[width=\textwidth]{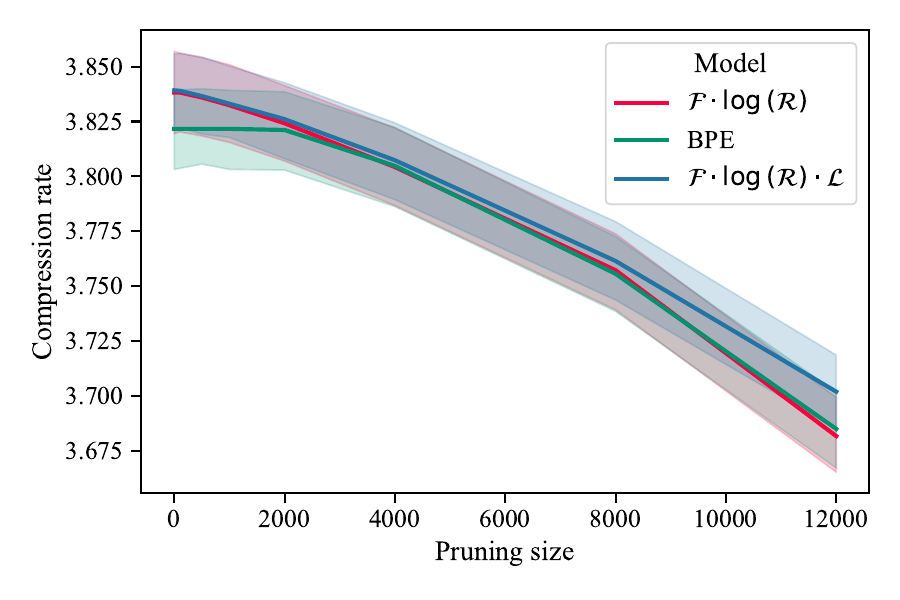}
        \subcaption{Pruning order: under-trained first.}
        \label{fig:scores_Kotlin}
    \end{subfigure}
    \caption{Compression rate for Kotlin for tokenizers with applied pruning \textbf{(a)} in the reverse order of token ids and \textbf{(b)} starting from the tokens with the lowest under-trained indicator values (distance from (0, 0) in indicator space).}
    \label{fig:pruning_Kotlin}
\end{figure*}

\begin{figure*}[t!]
    \begin{subfigure}[b]{0.48\textwidth}
        \includegraphics[width=\textwidth]{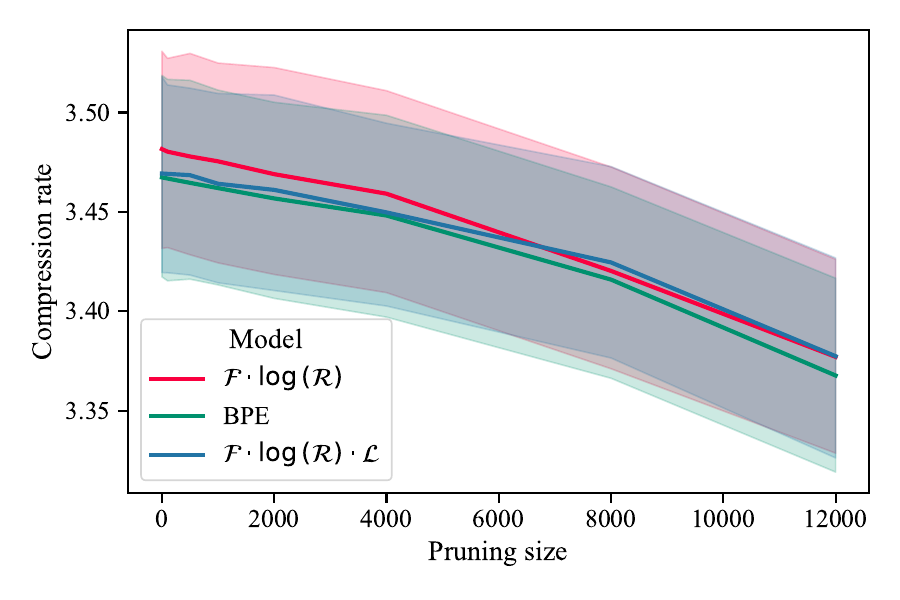}
        \subcaption{Pruning order: reverse merge order.}
        \label{fig:naive_Scala}
    \end{subfigure}
    \hfill
    \begin{subfigure}[b]{0.48\textwidth}
        \includegraphics[width=\textwidth]{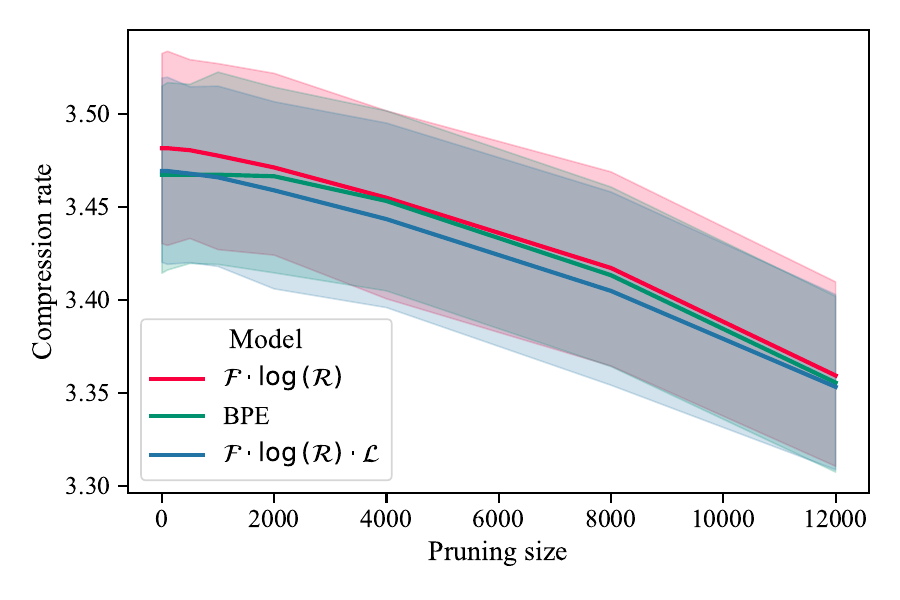}
        \subcaption{Pruning order: under-trained first.}
        \label{fig:scores_Scala}
    \end{subfigure}
    \caption{Compression rate for Scala for tokenizers with applied pruning \textbf{(a)} in the reverse order of token ids and \textbf{(b)} starting from the tokens with the lowest under-trained indicator values (distance from (0, 0) in indicator space).}
    \label{fig:pruning_Scala}
\end{figure*}

\begin{figure*}[t!]
    \begin{subfigure}[b]{0.48\textwidth}
        \includegraphics[width=\textwidth]{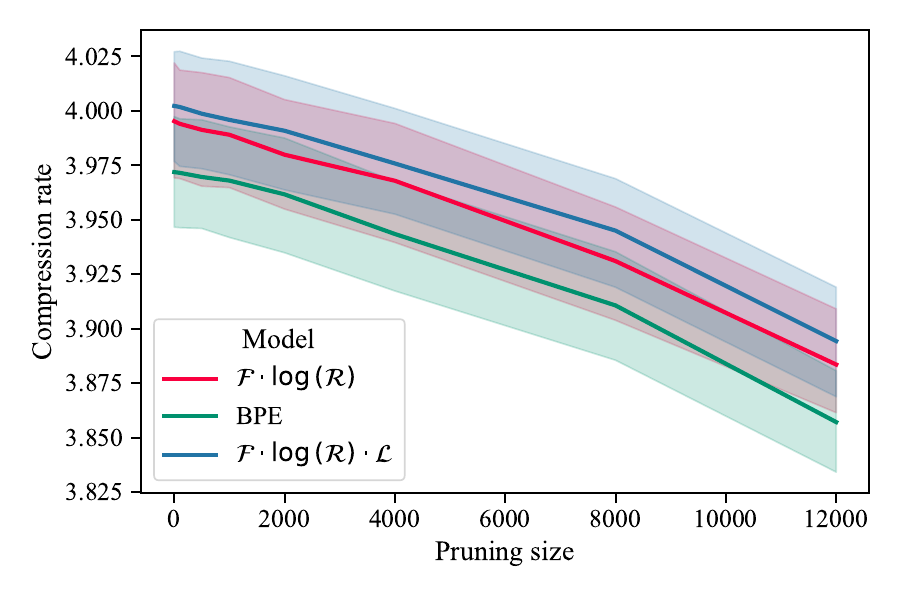}
        \subcaption{Pruning order: reverse merge order.}
        \label{fig:naive_Swift}
    \end{subfigure}
    \hfill
    \begin{subfigure}[b]{0.48\textwidth}
        \includegraphics[width=\textwidth]{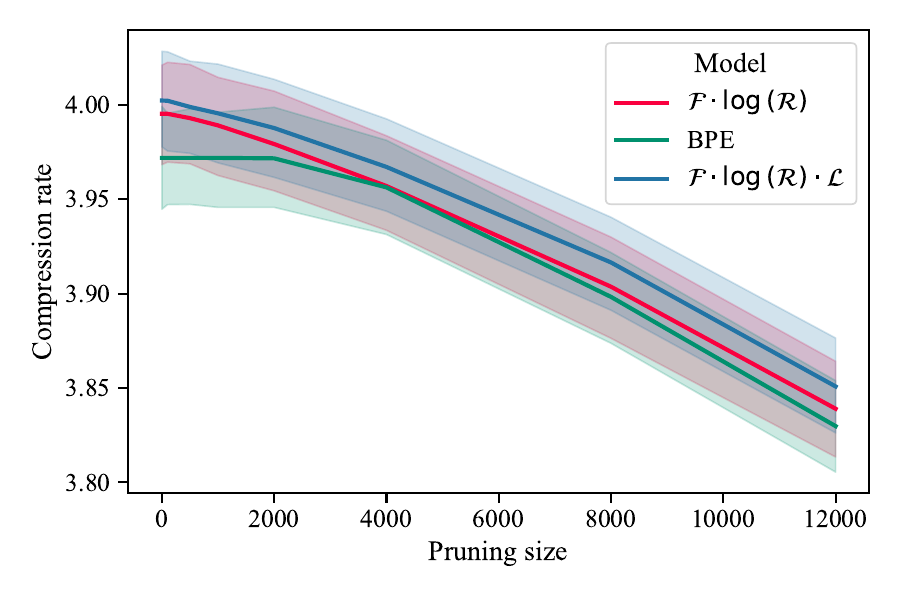}
        \subcaption{Pruning order: under-trained first.}
        \label{fig:scores_Swift}
    \end{subfigure}
    \caption{Compression rate for Swift for tokenizers with applied pruning \textbf{(a)} in the reverse order of token ids and \textbf{(b)} starting from the tokens with the lowest under-trained indicator values (distance from (0, 0) in indicator space).}
    \label{fig:pruning_Swift}
\end{figure*}

\begin{figure*}[t!]
    \begin{subfigure}[b]{0.48\textwidth}
        \includegraphics[width=\textwidth]{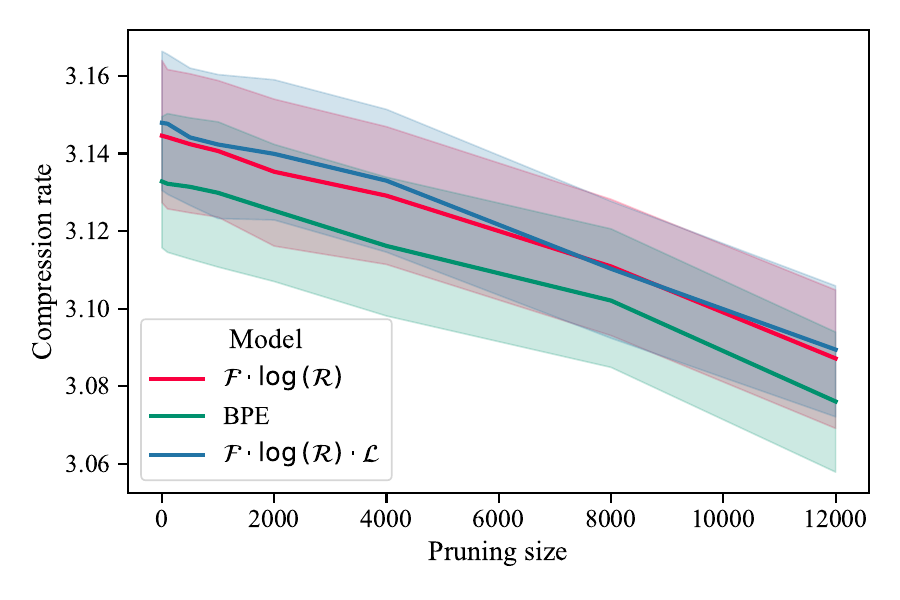}
        \subcaption{Pruning order: reverse merge order.}
        \label{fig:naive_Vue}
    \end{subfigure}
    \hfill
    \begin{subfigure}[b]{0.48\textwidth}
        \includegraphics[width=\textwidth]{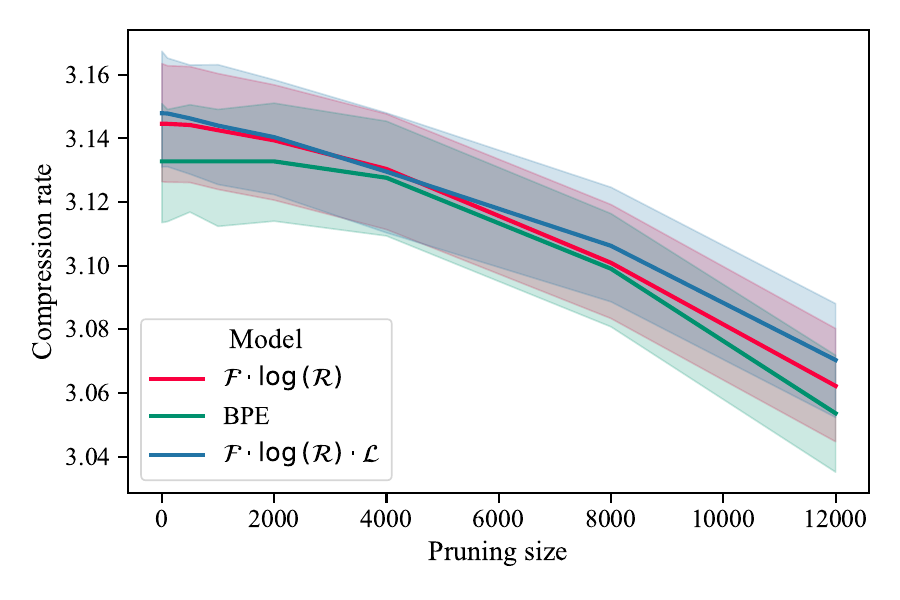}
        \subcaption{Pruning order: under-trained first.}
        \label{fig:scores_Vue}
    \end{subfigure}
    \caption{Compression rate for Vue for tokenizers with applied pruning \textbf{(a)} in the reverse order of token ids and \textbf{(b)} starting from the tokens with the lowest under-trained indicator values (distance from (0, 0) in indicator space).}
    \label{fig:pruning_Vue}
\end{figure*}

\begin{figure*}[t!]
    \begin{subfigure}[b]{0.48\textwidth}
        \includegraphics[width=\textwidth]{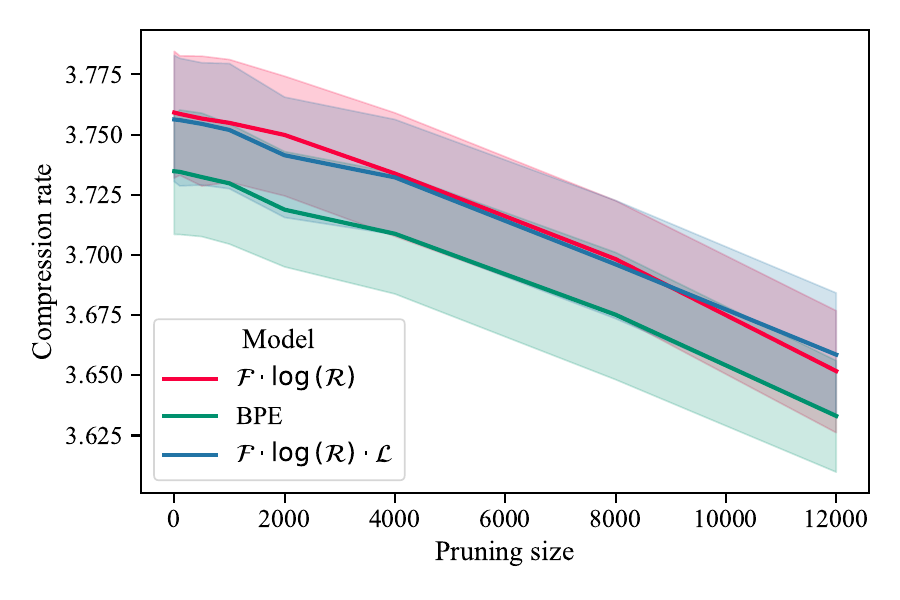}
        \subcaption{Pruning order: reverse merge order.}
        \label{fig:naive_Dart}
    \end{subfigure}
    \hfill
    \begin{subfigure}[b]{0.48\textwidth}
        \includegraphics[width=\textwidth]{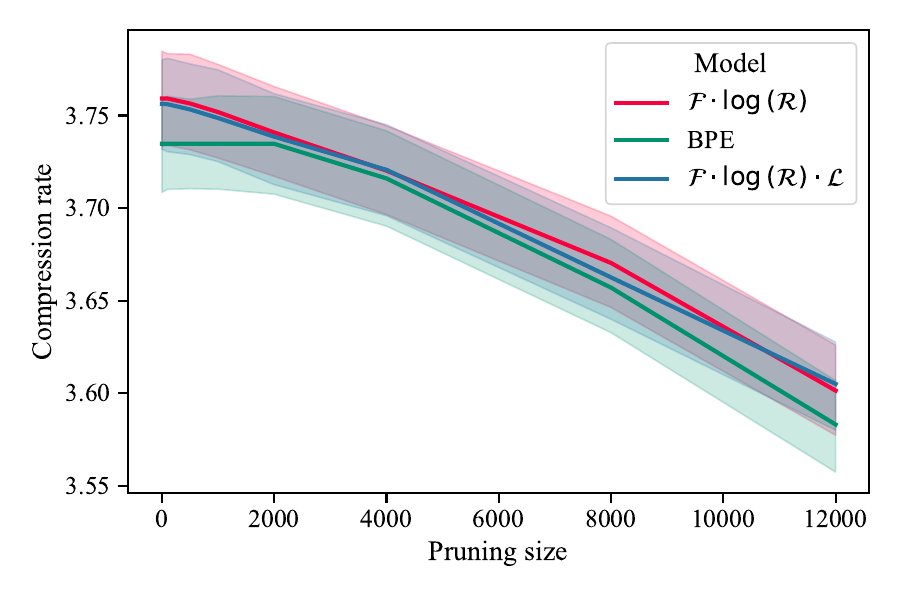}
        \subcaption{Pruning order: under-trained first.}
        \label{fig:scores_Dart}
    \end{subfigure}
    \caption{Compression rate for Dart for tokenizers with applied pruning \textbf{(a)} in the reverse order of token ids and \textbf{(b)} starting from the tokens with the lowest under-trained indicator values (distance from (0, 0) in indicator space).}
    \label{fig:pruning_Dart}
\end{figure*}

\begin{figure*}[t!]
    \begin{subfigure}[b]{0.48\textwidth}
        \includegraphics[width=\textwidth]{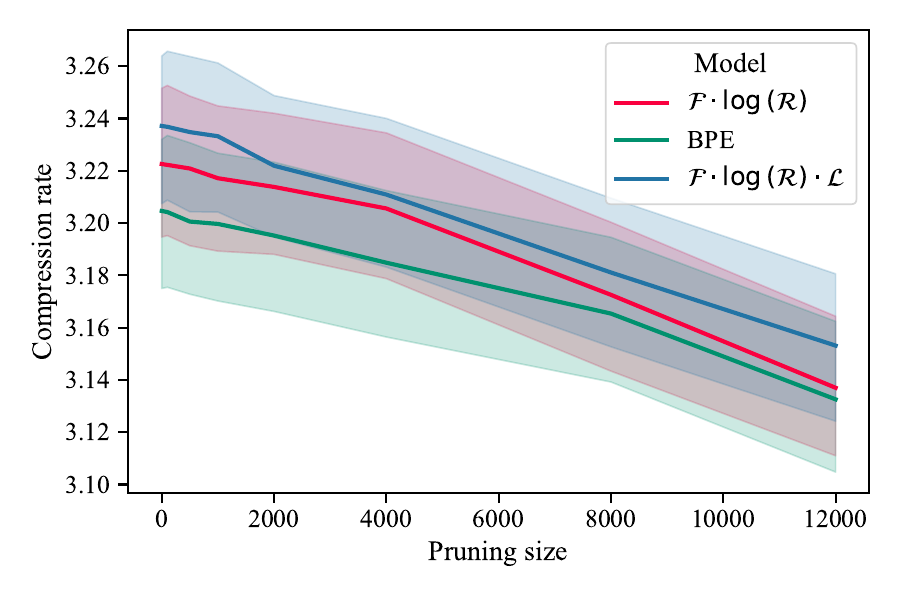}
        \subcaption{Pruning order: reverse merge order.}
        \label{fig:naive_Lua}
    \end{subfigure}
    \hfill
    \begin{subfigure}[b]{0.48\textwidth}
        \includegraphics[width=\textwidth]{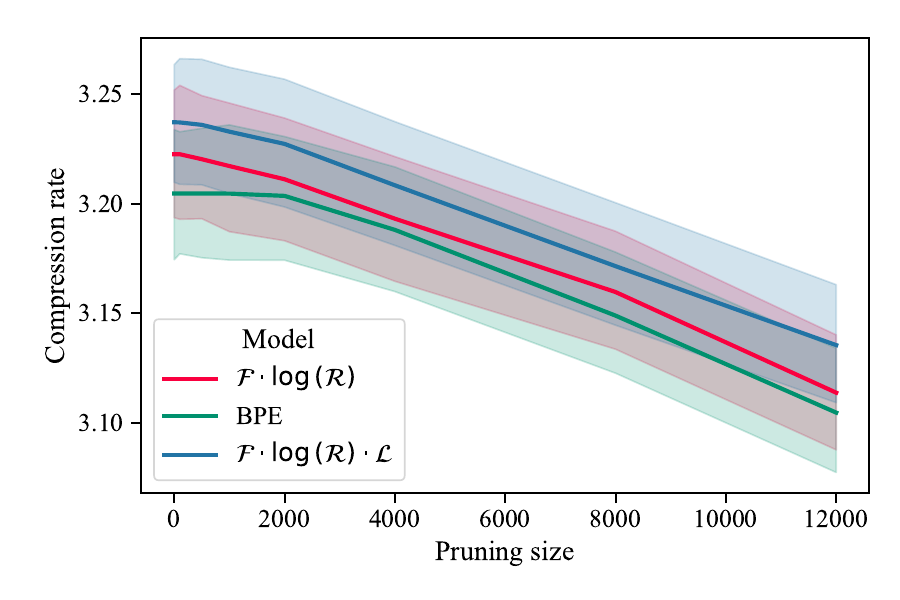}
        \subcaption{Pruning order: under-trained first.}
        \label{fig:scores_Lua}
    \end{subfigure}
    \caption{Compression rate for Lua for tokenizers with applied pruning \textbf{(a)} in the reverse order of token ids and \textbf{(b)} starting from the tokens with the lowest under-trained indicator values (distance from (0, 0) in indicator space).}
    \label{fig:pruning_Lua}
\end{figure*}

\begin{figure*}[t!]
    \begin{subfigure}[b]{0.48\textwidth}
        \includegraphics[width=\textwidth]{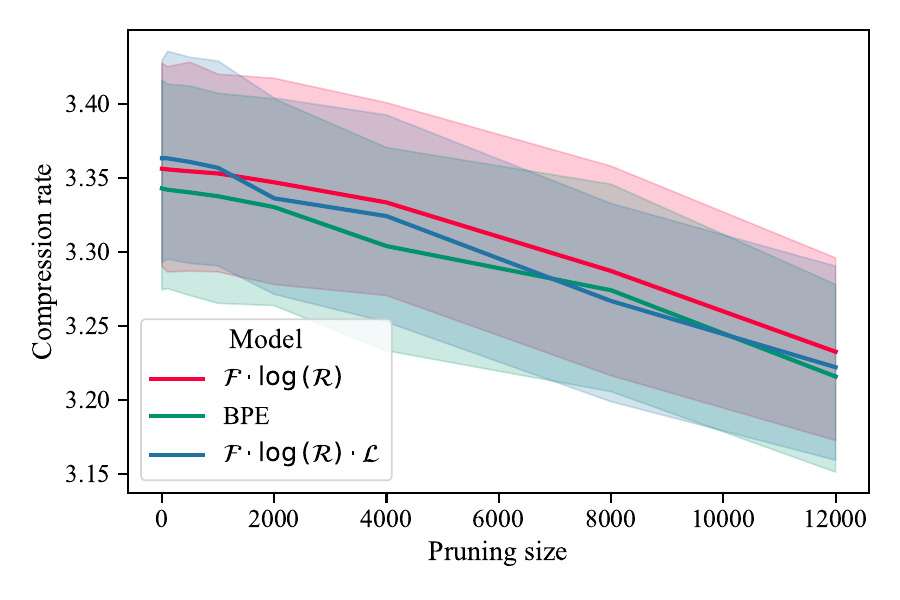}
        \subcaption{Pruning order: reverse merge order.}
        \label{fig:naive_Haskell}
    \end{subfigure}
    \hfill
    \begin{subfigure}[b]{0.48\textwidth}
        \includegraphics[width=\textwidth]{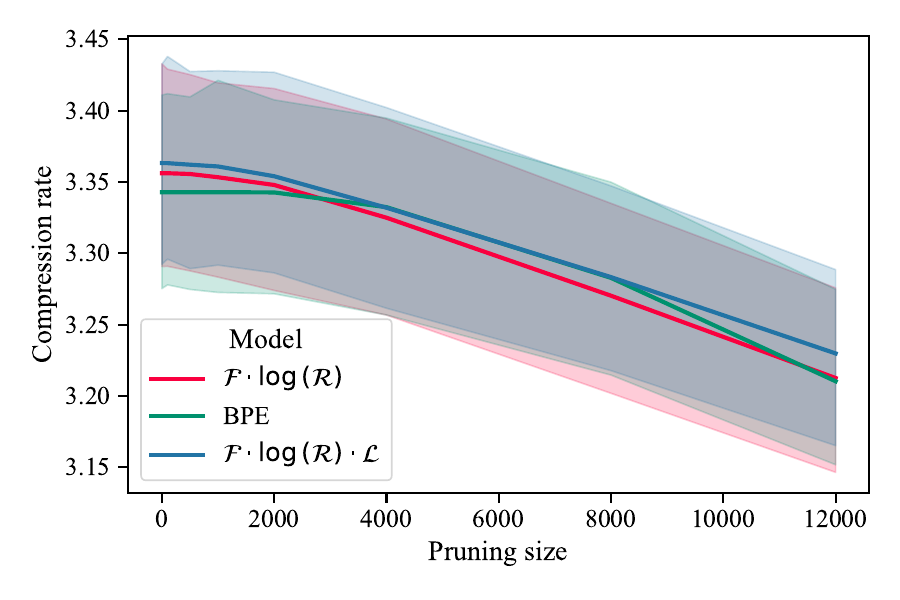}
        \subcaption{Pruning order: under-trained first.}
        \label{fig:scores_Haskell}
    \end{subfigure}
    \caption{Compression rate for Haskell for tokenizers with applied pruning \textbf{(a)} in the reverse order of token ids and \textbf{(b)} starting from the tokens with the lowest under-trained indicator values (distance from (0, 0) in indicator space).}
    \label{fig:pruning_Haskell}
\end{figure*}

\begin{figure*}[t!]
    \begin{subfigure}[b]{0.48\textwidth}
        \includegraphics[width=\textwidth]{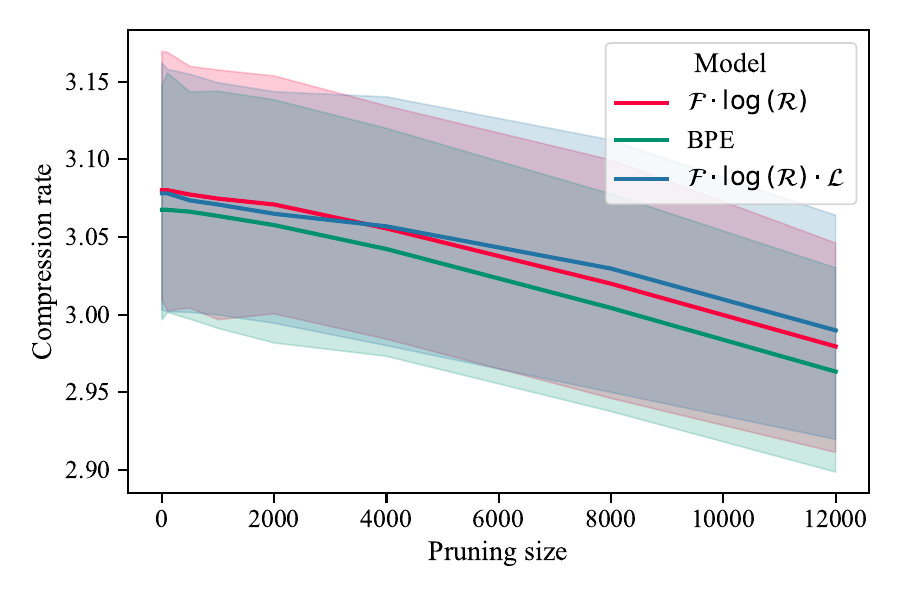}
        \subcaption{Pruning order: reverse merge order.}
        \label{fig:naive_Julia}
    \end{subfigure}
    \hfill
    \begin{subfigure}[b]{0.48\textwidth}
        \includegraphics[width=\textwidth]{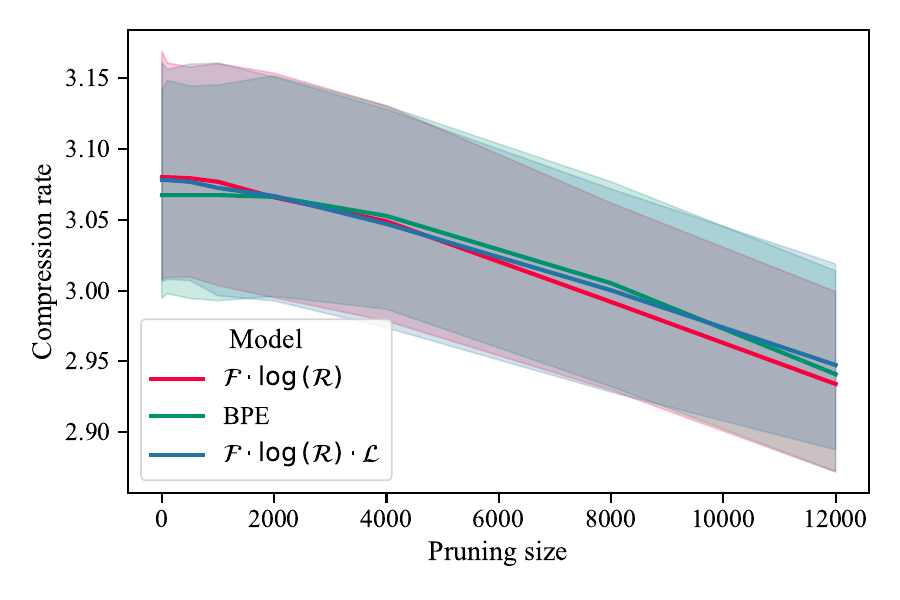}
        \subcaption{Pruning order: under-trained first.}
        \label{fig:scores_Julia}
    \end{subfigure}
    \caption{Compression rate for Julia for tokenizers with applied pruning \textbf{(a)} in the reverse order of token ids and \textbf{(b)} starting from the tokens with the lowest under-trained indicator values (distance from (0, 0) in indicator space).}
    \label{fig:pruning_Julia}
\end{figure*}

\begin{figure*}[t!]
    \begin{subfigure}[b]{0.48\textwidth}
        \includegraphics[width=\textwidth]{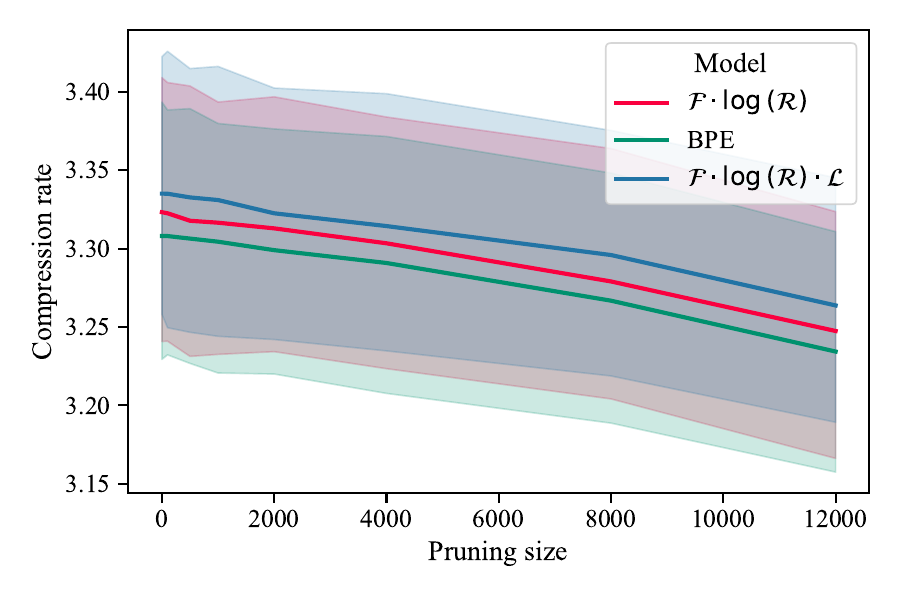}
        \subcaption{Pruning order: reverse merge order.}
        \label{fig:naive_OCaml}
    \end{subfigure}
    \hfill
    \begin{subfigure}[b]{0.48\textwidth}
        \includegraphics[width=\textwidth]{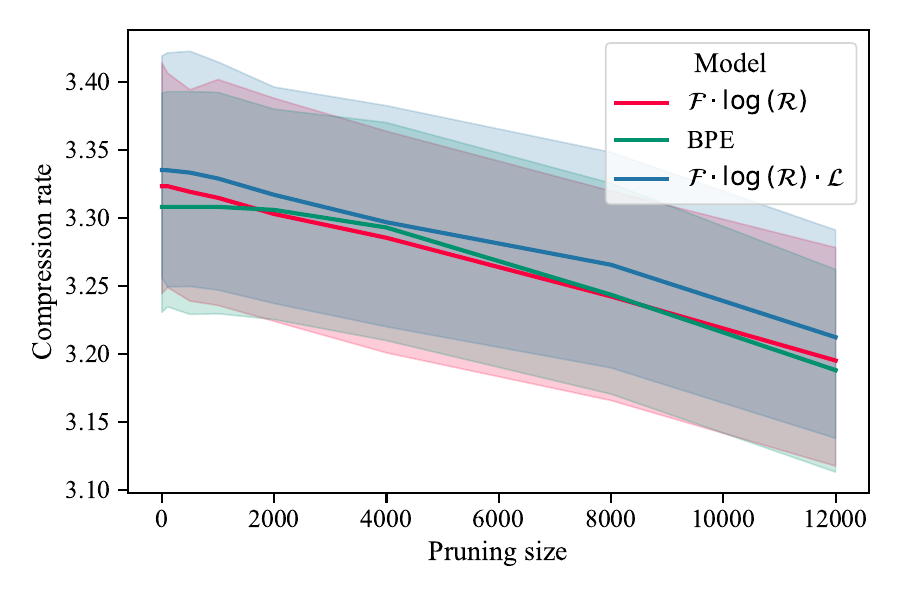}
        \subcaption{Pruning order: under-trained first.}
        \label{fig:scores_OCaml}
    \end{subfigure}
    \caption{Compression rate for OCaml for tokenizers with applied pruning \textbf{(a)} in the reverse order of token ids and \textbf{(b)} starting from the tokens with the lowest under-trained indicator values (distance from (0, 0) in indicator space).}
    \label{fig:pruning_OCaml}
\end{figure*}

In Figures~\ref{fig:pruning_Java}--\ref{fig:pruning_OCaml}, we show the compression rate of pruned tokenizers per language in the order of subsample size in the training set using pruning starting from the end of the merge list, i.e., in the reversed merge order, and leaf-based pruning that uses the distance from $(0, 0)$ in under-trained token indicator space as a priority criterion. In the first case, for most languages, the compression rate of our regularized tokenizers is consistently better than or comparable to that of BPE. In the second case, for all languages, the first several thousand removed tokens have no effect on the compression rate, as they are over-fitted to the training sample and are not utilized. In lower-resource languages, the confidence intervals for compression rates are becoming closer together. However, for some languages, such as Lua, the model with the language component shows better compression, which may suggest that this provides better support for lower-resource languages. This corresponds well to our findings in Appendix~\ref{app:multilingual}.

\section{LLM Usage Statement}

In this work, we utilized AI tools for paraphrasing and grammar correction (Grammarly) and assistance with data visualization design (Gemini, ChatGPT). We also used Claude Code to prettify and document the paper code in a GitHub repository.

\end{document}